\documentclass[letterpaper,journal]{IEEEtran}
\usepackage{amsmath,amsfonts,amssymb,amsthm}
\usepackage{algorithmic,algorithm,array}
\usepackage[caption=false,font=footnotesize,labelfont=rm,textfont=rm]{subfig}
\usepackage{textcomp,stfloats,url,verbatim,graphicx,cite}
\usepackage{xcolor,booktabs,multirow,enumitem,tikz}
\usepackage{microtype}
\newtheorem{definition}{Definition}
\newtheorem{lemma}{Lemma}
\newtheorem{theorem}{Theorem}
\theoremstyle{remark}
\newtheorem{remark}{Remark}
\begin{document}
\bstctlcite{gao_bib_style}
\title{Geometric Acceleration Obstacles: Physical Boundary Structure and Affine Safety Certificates}
\author{Xuheng Gao, Yuhan Pang, Bi-Ke Zhu, Ronghao Chen, Lingxiao Meng, Jiankun Wang, and Max Q.-H. Meng}
\maketitle
\begin{abstract}
Safe robot navigation in dynamic environments requires collision avoidance that respects motion constraints while remaining computationally efficient.
Conventional velocity-obstacle (VO) methods characterize collision-inducing motions in velocity space. 
Although some formulations account for acceleration-limited velocity transitions, their collision constraints remain tied to specific velocity-tracking models.
Acceleration obstacles (AO) address this limitation by representing collision-inducing inputs directly in acceleration space. 
However, existing AO methods either rely on conservative approximations or derive analytical structure expressions without verifying their geometric correspondence.
To address this gap, this paper presents Geometric Acceleration Obstacles (GAO), which provides, to the best of our knowledge, the first unified and physically validated characterization of AO geometry and its supporting structure.
Specifically, we derive configuration-dependent half-planes that globally support the collision set and establish their tightness, thereby separating certified collision-free acceleration inputs from collision-inducing parts.
These geometric results yield unified affine safety constraints. 
Combined with actuator limits, these constraints define a quadratic program (QP) for real-time navigation.
Extensive experiments demonstrate that GAO achieves competitive success rates and the lowest control effort among the evaluated methods, with millisecond-level computational latency.
Further circle-interchange simulations demonstrate scalability to 1,000 agents.
Physical experiments validate the approach in pedestrian encounters, reciprocal multi-agent navigation, and high-speed maneuvering.

Project website: \url{https://sites.google.com/view/gao-project}.
\end{abstract}

\section{Introduction}
\label{sec:introduction}

\IEEEPARstart{C}{ollision} avoidance is a fundamental capability for autonomous robots operating in dynamic environments, enabling them to navigate safely among moving obstacles and other agents.
This problem arises in crowded indoor navigation~\cite{crowded,safety}, warehouse logistics~\cite{warehouse}, and multi-robot systems~\cite{multi-robot}.
In practice, a robot's ability to maneuver away from collisions is constrained by its acceleration limits, which determine how rapidly it can change its speed and direction.
Biological studies of predator-prey interactions further highlight the role of acceleration and maneuverability in successful evasion~\cite{walker2005faster,wilson2018biomechanics}.
Consequently, safe avoidance requires not only selecting collision-free motions but also ensuring that the transitions toward them remain dynamically feasible under the robot's acceleration limits~\cite{coexisting2}.

\begin{figure*}[!t]
\centering
\includegraphics[width=\textwidth]{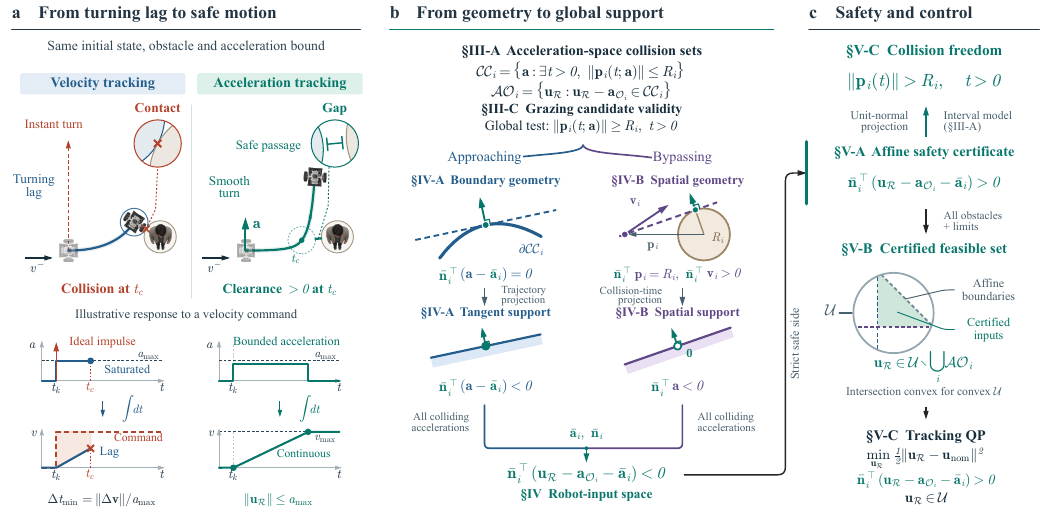}
\caption{Overview of the GAO framework. 
(a) Motivation. In the illustrated encounter with a pedestrian, saturated velocity tracking (blue) leads to contact at $t_c$, whereas GAO (green) achieves collision-free passage under the same initial conditions and acceleration bound. 
The lower $a$--$v$ plots are independent schematics of tracking lag and bounded-acceleration motion. 
(b) Geometric analysis. Trajectory-wide nonpenetration analysis identifies physically valid grazing candidates. 
Validated boundary tangents for approaching configurations and spatial supports for bypassing configurations establish global support of the collision set. 
These supports are translated into robot acceleration-input space. 
(c) Safety certification and control. The open half-planes opposite the collision set define affine safety certificates. 
Their intersection with convex actuator constraints yields a certified convex optimization of collision-free inputs for the tracking QP. 
Collision freedom holds when the certificates are strictly satisfied throughout each execution interval. }
\label{fig:gao_overview}
\end{figure*}

Conventional geometric methods, including velocity obstacles (VO)~\cite{vo} and collision cones (CC)~\cite{collisioncone}, represent collision-inducing motions in velocity space.
Reciprocal VO (RVO)~\cite{rvo} and optimal reciprocal collision avoidance (ORCA)~\cite{orca} extend this representation by distributing avoidance responsibility among interacting agents. 
However, selecting a collision-free target velocity does not guarantee collision-free motion during the transition from the current velocity, particularly under bounded acceleration (Fig.~\ref{fig:gao_overview}(a)).
% This limitation has motivated methods that account for the dynamics of reaching a selected velocity.
Dynamic window approach (DWA)~\cite{dwa} addresses this issue by restricting velocity selection to dynamically reachable commands and evaluating braking feasibility.
Other formulations incorporate control-induced motion into collision prediction:
generalized velocity obstacles (GVO)~\cite{gvo}  consider trajectories generated by control inputs, continuous control obstacles~\cite{discontinuous} account for motion-continuity constraints,
acceleration-velocity obstacles (AVO)~\cite{avo} characterize target velocities under proportional velocity tracking. 
Although these methods account for motion constraints, their collision representations remain tied to specific control parameterizations or motion-generation mechanisms.
In contrast, acceleration obstacles (AO)~\cite{ao_RAL} directly characterize collision-inducing acceleration inputs, without requiring a prescribed velocity-tracking mechanism.
However, characterizing the collision set in acceleration space does not by itself provide tractable safety constraints for real-time control.

% Although AO directly characterizes collision-inducing acceleration inputs, directly enforcing collision avoidance through its geometric representation can be difficult for real-time optimization.
To make AO suitable for real-time control, its geometric representation must be translated into tractable safety constraints.
Existing safety-control approaches derive tractable input constraints from predicted motion~\cite{brito2019mpcc,escape,predictivefilter}, collision-inevitability and reachability analyses~\cite{ics,hjreachability}, or safe-set invariance~\cite{cbf_safety,hocbf}.
More directly, Wang et al.~\cite{wang2017safety} derive affine acceleration constraints using braking-based safe sets and control barrier functions, while collision-cone CBFs~\cite{c3bf} incorporate relative-velocity geometry into barrier-based control constraints.
These studies demonstrate the value of tractable input constraints for safety-critical control. 
For AO, the challenge is to derive such constraints that exclude all collision-inducing inputs while retaining as many collision-free inputs as possible.

Existing AO studies have provided both collision-set characterization and analytical boundary construction, but important geometric questions remain.
Gonon et al.~\cite{ao_RAL} represent the finite-horizon AO as a union of acceleration-space disks and derive conservative half-plane constraints for control. 
However, their constructed local boundary curves can self-intersect, and the resulting constraints conservatively enclose the AO rather than exploit its exact boundary geometry.
Stern and Shiller~\cite{ao} subsequently derive analytical grazing candidates for AO boundaries.
Yet, grazing conditions alone do not guarantee nonpenetration over the entire trajectory, so these candidates do not necessarily correspond to the physical AO boundary. 
Moreover, the geometric structure of AO depends on the initial relative-motion configuration, requiring a unified analysis and corresponding geometric characterization.

To address these limitations, we present Geometric Acceleration Obstacles (GAO), which establishes a complete geometric characterization of AO and derives unified affine safety certificates from its validated structure. 
% We first establish a trajectory-wide nonpenetration criterion to distinguish physically valid grazing boundaries from penetrating candidates. 
% We then reveal the configuration-dependent geometry of AO and derive global supporting half-planes for both approaching and bypassing configurations. These results yield a unified affine constraint form for acceleration-space control, with proofs of safety and geometric tightness. 
% The resulting constraints, combined with actuator limits, define a certified feasible region for real-time quadratic programming.
The resulting contributions are:
\begin{enumerate}[beginpenalty=10000]
    \item \textbf{Physical characterization of AO geometry.}
    A trajectory-wide nonpenetration criterion distinguishes physically valid grazing candidates from hidden penetrating candidates, establishing their correspondence with the actual AO boundary.

    \item \textbf{Configuration-dependent global support.}
    We characterize the AO geometry under approaching and bypassing configurations and derive tangent-based and spatially derived global supports, with proofs of translational and angular tightness (Fig.~\ref{fig:gao_overview}(b)).

    \item \textbf{Unified affine certificates and experiment validation.}
    We derive pairwise robot-input certificates and combine them with convex actuator limits to obtain a convex certified inner approximation of collision-free inputs for QP-based control. Large-scale simulations and physical robot experiments validate the proposed framework (Fig.~\ref{fig:gao_overview}(c)).
\end{enumerate}

The remainder of this paper is organized as follows.
Section~\ref{sec:related} reviews related work.
Section~\ref{sec:problem_formulation} formulates the relative-motion model and examines the global validity of grazing candidates.
Section~\ref{sec:geometry_certificates} establishes the geometric safety certificates.
Section~\ref{sec:optimization} presents the control formulation.
Sections~\ref{sec:sim_exp} and~\ref{sec:phy_exp} report simulation and physical experiments, respectively.
Section~\ref{sec:conclusion} concludes this work.
Appendix~\ref{sec:support_tightness_details} provides supplementary support and tightness proofs.

\section{Related Work}
\label{sec:related}

\subsection{Geometric Collision Avoidance}
Geometric collision-avoidance methods represent potentially colliding motions as forbidden sets. 
The velocity obstacle (VO) formulation characterizes robot velocities that lead to future collisions with moving obstacles under constant-velocity motion~\cite{vo}.
Collision cones (CC) characterize collision-inducing relative velocities~\cite{collisioncone}.
Translating the cone by the obstacle velocity yields the corresponding VO in the robot's velocity space.
Nonlinear velocity obstacles (NLVOs) extend this representation to obstacles following prescribed nonlinear trajectories~\cite{nlvo}. 
For multi-agent navigation, reciprocal velocity obstacles (RVO) distribute avoidance responsibility among interacting agents~\cite{rvo}, while optimal reciprocal collision avoidance (ORCA) constructs sufficient pairwise velocity constraints that reduce action selection to a low-dimensional linear program~\cite{orca}.  
Subsequent developments address reciprocal oscillations~\cite{hrvo} and parallel collision avoidance at large scales~\cite{clearpath}.
VR-ORCA adjusts pairwise avoidance responsibilities based on neighboring-agent configurations~\cite{vr_orca}, while AVOCADO adapts responsibility allocation using cooperation estimates obtained through opinion dynamics~\cite{avocado}.
A recent survey provides a broader taxonomy of VO-based methods~\cite{vesentini2024survey}.

Beyond velocity-space representations, several approaches account for the dynamics and continuity of executable motions. 
The dynamic window approach (DWA) restricts candidate velocities to dynamically reachable commands and evaluates their braking feasibility~\cite{dwa}. 
Generalized velocity obstacles (GVO) extend collision-set construction to admissible control inputs of kinodynamic systems~\cite{gvo}, whereas continuous control obstacles (CCO) represent collision-inducing control sequences subject to motion-continuity constraints~\cite{discontinuous}. 
Acceleration-velocity obstacles (AVO) characterize target velocities whose proportional velocity-tracking responses lead to collisions, explicitly accounting for acceleration-limited transitions~\cite{avo}. 
Nonholonomic extensions further connect reciprocal avoidance to executable robot motions~\cite{nh_orca,alonsomora2018ccnr}. 
Bareiss and van den Berg unify VO, AVO, CCO, and LQR-obstacles within a generalized control-obstacle framework, extending reciprocal collision avoidance to heterogeneous robots with nonlinear dynamics~\cite{bareiss2015generalized}.

Unlike velocity-tracking formulations such as AVO, acceleration obstacles (AO) directly characterize collision-inducing accelerations under second-order motion dynamics~\cite{ao_RAL,ao}. 
This representation provides an intuitive geometric basis for analyzing the relationship between acceleration commands and future collisions.

\subsection{Safety-Critical Planning and Control}

Optimization-based planning and control provide a general framework for incorporating collision-avoidance conditions into robot motion generation.
Model predictive control (MPC) formulates collision avoidance as a finite-horizon optimization problem that accounts for robot dynamics, state and input constraints, and predicted obstacle motion. Model predictive contouring control balances path progress with collision avoidance in unstructured dynamic environments~\cite{brito2019mpcc}. For aerial robots, nonlinear MPC propagates state-estimation uncertainty into predicted separation constraints~\cite{kamel2017robust}. Distributed MPC generates multi-agent trajectories with on-demand collision constraints~\cite{luis2019dmpc}, while MADER jointly optimizes trajectories and separating planes in asynchronous multi-agent environments~\cite{escape}. These approaches enforce collision avoidance through predicted motion and spatial separation within receding-horizon or trajectory-optimization formulations. Predictive safety filters further modify nominal inputs when necessary using model-based safety policies and state and input constraints~\cite{predictivefilter}.

When surrounding agents respond to the robot's actions, prediction and control become coupled. Interactive MPC incorporates pedestrian intentions and interactions into robot navigation~\cite{chen2022interactivempc}, while SICNav jointly optimizes robot decisions and predicts human responses through bilevel optimization with an ORCA-based behavioral model~\cite{sicnav}. Beyond interaction modeling, uncertainty-aware planning accounts for imperfect state estimates and alternative future obstacle motions. Du Toit and Burdick integrate motion prediction, state estimation, and chance constraints into partially closed-loop planning~\cite{dutoit2012dynamicuncertain}. Chance-constrained nonlinear MPC bounds collision probabilities under state uncertainty~\cite{zhu2019chance}, while multimodal motion prediction~\cite{zhang2023prescient} and conformal prediction~\cite{lindemann2023conformal} provide complementary ways to represent and quantify prediction uncertainty. NH-TTC offers a related anticipatory collision-avoidance framework for acceleration-controlled, differential-drive, and car-like robots~\cite{nhttc}.

An alternative perspective is to establish safety through reachability analysis and safe-set invariance.
Inevitable collision states (ICS) identify states from which every admissible future trajectory results in collision~\cite{ics}, while Hamilton--Jacobi reachability characterizes backward-reachable sets under dynamic interactions between control and disturbance inputs~\cite{hjreachability}. Control barrier functions (CBFs) establish state-dependent conditions for forward invariance and translate them into control-input inequalities for quadratic-program-based safety enforcement~\cite{cbf_safety}. Exponential and high-order CBFs extend this approach to systems with higher relative degree~\cite{ecbf,hocbf}. Robust CBFs account for bounded disturbances and input constraints~\cite{robustcbf}, while backup CBFs construct control-invariant sets using backup policies~\cite{backupcbf}. Sampled-data CBF formulations further address safety between control updates and in the presence of input delays~\cite{sampledcbf,delaycbf}. These guarantees depend on the adopted dynamics, safe sets, and execution assumptions. MPC-CBF further combines trajectory optimization with discrete-time CBF conditions~\cite{zeng2021safety}.

Geometry-informed safety constraints provide a more direct connection between collision geometry and control inputs. Collision-cone CBFs incorporate relative-velocity geometry into barrier constraints for acceleration-controlled unicycle and bicycle models~\cite{c3bf}. Subsequent experiments validate these controllers on ground robots~\cite{c3bf_exp}. For bounded-acceleration robots, Wang et al. derive affine acceleration-input constraints from relative separation and braking capability using control barrier functions~\cite{wang2017safety}. 
These studies demonstrate how collision-related geometric and safety conditions can be translated into tractable acceleration-input constraints. However, they do not address the geometric characterization of the AO collision set itself or derive affine certificates directly from its boundary structure.

\subsection{Acceleration-Obstacle Geometry and Safety Constraints}
Acceleration obstacles extend velocity-obstacle representations to collision-inducing acceleration inputs under second-order motion dynamics. 
Gonon et al.~\cite{ao_RAL} formulate the finite-horizon AO as an exact union of acceleration-space disks, each corresponding to accelerations that cause collision at a particular future time. 
They analyze the local boundary curves and their normal directions, showing that these curves may self-intersect and include points that do not belong to the actual AO boundary. 
To obtain tractable control constraints, they develop a sampling-free method based on spatial occlusion geometry and construct conservative half-plane enclosures with proven containment guarantees. 
This approach provides an important connection between acceleration-space collision sets and affine safety constraints. 
However, the resulting constraints conservatively enclose the collision set rather than directly exploiting its physical geometry.

Stern and Shiller~\cite{ao} subsequently derive analytical AO boundary candidates from grazing trajectories by solving contact and tangency conditions in local normal and tangential coordinates. 
Their formulation provides an explicit parameterization of candidate boundary points and extends the analysis to nonlinear acceleration obstacles. 
However, satisfying grazing conditions and requiring a positive contact time do not establish nonpenetration over the entire trajectory. 
A candidate may satisfy the prescribed contact conditions while the corresponding trajectory penetrates the obstacle at another time. 
Consequently, the analytical candidate curve does not necessarily coincide with the physical AO boundary. 
Moreover, a complete geometric characterization must account for the dependence of AO structure on the initial relative-motion configuration, rather than relying on boundary candidates alone.

These limitations reveal two distinct requirements for deriving affine safety constraints directly from AO geometry. 
First, candidate boundary points must be validated against the entire future trajectory to establish their physical correspondence with the actual collision-set boundary. 
Second, the resulting boundary structure must be analyzed across different relative-motion configurations to identify geometric properties that support the complete collision set.  
GAO addresses these requirements by establishing trajectory-wide boundary validity and deriving configuration-dependent geometric supports, which together provide a unified basis for affine safety certificates.

\section{Problem Formulation}
\label{sec:problem_formulation}
We formulate robot–obstacle motion in relative coordinates, define the corresponding collision set in acceleration space and then assess the physical validity.

\subsection{Collision Model}
\label{subsec:collision_model}
Let $\mathcal{W} \subset \mathbb{R}^n$ denote the workspace. The analysis focuses on planar motion, $n=2$.
The environment contains $m$ dynamic obstacles, denoted by $\mathcal{O}_i$ for $i \in \{1,\dots,m\}$.
Each obstacle is modeled as a compact set with reference position $\mathbf{p}_{\mathcal{O}_i}(t) \in \mathbb{R}^2$.
We consider a holonomic robot $\mathcal{R}$ with reference position $\mathbf{p}_{\mathcal{R}}(t) \in \mathbb{R}^2$.
The robot state is
\begin{equation}
\mathbf{x}(t) = \big[ \mathbf{p}_{\mathcal{R}}^\top(t), \mathbf{v}_{\mathcal{R}}^\top(t) \big]^\top \in \mathcal{X},
\end{equation}
where $\mathcal{X} \subseteq \mathbb{R}^{4}$ is the state space, and $\mathbf{v}_{\mathcal{R}}(t) \in \mathbb{R}^2$ represents the velocity.
We model the robot's motion as a double-integrator system,
\begin{equation}
\dot{\mathbf{x}} = \begin{bmatrix} \dot{\mathbf{p}}_{\mathcal{R}} \\ \dot{\mathbf{v}}_{\mathcal{R}} \end{bmatrix} = \begin{bmatrix} \mathbf{v}_{\mathcal{R}} \\ \mathbf{u}_{\mathcal{R}} \end{bmatrix},
\end{equation}
where $\mathbf{u}_{\mathcal{R}} \in \mathcal{U} \subset \mathbb{R}^2$ is the acceleration command, and $\mathcal{U}$ represents actuator limits.

We use configuration-space inflation to convert finite-size collision checking into point-robot collision checking \cite{co}.
For obstacle $i$, the C-space obstacle is
\begin{equation}
\mathcal{CO}_i = \mathcal{O}_i \oplus (-\mathcal{R})
= \{ \mathbf{o} - \mathbf{r} \mid \mathbf{o} \in \mathcal{O}_i, \mathbf{r} \in \mathcal{R} \},
\end{equation}
where $\oplus$ denotes the Minkowski sum.
For real-time control, we use disk bounds for the robot and obstacles. Let $r_{\mathcal R}$ be the robot radius and $r_{\mathcal O_i}$ be the radius of obstacle $i$. The corresponding C-space obstacle has a radius
\begin{equation}
R_i = r_{\mathcal{R}} + r_{\mathcal{O}_i}.
\end{equation}

For notational simplicity, we now consider one robot-obstacle pair and omit the obstacle index $i$.
Let $\mathbf{p}(t)=\mathbf{p}_{\mathcal{R}}(t)-\mathbf{p}_{\mathcal{O}}(t)$ and
$\mathbf{v}(t)=\mathbf{v}_{\mathcal{R}}(t)-\mathbf{v}_{\mathcal{O}}(t)$ denote the relative position and velocity.
Set the current time to $t_0 = 0$, with relative states $\mathbf{p}_0 \triangleq \mathbf{p}(0)$ and $\mathbf{v}_0 \triangleq \mathbf{v}(0)$.
Throughout the geometric analysis, $\|\mathbf p_0\|>R$.

Within each control cycle, the robot and obstacle are modeled with constant accelerations.
Let $\mathbf{u}_{\mathcal R}$ and $\mathbf{a}_{\mathcal O}$ denote the robot acceleration input and the obstacle acceleration, respectively. 
The relative acceleration is $\mathbf{a} = \mathbf{u}_{\mathcal{R}} - \mathbf{a}_{\mathcal{O}}$.
Accordingly, the relative motion evolves as
\begin{equation}\label{eq:relative_trajectory}
    \mathbf{p}(t; \mathbf{a}) = \mathbf{p}_0 + \mathbf{v}_0 t + \frac{1}{2} \mathbf{a} t^2, \quad  t \ge 0.
\end{equation}
In relative coordinates, the C-space obstacle becomes $\mathcal{CO}=\{\mathbf q \in \mathbb{R}^2:\|\mathbf q\|\le R\}$.
A future collision occurs if $\|\mathbf{p}(t;\mathbf{a})\|\le R$ for some $t>0$.
A strictly collision-free trajectory therefore requires $\|\mathbf p(t;\mathbf{a})\|>R$ for all $t>0$.

\begin{definition}[Collision Cone]
The collision cone, denoted by $\mathcal{CC}(\mathbf{p}_0, \mathbf{v}_0)$, is defined as the set of all relative accelerations that lead to a future collision:
\begin{equation}
\mathcal{CC}(\mathbf{p}_0, \mathbf{v}_0)
\triangleq
\left\{
\mathbf{a} \in \mathbb{R}^2 \,\middle|\,
\exists t > 0,\;
\left\| \mathbf{p}(t; \mathbf{a}) \right\| \le R
\right\}.
\end{equation}
\end{definition}

\begin{remark}
\label{rem:collision_cone_terminology}
The terminology follows the velocity obstacles literature, but $\mathcal{CC}$ is not generally a linear cone in acceleration space. Its boundary can be nonlinear.
\end{remark}

\begin{definition}[Acceleration Obstacles]\label{def:ao}
Mapping the collision cone to robot-input space defines the Acceleration Obstacles $\mathcal{AO}(\mathbf p_0,\mathbf v_0,\mathbf a_{\mathcal O})$:
\begin{equation}
\mathcal{AO}(\mathbf{p}_0, \mathbf{v}_0, \mathbf{a}_{\mathcal{O}})
\triangleq
\left\{
\mathbf{u} \in \mathbb R^2
\;\middle|\;
\mathbf{u} - \mathbf{a}_{\mathcal{O}} \in \mathcal{CC}(\mathbf{p}_0, \mathbf{v}_0)
\right\}.
\end{equation}
\end{definition}

\begin{figure}[t]
\centering
\includegraphics[width=\linewidth]{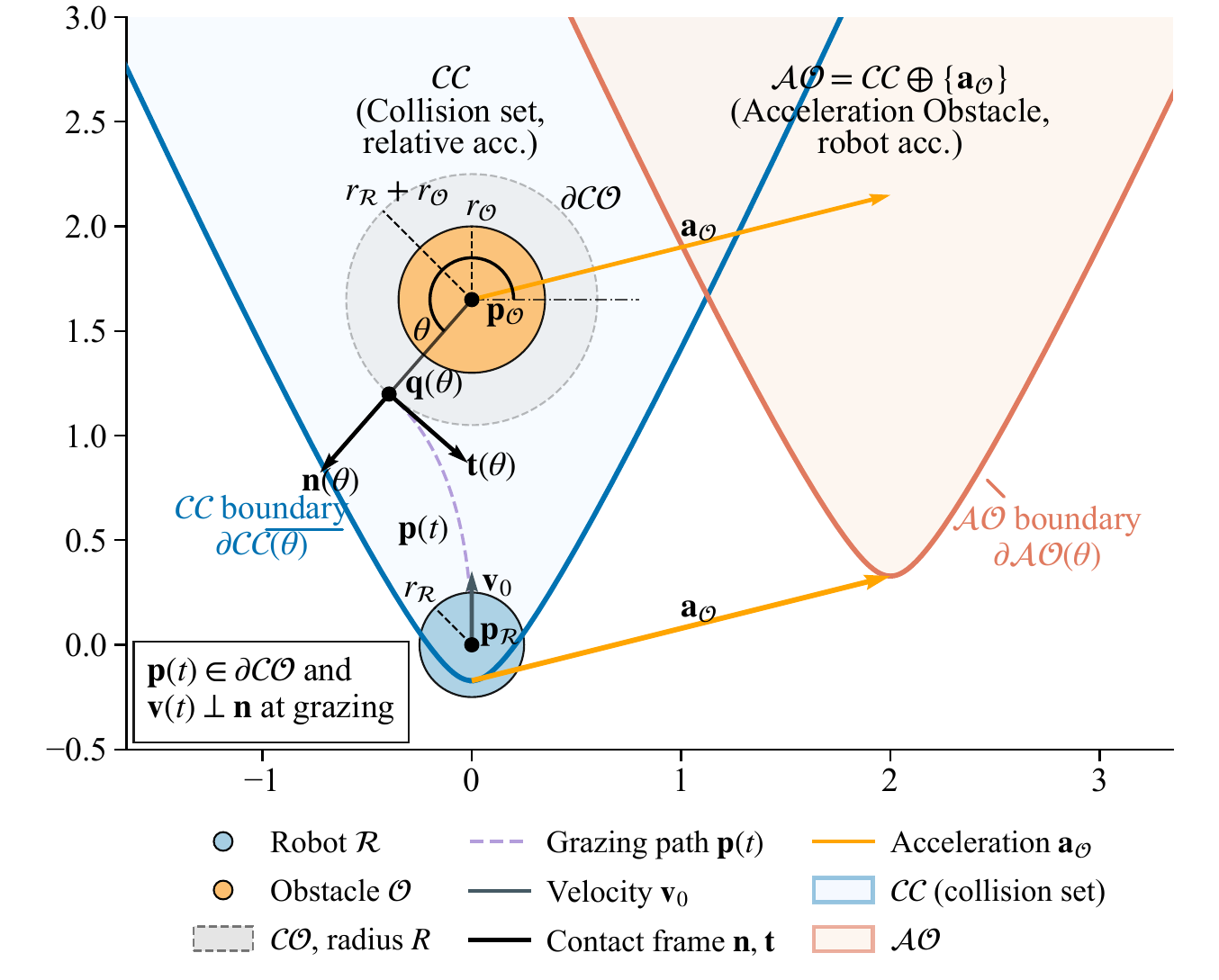}
\caption{Collision cone ($\mathcal{CC}$) and acceleration obstacles ($\mathcal{AO}$). The robot $\mathcal{R}$ (radius $r_\mathcal{R}$) and obstacle $\mathcal{O}$ (radius $r_\mathcal{O}$) define the relative C-space obstacle ($\mathcal{CO}$) of radius $R = r_\mathcal{R} + r_\mathcal{O}$. In the valid approaching domain, the boundary ($\partial \mathcal{CC}$) is parameterized by accelerations $\mathbf a(\theta)$ generating grazing trajectories $\mathbf{p}(t;\mathbf{a})$ that contact $\partial \mathcal{CO}$ at $\mathbf{q}(\theta)$. The inadmissible robot inputs form $\mathcal{AO}$, obtained by translating $\mathcal{CC}$ by obstacle acceleration $\mathbf{a}_{\mathcal{O}}$: $\mathcal{AO} = \mathcal{CC} \oplus \{ \mathbf{a}_{\mathcal{O}} \}$.}
\label{fig:sketch}
\end{figure}

As illustrated in Fig.~\ref{fig:sketch}, the mapping is a pure translation, $\mathcal{AO} = \mathcal{CC} \oplus \{ \mathbf{a}_{\mathcal{O}} \}$. It preserves the boundary geometry, tangent directions, and curvature of $\mathcal{CC}$.
We therefore analyze $\mathcal{CC}$ and translate the resulting constraints to robot-input space for control.

The boundary of $\mathcal{CC}$ marks the transition to collision-free accelerations and provides a geometric reference for selecting avoidance controls. We next derive analytical candidates for this boundary from trajectories that make tangential contact with the obstacle.

\subsection{Grazing Candidate Construction}
\label{subsec:grazing_construction}
We follow the derivation of \cite{ao} to construct analytical grazing candidates.
Consider a grazing trajectory that reaches $\partial\mathcal{CO}$ at a contact time $t_c>0$ with zero normal velocity. For the disk obstacle, the contact and tangency conditions are
\begin{equation}\label{eq:grazing_contact_conditions}
\begin{cases}
\|\mathbf p(t_c)\|=R,\\
\mathbf p(t_c)\cdot\mathbf v(t_c)=0.
\end{cases}
\end{equation}

We parameterize $\partial\mathcal{CO}$ by an angular variable $\theta$, as illustrated in Fig.~\ref{fig:sketch}. At a candidate contact point $\mathbf q(\theta)$, define the local orthonormal frame
\begin{equation}
\mathbf n(\theta)=\begin{bmatrix}\cos\theta\\\sin\theta\end{bmatrix},\qquad
\mathbf t(\theta)=\begin{bmatrix}-\sin\theta\\\cos\theta\end{bmatrix},
\end{equation}
so that $\mathbf q(\theta)=R\mathbf n(\theta)$.
Let $\mathbf s(\theta)=\mathbf q(\theta)-\mathbf p_0$ be the displacement from the current relative position to the candidate contact point. Projecting $\mathbf s$ and $\mathbf v_0$ onto the local frame gives
\begin{equation}
\begin{aligned}
s_n(\theta)&=\mathbf s(\theta)\cdot\mathbf n(\theta),&
s_t(\theta)&=\mathbf s(\theta)\cdot\mathbf t(\theta),\\
v_n(\theta)&=\mathbf v_0\cdot\mathbf n(\theta),&
v_t(\theta)&=\mathbf v_0\cdot\mathbf t(\theta).
\end{aligned}
\end{equation}

We solve for the candidate acceleration $\mathbf a(\theta)$ in the local frame, omitting explicit dependence on $\theta$ in the intermediate steps.
With $\mathbf p(t_c)=\mathbf q(\theta)$, the normal projection of~\eqref{eq:relative_trajectory} and the tangency condition give
\begin{equation}\label{eq:grazing_component_conditions}
\begin{cases}
v_nt_c+\tfrac12a_nt_c^2=s_n,\\
v_n+a_nt_c=0.
\end{cases}
\end{equation}
For $s_nv_n\ne0$, solving these equations yields
\begin{equation}\label{eq:grazing_normal_solution}
a_n=-\frac{v_n^2}{2s_n},\qquad t_c=\frac{2s_n}{v_n}.
\end{equation}
The tangential displacement must satisfy
\begin{equation}\label{eq:grazing_tangential_condition}
v_tt_c+\tfrac12a_tt_c^2=s_t.
\end{equation}
Substituting $t_c=2s_n/v_n$ into~\eqref{eq:grazing_tangential_condition} gives
\begin{equation}\label{eq:grazing_tangential_solution}
a_t=\frac{s_tv_n^2}{2s_n^2}-\frac{v_tv_n}{s_n}.
\end{equation}
Combining the normal and tangential components gives the analytical grazing candidate curve
\begin{equation}\label{eq:cc_boundary}
\mathbf a(\theta)
=\underbrace{\left(-\frac{v_n^2}{2s_n}\right)}_{a_n(\theta)}\mathbf n(\theta)
+\underbrace{\left(\frac{s_tv_n^2}{2s_n^2}-\frac{v_tv_n}{s_n}\right)}_{a_t(\theta)}\mathbf t(\theta).
\end{equation}

Future contact requires $t_c>0$. For the nonsingular parameterization, substituting the contact time from~\eqref{eq:grazing_normal_solution} yields
\begin{equation}\label{eq:causality}
t_c>0\quad\Longleftrightarrow\quad\frac{2s_n}{v_n}>0
\quad\Longleftrightarrow\quad s_nv_n>0.
\end{equation}
The nonsingular candidate domain and analytical envelope are
\begin{equation}\label{eq:envelope_domain}
\Theta_g=\{\theta\bmod 2\pi:s_nv_n>0\},\qquad
\Gamma_g=\{\mathbf a(\theta):\theta\in\Theta_g\}.
\end{equation}

Within $\Theta_g$, the admissibility of a candidate contact depends on the initial relative motion.
We distinguish approaching and bypassing configurations using the constant-velocity ray $\mathbf p_0+t\mathbf v_0$, $t\ge0$, and the following line-of-sight cone.

\begin{definition}[Line-of-Sight Cone]
The line-of-sight cone is the set of directions from $\mathbf{p}_0$ that intersect $\mathcal{CO}$, bounded by the two tangent rays from $\mathbf{p}_0$ to $\partial \mathcal{CO}$. It is defined as
\begin{equation}\label{eq:los_cone}
\mathcal{K} \triangleq
\left\{\mathbf{\eta} \in \mathbb{R}^2
\;\middle|\;\exists \tau > 0,\;\mathbf{p}_0 + \tau \mathbf{\eta} \in \mathcal{CO}\right\}.
\end{equation}
\end{definition}

A strict approaching configuration occurs when the constant-velocity ray enters the interior of $\mathcal{CO}$, equivalently, $\mathbf v_0\in\operatorname{int}(\mathcal K)$.
For nonzero relative velocity, a strict bypassing configuration occurs when the ray does not intersect $\mathcal{CO}$ and maintains positive clearance.
Zero relative velocity and nonzero velocity directed toward a tangent point are treated as degenerate cases and addressed separately in Section~\ref{subsec:bypassing_geometry}.

To characterize admissible grazing contact locations in approaching motion, we distinguish the visible and invisible surfaces of the obstacle as viewed from $\mathbf p_0$.
These surfaces are defined by the sign of the normal displacement $s_n(\theta)$:
\begin{equation}
\begin{aligned}
\mathcal{S}_{vis} &= \{ \mathbf q(\theta)\in\partial\mathcal{CO} \mid s_n(\theta) < 0 \}, \\
\mathcal{S}_{inv} &= \{ \mathbf q(\theta)\in\partial\mathcal{CO} \mid s_n(\theta) \ge 0 \}.
\end{aligned}
\end{equation}

\begin{theorem}[Admissible Approaching Grazing Domain] \label{thm:admissibility}
For the disk C-space obstacle with $\mathbf v_0\in\operatorname{int}\mathcal K$, any physically admissible grazing contact satisfies $\mathbf q(\theta)\in\mathcal S_{vis}$ and
\begin{equation}
s_n(\theta)<0,\qquad v_n(\theta)<0.
\end{equation}
Its contact parameter therefore lies in
\begin{equation}
\label{eq:theta_app}
\Theta_{\mathrm{app}} = \left\{ \theta \in [0, 2\pi) \;\middle|\; s_n(\theta) < 0, \; v_n(\theta) < 0 \right\}.
\end{equation}
\end{theorem}

\begin{figure}[htbp]
\centering
\includegraphics[width=0.75\linewidth]{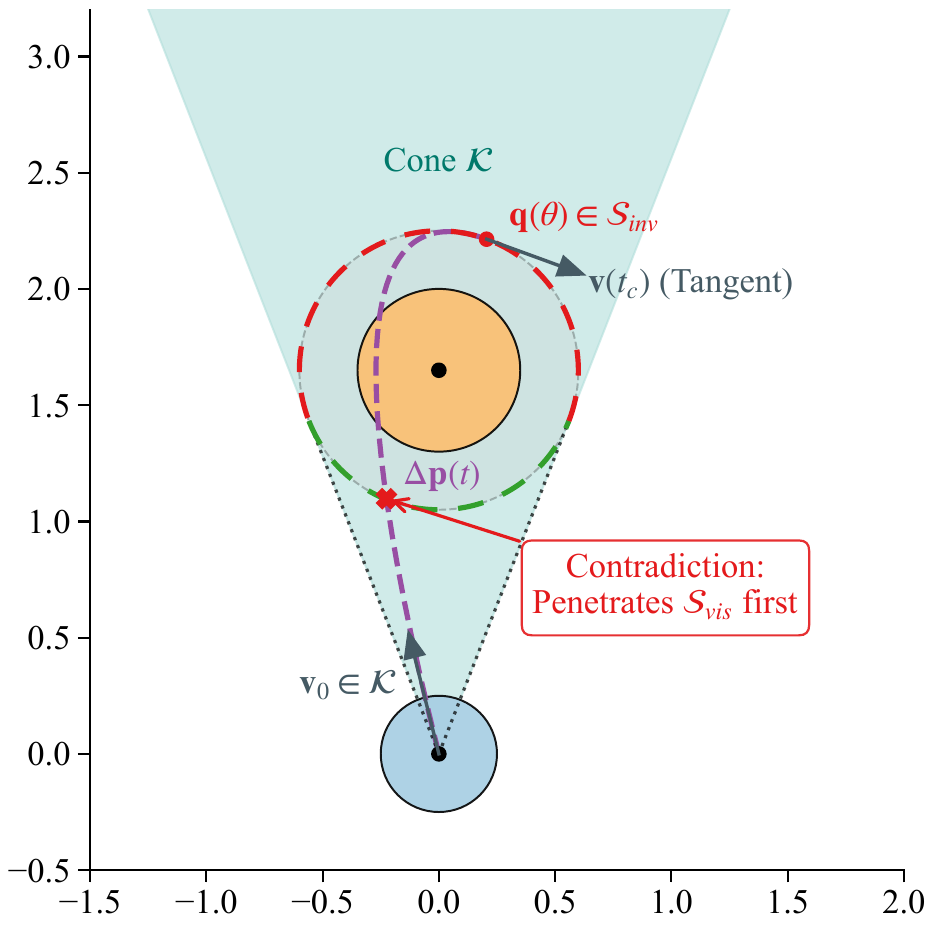}
\caption{First-contact geometry in Theorem~\ref{thm:admissibility}. For $\mathbf v_0\in\operatorname{int}\mathcal K$, a grazing trajectory to $\mathbf q(\theta)\in\mathcal S_{inv}$ satisfies $\Delta\mathbf p(t)\in\operatorname{int}\mathcal K$ for $0<t<t_c$. It therefore crosses $\mathcal S_{vis}$ and penetrates the obstacle before $t_c$, contradicting first-contact admissibility.}
\label{fig:contradiction}
\end{figure}

\begin{IEEEproof}
A grazing contact at $s_n=0$ requires $v_n=0$ by~\eqref{eq:grazing_component_conditions}, so the initial velocity lies along the contact tangent, contrary to strict approaching motion.

Suppose a physically admissible grazing event at $t_c>0$ occurs at $\mathbf q(\theta)\in\mathcal S_{inv}$ with $s_n>0$. The displacement $\Delta\mathbf p(t)=\mathbf p(t)-\mathbf p_0=\mathbf v_0t+\tfrac12\mathbf a t^2$ satisfies $\Delta\mathbf p(t_c)=\mathbf s(\theta)$, which determines
\begin{equation}\label{eq:traj_acc}
\mathbf a=\frac{2}{t_c^2}\bigl(\mathbf s(\theta)-\mathbf v_0t_c\bigr).
\end{equation}
Substitution gives the positive-combination representation
\begin{equation}\label{eq:traj_proof}
\Delta\mathbf p(t)
=\underbrace{\left(t-\frac{t^2}{t_c}\right)}_{c_1(t)}\mathbf v_0
+\underbrace{\frac{t^2}{t_c^2}}_{c_2(t)}\mathbf s(\theta).
\end{equation}
For $0<t<t_c$, both coefficients are positive. Since $\mathbf v_0\in\operatorname{int}\mathcal K$ and $\mathbf s(\theta)\in\mathcal K$, convexity of the line-of-sight cone gives $\Delta\mathbf p(t)\in\operatorname{int}\mathcal K$ throughout this interval.

Such a trajectory must penetrate the visible surface $\mathcal S_{vis}$ before reaching $\mathbf q(\theta)$ (Fig.~\ref{fig:contradiction}), contradicting admissibility .
Therefore, under strict approaching motion, any physically admissible grazing contact must occur on the visible surface $\mathcal S_{vis}$.
It follows that $s_n<0$ and $v_n=2s_n/t_c<0$, so the contact parameter must belong to $\Theta_{\mathrm{app}}$.
\end{IEEEproof}

For bypassing motion, we retain only the future-contact condition~\eqref{eq:causality} to define the analytical candidate domain:
\begin{equation}\label{eq:theta_byp}
\Theta_{\mathrm{byp}}
=\{\theta\bmod 2\pi:s_n(\theta)v_n(\theta)>0\}
=\Theta_g.
\end{equation}

Equation~\eqref{eq:cc_boundary} gives the grazing candidates, with domains defined for approaching and bypassing motion.
A physical boundary trajectory must satisfy the grazing conditions at any finite contact time, but grazing alone does not exclude penetration at other times.
We therefore examine global nonpenetration to determine when these candidates correspond to the physical collision boundary.

\subsection{Grazing Candidate Validity}
\label{subsec:grazing_validity}
We assess candidate validity by evaluating clearance along the entire future trajectory.

For a candidate $\mathbf a=\mathbf a(\theta)$, define the clearance function $g_{\mathbf a}(t):=\|\mathbf p(t;\mathbf a)\|^2-R^2$. It is positive outside the disk, zero at contact, and negative during penetration. Let $z=t-t_c$ denote the time offset from contact. Define $w=v_t+a_tt_c$ as the tangential velocity at contact and $A=\|\mathbf a\|$ as the acceleration magnitude. Using $\mathbf p(t_c)=R\mathbf n$ and $\mathbf v(t_c)=w\mathbf t$, the trajectory clearance can be written as
\begin{equation}\label{eq:grazing_factorization}
\begin{aligned}
g_{\mathbf a}(t)&=z^2Q_\theta(z),\\
Q_\theta(z)&=\tfrac14A^2z^2+wa_tz+w^2+Ra_n.
\end{aligned}
\end{equation}
To test global nonpenetration for $t>0$, we minimize $Q_\theta(t-t_c)$ over $t\ge0$.
Including $t=0$ does not affect this test because the initial separation satisfies $d=\|\mathbf p_0\|>R$.
For $A>0$, let $z_v=-2wa_t/A^2$ and define $t_Q=\max\{0,t_c+z_v\}$ as the nonnegative time at which this minimum is attained:
\begin{equation}\label{eq:qmin_closed}
\begin{aligned}
Q_{\min}(\theta)&:=\min_{t\ge0}Q_\theta(t-t_c)\\
&=\begin{cases}
(d^2-R^2)/t_c^2,&t_Q=0,\\
Ra_n+w^2a_n^2/A^2,&t_Q>0.
\end{cases}
\end{aligned}
\end{equation}

\begin{lemma}[Global nonpenetration of grazing trajectories]\label{thm:grazing_validity}
Assume $d>R>0$, a nonzero constant relative acceleration $\mathbf a$, and a finite grazing time $t_c>0$. Then $Q_{\min}\ge0$ if and only if $\|\mathbf p(t;\mathbf a)\|\ge R$ for every $t\ge0$. If $Q_{\min}<0$, the trajectory strictly penetrates the disk at some positive time and $\mathbf a\in\operatorname{int}\mathcal{CC}$.
\end{lemma}
\begin{IEEEproof}
If $Q_{\min}\ge0$, the factorization gives $g_{\mathbf a}(t)\ge0$ for all $t\ge0$. Conversely, if $Q_{\min}<0$, continuity gives some $z>-t_c$, $z\ne0$, with $Q_\theta(z)<0$, hence $g_{\mathbf a}(t_c+z)<0$. This strict penetration persists under small acceleration perturbations by continuity of~\eqref{eq:relative_trajectory}, so $\mathbf a\in\operatorname{int}\mathcal{CC}$.
\end{IEEEproof}

By Lemma~\ref{thm:grazing_validity}, candidates with $Q_{\min}<0$ lie in $\operatorname{int}\mathcal{CC}$, whereas those with $Q_{\min}\ge0$ generate globally nonpenetrating trajectories. We apply this criterion to the approaching and bypassing configurations defined in Section~\ref{subsec:grazing_construction}.

\subsubsection{Approaching Boundary Correspondence}
\label{subsec:approaching_validity}
Under the model in Section~\ref{subsec:collision_model}, let $\mathbf v_0\in\operatorname{int}\mathcal K$.
Consider a candidate relative acceleration $\mathbf a=\mathbf a(\theta)$ with $\theta\in\Theta_{\mathrm{app}}$ from Theorem~\ref{thm:admissibility}.
The domain conditions $s_n<0$ and $v_n<0$ give $t_c=2s_n/v_n>0$ and $a_n=-v_n^2/(2s_n)>0$. Both values in~\eqref{eq:qmin_closed} are therefore positive. Hence $Q_{\min}>0$; Lemma~\ref{thm:grazing_validity} and~\eqref{eq:grazing_factorization} establish global nonpenetration with grazing contact only at $t_c$. Thus these approaching grazing candidates lie on the physical collision boundary, $\mathbf a\in\partial\mathcal{CC}$.

\subsubsection{Limits of Bypassing Grazing Candidates}
\label{subsec:bypassing_validity_limit}
For bypassing motion, the future-contact domain also admits $s_n>0$ and $v_n>0$, giving $a_n<0$. For these candidates, global nonpenetration is determined by checking $Q_{\min}\ge0$ using~\eqref{eq:qmin_closed}. If $Q_{\min}<0$, the trajectory penetrates the obstacle despite satisfying the grazing conditions, and Lemma~\ref{thm:grazing_validity} gives $\mathbf a\in\operatorname{int}\mathcal{CC}$.

Thus, the bypassing grazing candidate curve may contain points inside the collision set and need not coincide with its physical boundary. Section~\ref{sec:geometry_certificates} discusses this difference in more detail.

\section{Collision-Set Geometry and Safety Certificates}
\label{sec:geometry_certificates}
Based on the collision-set characterization established in Section~\ref{sec:problem_formulation}, this section derives affine safety certificates from the geometric properties of the collision set. The construction exploits either local boundary structure or global one-sided support, depending on the available geometric information.
For approaching configurations, curvature and local convexity characterize validated boundary branches, while trajectory projection establishes support of the entire collision set by their tangents and certifies the opposite half-planes. For bypassing configurations, although the collision set still admits a geometric boundary, its local grazing representation does not directly yield a reliable supporting constraint; instead, a global one-sided support is constructed from the collision-set geometry.

\subsection{Approaching Configurations}
\label{subsec:approaching_geometry}
Constructing an affine safety constraint requires identifying a line that separates safe accelerations from the collision set. We first examine the local geometry of the approaching boundary branches, then establish the global separation needed for a safety certificate.

\subsubsection{Local Boundary Geometry}
\label{subsec:approaching_local_convexity}
\begin{figure}[t]
\centering
\includegraphics[width=\columnwidth]{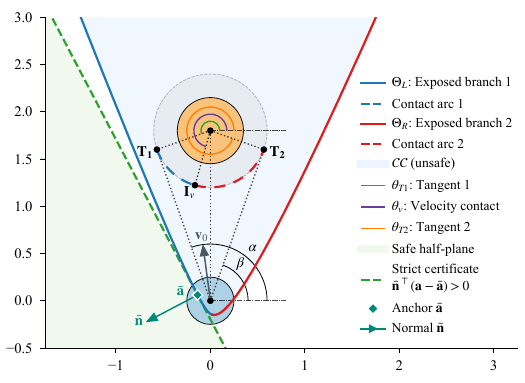}
\caption{Approaching boundary geometry and tangent safety certificate. The velocity-intersection parameter $\theta_v$ partitions the admissible domain into $\Theta_L$ and $\Theta_R$, which map to the solid acceleration-space branches $\Gamma_L$ and $\Gamma_R$. The tangent at $\bar{\mathbf a}$ with normal $\bar{\mathbf n}$ separates $\mathcal{CC}$ from the shaded safe half-plane $H_{\mathrm{AO}}(\bar\theta)$. The dashed support line, including the anchor, is excluded from the certified half-plane. 
% The spatial configuration is superimposed on the relative-acceleration geometry, and light blue shading denotes $\mathcal{CC}$.
}
\label{fig:approaching_geometry_certificate}
\end{figure}

Let $\theta_{T_1}$ and $\theta_{T_2}$ be the two roots of $s_n(\theta)=0$, with tangent points $\mathbf T_k=\mathbf q(\theta_{T_k})$, $k\in\{1,2\}$.
The associated directed spatial tangent rays have angles $\alpha$ and $\beta$, respectively, as illustrated in Fig.~\ref{fig:approaching_geometry_certificate}.
We label them such that $\theta_{T_1}<\theta_{T_2}$.
The ray from $\mathbf{p}_0$ along $\mathbf{v}_0$ intersects $\mathcal{S}_{vis}$ at a unique point $\mathbf{I}_v$ with parameter $\theta_v \in (\theta_{T_1}, \theta_{T_2})$.
The velocity-intersection parameter partitions the validated boundary curve parameterized by $\Theta_{\mathrm{app}}$ into two connected passing branches:
\begin{equation}
\Theta_L = (\theta_{T_1}, \theta_v],
\qquad
\Theta_R = (\theta_v, \theta_{T_2}).
\end{equation}
The branch $\Theta_L$ corresponds to the left-passing avoidance mode, whereas $\Theta_R$ corresponds to the right-passing avoidance mode.
The corresponding acceleration-space boundary branches are
\begin{equation}
\Gamma_j=\{\mathbf a(\theta):\theta\in\Theta_j\},\qquad j\in\{L,R\}.
\end{equation}

For $j\in\{L,R\}$ and $\theta\in\Theta_j\subset\Theta_{\mathrm{app}}$, write $\mathbf a=a_n\mathbf n+a_t\mathbf t$ in the local frame at $\mathbf q=R\mathbf n$. The dependence on $\theta$ is suppressed below.
With $\mathbf p_0$ and $\mathbf v_0$ fixed, $\mathrm d\mathbf n/\mathrm d\theta=\mathbf t$ and $\mathrm d\mathbf t/\mathrm d\theta=-\mathbf n$ give $\mathrm d s_n/\mathrm d\theta=s_t$, $\mathrm d s_t/\mathrm d\theta=R-s_n$, $\mathrm d v_n/\mathrm d\theta=v_t$, and $\mathrm d v_t/\mathrm d\theta=-v_n$.
The coefficients in~\eqref{eq:cc_boundary} then satisfy
\begin{equation}
\frac{\mathrm d a_n}{\mathrm d\theta}=-\frac{v_t v_n}{s_n}+\frac{s_t v_n^2}{2s_n^2}=a_t.
\end{equation}
Thus, the normal component cancels when differentiating in the rotating frame:
\begin{equation}
\label{eq:oriented_tangent}
\begin{aligned}
\frac{\mathrm d\mathbf a}{\mathrm d\theta}
&=\left(\frac{\mathrm d a_n}{\mathrm d\theta}-a_t\right)\mathbf n+\left(a_n+\frac{\mathrm d a_t}{\mathrm d\theta}\right)\mathbf t\\
&=\lambda(\theta)\mathbf t(\theta),
\end{aligned}
\end{equation}
where
\begin{equation}
\label{eq:positive_lambda}
\begin{aligned}
\lambda(\theta)&=a_n+\frac{\mathrm d a_t}{\mathrm d\theta}\\
&=\frac{R s_n v_n^2-2(s_n v_t-s_t v_n)^2}{2s_n^3}.
\end{aligned}
\end{equation}
Since $R>0$, $s_n<0$, and $v_n<0$, both numerator and denominator are strictly negative, giving $\lambda(\theta)>0$.
Let $\phi(\theta)=\arg(\mathrm d\mathbf a/\mathrm d\theta)$ be the continuous tangent angle of $\Gamma_j$ with increasing $\theta$.
Substituting~\eqref{eq:oriented_tangent} and using $\lambda(\theta)>0$ gives
\begin{equation}
\label{eq:tangent_angle}
\phi(\theta)=\arg\!\bigl(\lambda(\theta)\mathbf t(\theta)\bigr)=\theta+\frac{\pi}{2}.
\end{equation}
Together with $\lambda(\theta)>0$, this tangent-angle relation leads to the local convexity established in Theorem~\ref{thm:approaching_local_convexity}.

\begin{theorem}[Local Convexity]
\label{thm:approaching_local_convexity}
Each connected validated approaching branch $\Gamma_j$, $j\in\{L,R\}$, has $\kappa(\theta)>0$ for all $\theta\in\Theta_j$ and is locally strictly convex.
\end{theorem}

\begin{IEEEproof}
Let $\ell$ denote arc length along the branch in the direction of increasing $\theta$. Since $\|\mathbf t(\theta)\|=1$,~\eqref{eq:oriented_tangent} gives
\begin{equation}
\frac{\mathrm d\ell}{\mathrm d\theta}
=\left\|\frac{\mathrm d\mathbf a}{\mathrm d\theta}\right\|
=|\lambda(\theta)|=\lambda(\theta)>0,
\end{equation}
where $\lambda(\theta)>0$ on the validated approaching branch. Together with $\mathrm d\phi/\mathrm d\theta=1$ from~\eqref{eq:tangent_angle}, the chain rule gives
\begin{equation}
\label{eq:approaching_curvature}
\kappa(\theta)=\frac{\mathrm d\phi}{\mathrm d\ell}
=\frac{\mathrm d\phi/\mathrm d\theta}{\mathrm d\ell/\mathrm d\theta}
=\frac{1}{\lambda(\theta)}>0.
\end{equation}
Since $\kappa(\theta)>0$, each validated approaching branch is locally strictly convex.
\end{IEEEproof}

\subsubsection{Safety Certificate}
\label{subsec:approaching_certificate}

Theorem~\ref{thm:approaching_local_convexity} establishes local convexity of the approaching boundary branches. To construct an affine safety constraint from a boundary tangent, we must further show that the entire collision set lies on one side of the tangent line. Theorem~\ref{thm:approaching_tangent_certificate} establishes this global support property through trajectory projection.

\begin{theorem}[Tangent Support and Safety Certificate]
\label{thm:approaching_tangent_certificate}
Consider the disk C-space model of Section~\ref{subsec:collision_model}, with $d>R$ and constant relative acceleration. Choose a connected validated approaching branch $\Gamma_j$, $j\in\{L,R\}$, and an anchor parameter $\bar\theta\in\Theta_j$, with $\bar{\mathbf a}=\mathbf a(\bar\theta)$ and corresponding unit normal $\bar{\mathbf n}=\mathbf n(\bar\theta)$. The tangent line $\bar{\mathbf n}^{\top}(\mathbf a-\bar{\mathbf a})=0$ supports $\mathcal{CC}$ at $\bar{\mathbf a}$, with
\begin{equation}
\label{eq:approaching_collision_support}
\mathcal{CC}\subseteq\{\mathbf a:\bar{\mathbf n}^{\top}(\mathbf a-\bar{\mathbf a})\le0\}.
\end{equation}
Every acceleration in the half-plane
\begin{equation}\label{eq:approaching_halfplane}
H_{\mathrm{AO}}(\bar\theta)=\{\mathbf a:\bar{\mathbf n}^{\top}(\mathbf a-\bar{\mathbf a})>0\}
\end{equation}
generates a collision-free trajectory, with $\|\mathbf p(t;\mathbf a)\|>R$ for all $t\ge0$. Fig.~\ref{fig:approaching_geometry_certificate} illustrates this structure.
\end{theorem}

\begin{IEEEproof}
Equation~\eqref{eq:oriented_tangent} and Theorem~\ref{thm:approaching_local_convexity} establish tangency and local support at $\bar{\mathbf a}$.
At the anchor, let $t_c=t_c(\bar\theta)>0$ and $v_n=\bar{\mathbf n}^{\top}\mathbf v_0<0$. The contact and tangency relations give
\begin{equation}\label{eq:grazing_support_square}
\bar{\mathbf n}^{\top}\mathbf p(t;\bar{\mathbf a})
=R+\frac{-v_n}{2t_c}(t-t_c)^2\ge R.
\end{equation}
For any $\mathbf a\in H_{\mathrm{AO}}(\bar\theta)$, the acceleration difference has positive projection onto $\bar{\mathbf n}$. The relative dynamics therefore yield
\begin{equation}\label{eq:approaching_projection_safety}
\begin{aligned}
\bar{\mathbf n}^{\top}\mathbf p(t;\mathbf a)
&=\bar{\mathbf n}^{\top}\mathbf p(t;\bar{\mathbf a})
+\tfrac12t^2\bar{\mathbf n}^{\top}(\mathbf a-\bar{\mathbf a})\\
&>R,\qquad t>0.
\end{aligned}
\end{equation}
Since $\bar{\mathbf n}$ is a unit vector,
\begin{equation}\label{eq:approaching_nonpenetration}
\|\mathbf p(t;\mathbf a)\|\ge\bar{\mathbf n}^{\top}\mathbf p(t;\mathbf a)>R,
\qquad t>0.
\end{equation}
Together with $d>R$ at $t=0$, this proves that every acceleration in $H_{\mathrm{AO}}(\bar\theta)$ avoids contact.

Thus the half-plane~\eqref{eq:approaching_halfplane} contains no colliding acceleration, placing $\mathcal{CC}$ on the nonpositive side as in~\eqref{eq:approaching_collision_support}. Contact at $t_c$ gives $\bar{\mathbf a}\in\mathcal{CC}$, so the tangent line meets the collision set at the anchor and supports it globally.
\end{IEEEproof}

The tangent with normal $\bar{\mathbf n}$ globally supports the collision set at the smooth boundary point $\bar{\mathbf a}$, yielding both translational and angular tightness.
Any parallel shift that enlarges the certified half-plane, or any rotation about $\bar{\mathbf a}$ to a non-tangent orientation, admits colliding accelerations.

Substituting $\mathbf a=\mathbf u_{\mathcal R}-\mathbf a_{\mathcal O}$ gives the robot-input certificate
\begin{equation}\label{eq:ao_tangent_input}
\bar{\mathbf n}^{\top}\big[\mathbf u_{\mathcal R}-(\bar{\mathbf a}+\mathbf a_{\mathcal O})\big]>0.
\end{equation}

\subsection{Bypassing Configurations}
\label{subsec:bypassing_geometry}

\begin{figure}[htbp]
\centering
\includegraphics[width=\linewidth]{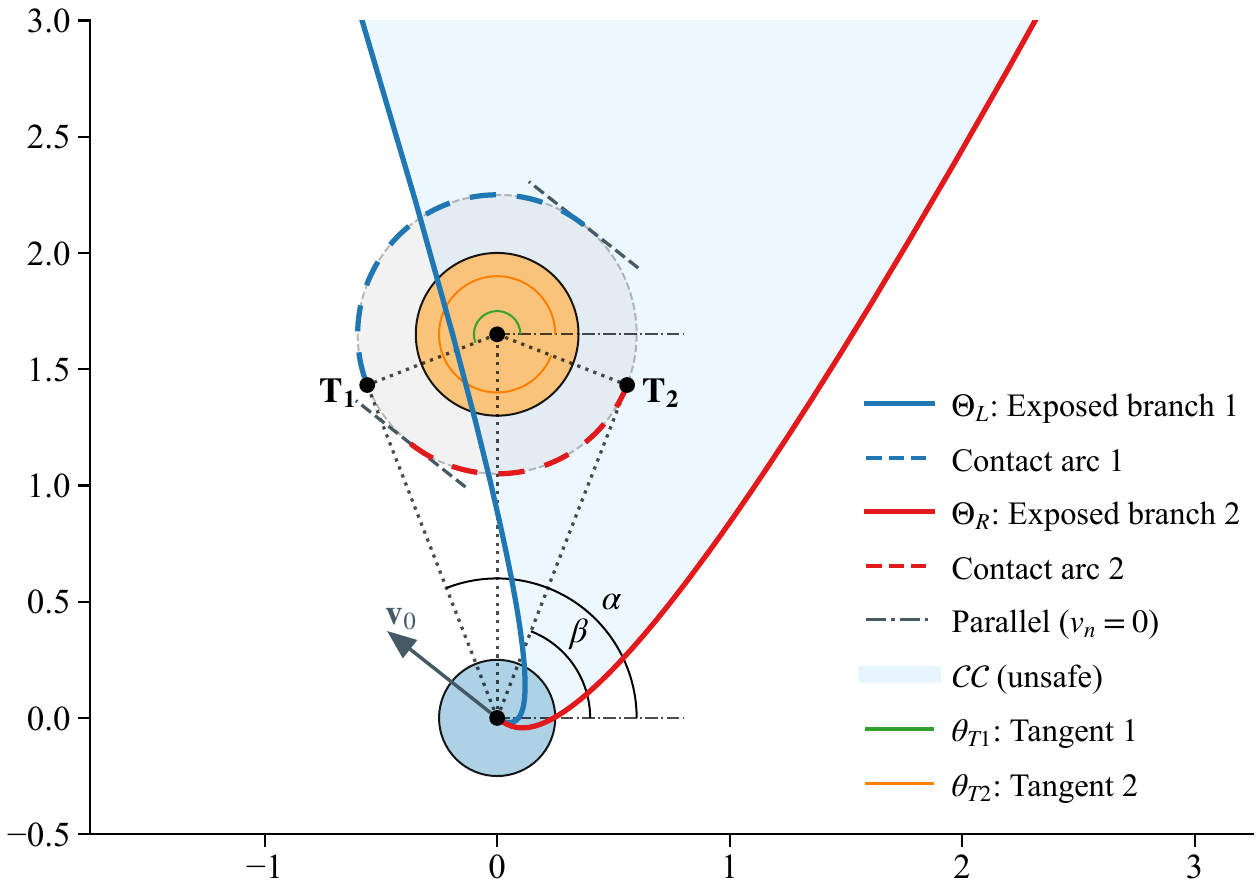}
\caption{Representative bypassing configuration with exposed grazing candidates. The dashed spatial contact arcs map to the solid curves bounding the shaded collision set in relative-acceleration space. 
% The spatial configuration is superimposed to show this correspondence.
}
\label{fig:bypassing_geometry}
\end{figure}

Fig.~\ref{fig:bypassing_geometry} illustrates a representative bypassing configuration. The two dashed contact arcs map to the solid acceleration-space curves bounding the shaded collision region.

\begin{figure}[t]
\centering
% Crop embedded panel headings; LaTeX supplies the subfigure labels.
\subfloat[]{%
\includegraphics[width=.48\columnwidth,trim=3bp 25bp 130bp 14bp,clip]{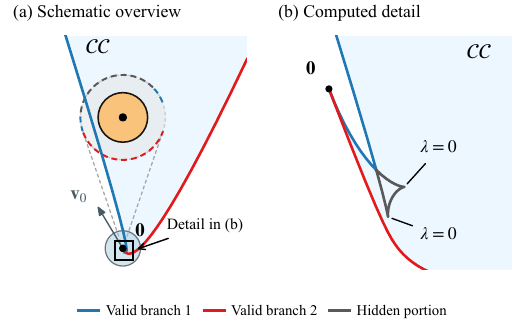}\label{fig:hidden_structure_overview}}\hfill
\subfloat[]{%
\includegraphics[width=.48\columnwidth,trim=130bp 25bp 3bp 14bp,clip]{figs/method/bypassing_hidden_structure.pdf}\label{fig:hidden_structure_detail}}
\par
\includegraphics[width=\columnwidth,trim=0bp 2bp 0bp 143bp,clip]{figs/method/bypassing_hidden_structure.pdf}
\captionsetup{skip=2pt}
\caption{Hidden penetrating candidates in bypassing motion. (a) Schematic overview of the spatial geometry and acceleration-space curves. 
(b) Detail for box region in (a). 
Blue and red portions in (b) have $Q_{\min}\ge0$, whereas the gray portion has $Q_{\min}<0$ and lies inside $\mathcal{CC}$. The labels $\lambda=0$ indicate stationary points of the candidate curve. 
Shading region denotes $\mathcal{CC}$.}
\label{fig:hidden_structure}
\end{figure}

The grazing construction can also produce hidden penetrating candidates, as shown in Fig.~\ref{fig:hidden_structure}. Such candidates lie in the interior of $\mathcal{CC}$ and therefore do not represent its physical boundary (Lemma~\ref{thm:grazing_validity}). Although the collision set's supporting geometry can be analyzed directly in acceleration space, a more intuitive derivation follows from the spatial geometry.

\begin{figure}[!htbp]
\centering
% Crop embedded headings and the bottom mu definition without retaining its space.
\subfloat[]{%
\includegraphics[width=.48\columnwidth,trim=3bp 17bp 128bp 14bp,clip]{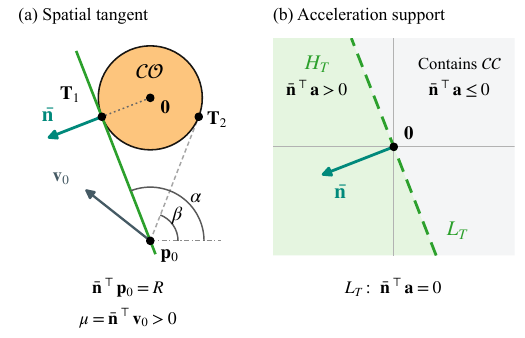}\label{fig:byp_regime_spatial}}\hfill
\subfloat[]{%
\includegraphics[width=.48\columnwidth,trim=128bp 17bp 3bp 14bp,clip]{figs/method/bypassing_support.pdf}\label{fig:byp_regime_acceleration}}
\caption{Correspondence between spatial tangency and acceleration-space support in bypassing motion. (a) The selected spatial tangent has outward normal $\bar{\mathbf n}$ satisfying $\bar{\mathbf n}^{\top}\mathbf v_0>0$. (b) The same normal defines the global support line $L_T$ through the acceleration origin, separating $\mathcal{CC}$ from the certified half-plane $H_T$.}
\label{fig:byp_regime}
\end{figure}

We seek a global support line for $\mathcal{CC}$. Spatial tangents bound the possible collision positions, and the relative dynamics connect these bounds to the colliding accelerations.
The spatial tangents through $\mathbf p_0$ at the previously introduced points $\mathbf T_1$ and $\mathbf T_2$ are shown in Fig.~\ref{fig:byp_regime}\subref{fig:byp_regime_spatial}.
Their outward unit normals are denoted by $\mathbf n_1$ and $\mathbf n_2$.

For $d=\|\mathbf p_0\|>R$, set
\begin{equation}
\mathbf e=\frac{\mathbf p_0}{d},\qquad
\mathbf e^\perp=
\begin{bmatrix}
0&-1\\[1mm]
1&0
\end{bmatrix}\mathbf e,
\end{equation}
which gives the two tangent normals and contact points
\begin{equation}
\label{eq:tangent_supports}
\mathbf n_{1,2}
=\frac{R}{d}\mathbf e
\mp
\sqrt{1-\frac{R^2}{d^2}}\,\mathbf e^\perp,\qquad
\mathbf T_{1,2}=R\mathbf n_{1,2}.
\end{equation}
For $k\in\{1,2\}$, the identity $\mathbf n_k^\top\mathbf p_0=\mathbf n_k^\top\mathbf T_k=R$ shows that each line $\mathbf n_k^\top\mathbf p=R$ supports the disk and passes through $\mathbf p_0$. Among these two outward normals, we select $\bar{\mathbf n}$ using the sign of the planar cross product of $\mathbf p_0$ and $\mathbf v_0$:
\begin{equation}
\label{eq:byp_normal_selection}
\begin{aligned}
\sigma&=\operatorname{sgn}(p_{0x}v_{0y}-p_{0y}v_{0x}),\\
\bar{\mathbf n}&=
\begin{cases}
\mathbf n_1,&\sigma<0,\\
\mathbf n_2,&\sigma>0.
\end{cases}
\end{aligned}
\end{equation}
When $\sigma=0$, both normals have the same projection onto $\mathbf v_0$, so either may be selected. For every nonzero strict bypassing velocity, this selection gives
\begin{equation}
\label{eq:spatial_tangent_selection}
\mu:=\bar{\mathbf n}^\top\mathbf v_0>0 .
\end{equation}
The case $\mu=0$ is treated separately in Remark~\ref{rem:degenerate}.
Theorem~\ref{thm:byp_global_support} establishes the desired global support and the associated safety certificate for the selected normal $\bar{\mathbf n}$.

\begin{theorem}[Global Support and Safety Certificate]
\label{thm:byp_global_support}
In strict bypassing motion, the selected tangent normal $\bar{\mathbf n}$ satisfying~\eqref{eq:spatial_tangent_selection} yields
\begin{equation}
\label{eq:byp_global_support}
\begin{aligned}
\mathcal{CC}&\subseteq\{\mathbf a:\bar{\mathbf n}^\top\mathbf a\le0\},\\
\sup_{\mathbf a\in\mathcal{CC}}\bar{\mathbf n}^\top\mathbf a&=0.
\end{aligned}
\end{equation}
Hence, a global support line through the origin is
\begin{equation}
\label{eq:byp_support_line}
L_T=\{\mathbf a:\bar{\mathbf n}^\top\mathbf a=0\}.
\end{equation}
Every relative acceleration in the half-plane
\begin{equation}
\label{eq:support_halfplane}
H_T=\{\mathbf a:\bar{\mathbf n}^\top\mathbf a>0\}
\end{equation}
satisfies
\begin{equation}
\|\mathbf p_0+\mathbf v_0t+\tfrac12\mathbf a t^2\|>R,
\qquad \forall t\ge0 .
\end{equation}
Thus the trajectory is collision-free.
\end{theorem}
\begin{IEEEproof}
For any $\mathbf a\in\mathcal{CC}$, there is a collision time $t_{\mathrm{col}}>0$ such that $\|\mathbf p(t_{\mathrm{col}};\mathbf a)\|\le R$.
Rearranging gives $\mathbf a=2(\mathbf p(t_{\mathrm{col}};\mathbf a)-\mathbf p_0-t_{\mathrm{col}}\mathbf v_0)/t_{\mathrm{col}}^2$. Since $\bar{\mathbf n}$ is unit length, $\bar{\mathbf n}^\top\mathbf p(t_{\mathrm{col}};\mathbf a)\le R$, and thus
\begin{equation}
\label{eq:byp_collision_projection}
\begin{aligned}
\bar{\mathbf n}^\top\mathbf a
&=\frac{2}{t_{\mathrm{col}}^2}\bigl(\bar{\mathbf n}^\top\mathbf p(t_{\mathrm{col}};\mathbf a)-R-t_{\mathrm{col}}\mu\bigr)\\
&\le-\frac{2\mu}{t_{\mathrm{col}}}<0.
\end{aligned}
\end{equation}
Thus $\mathcal{CC}\subseteq\{\mathbf a:\bar{\mathbf n}^\top\mathbf a\le0\}$ for all colliding accelerations.

For any finite arrival time $t_{\mathrm{ctr}}>0$ at the disk center, consider the acceleration
\begin{equation}
\label{eq:byp_origin_family}
\mathbf b_{t_{\mathrm{ctr}}}
=-\frac{2(\mathbf p_0+t_{\mathrm{ctr}}\mathbf v_0)}{t_{\mathrm{ctr}}^2}.
\end{equation}
This acceleration reaches the obstacle center at $t_{\mathrm{ctr}}$, so $\mathbf b_{t_{\mathrm{ctr}}}\in\mathcal{CC}$. Each chosen time therefore defines a colliding acceleration with normal projection
\begin{equation}
\label{eq:byp_origin_projection}
\bar{\mathbf n}^\top\mathbf b_{t_{\mathrm{ctr}}}
=-\frac{2R}{t_{\mathrm{ctr}}^2}-\frac{2\mu}{t_{\mathrm{ctr}}}<0.
\end{equation}
Together with~\eqref{eq:byp_collision_projection}, this yields
\begin{equation}
\label{eq:byp_support_supremum}
0=\lim_{t_{\mathrm{ctr}}\to\infty}\bar{\mathbf n}^\top\mathbf b_{t_{\mathrm{ctr}}}
\le\sup_{\mathbf a\in\mathcal{CC}}\bar{\mathbf n}^\top\mathbf a
\le0.
\end{equation}
The half-plane inclusion and zero supremum establish $L_T$ as a global support line through the origin.

To verify the safety certificate, let $\mathbf a\in H_T$. Then $\mu>0$ and $\bar{\mathbf n}^\top\mathbf a>0$ give
\begin{equation}
\label{eq:byp_projection}
\begin{aligned}
h(t):=\bar{\mathbf n}^\top\mathbf p(t)
&=R+\mu t+\tfrac12\bar{\mathbf n}^\top\mathbf a\,t^2\\
&>R,\qquad t>0.
\end{aligned}
\end{equation}
Since $\bar{\mathbf n}$ is unit length, $\|\mathbf p(t)\|\ge h(t)>R$ for $t>0$. Initial separation follows from $d>R$.
\end{IEEEproof}

The global support and safety certificate follow directly from the relative dynamics, without assuming convexity of $\mathcal{CC}$ or smoothness of its boundary.

\begin{remark}[Tangent-velocity and zero-velocity limits]
\label{rem:degenerate}
The same safety constraint~\eqref{eq:support_halfplane} applies to the nonzero tangent-ray and zero-velocity cases identified in Section~\ref{subsec:grazing_construction}. In both cases, the tangent normal can be chosen such that $\mu=0$, reducing the projection~\eqref{eq:byp_projection} to $h(t)=R+\tfrac12\bar{\mathbf n}^\top\mathbf a\,t^2$. With $\bar{\mathbf n}^\top\mathbf a>0$, $h(t)>R$ for every $t>0$; together with $d>R$, this establishes strict collision freedom.
\end{remark}

Substituting $\mathbf a=\mathbf u_{\mathcal R}-\mathbf a_{\mathcal O}$ gives the strict robot-input certificate
\begin{equation}
\label{eq:byp_support_input}
\bar{\mathbf n}^\top(\mathbf u_{\mathcal R}-\mathbf a_{\mathcal O})>0.
\end{equation}

A sufficient safety certificate may restrict the available accelerations more than necessary. To examine whether the present affine constraint can be relaxed without admitting colliding accelerations, Appendix~\ref{sec:support_tightness_details} analyzes its tightness under inward translation and unsafe-side rotation.

\section{Control Synthesis}
\label{sec:optimization}

This section combines the geometric certificates of Section~\ref{sec:geometry_certificates} with actuator limits to form a certified feasible set. We formulate a quadratic program for target tracking.

\subsection{Unified Affine Safety Constraints}

The robot-input certificates~\eqref{eq:ao_tangent_input} and~\eqref{eq:byp_support_input} share the same affine form
\begin{equation}\label{eq:unified_affine}
\bar{\mathbf n}_i^\top\mathbf u_{\mathcal R}>b_i,\qquad
b_i=\bar{\mathbf n}_i^\top(\bar{\mathbf a}_i+\mathbf a_{\mathcal O_i}),
\end{equation}
where $\bar{\mathbf n}_i$ is the unit normal pointing toward the safe half-plane and $\bar{\mathbf a}_i$ is the relative-acceleration anchor. 
Both are selected from the current relative state as follows.
Using the sign $\sigma$ defined in~\eqref{eq:byp_normal_selection}, the passing rule selects the right-passing tangent $\mathbf T_2$ for $\sigma>0$ and the left-passing tangent $\mathbf T_1$ for $\sigma\le0$.

\subsubsection{Approaching anchor selection}

For $\mathbf v_0\in\operatorname{int}\mathcal K$, the parameter $\gamma\in(0,1)$ controls the balance between lateral passing and braking. 
We interpolate between the tangent parameter $\theta_T\in\{\theta_{T_1},\theta_{T_2}\}$ selected by the passing rule and the velocity-intersection parameter $\theta_v$ defined in Section~\ref{subsec:approaching_geometry}:
\begin{equation}\label{eq:anchor_interpolation}
    \bar{\theta} = (1 - \gamma)\theta_T + \gamma\theta_v.
\end{equation}
The selected parameter $\bar\theta\in\Theta_{\mathrm{app}}$ defines the boundary anchor $\bar{\mathbf a}=\mathbf a(\bar\theta)$ and the unit normal $\bar{\mathbf n}=\mathbf n(\bar\theta)$.
The resulting affine constraint satisfies the safety condition in~\eqref{eq:ao_tangent_input}.

\subsubsection{Bypassing support selection}

Set $\bar{\mathbf a}=\mathbf0$ and choose $\bar{\mathbf n}=\mathbf n_2$ for $\sigma>0$ and $\bar{\mathbf n}=\mathbf n_1$ for $\sigma\le0$.
The resulting affine constraint is the spatial-support certificate in~\eqref{eq:byp_support_input}, with $\bar{\mathbf n}^\top\mathbf v_0>0$ for strict bypassing motion. 
Remark~\ref{rem:degenerate} extends the certificate to tangent-velocity and zero-velocity cases, where $\bar{\mathbf n}^\top\mathbf v_0=0$.

\subsection{Certified Feasible Set}

\begin{figure*}[t]
\centering
% Equal-width crops retain the axes; LaTeX supplies subfigure labels and titles.
\subfloat[Robot and moving obstacles]{%
\includegraphics[width=.32\textwidth,trim=-3.5bp 15bp 352.176bp 11bp,clip]{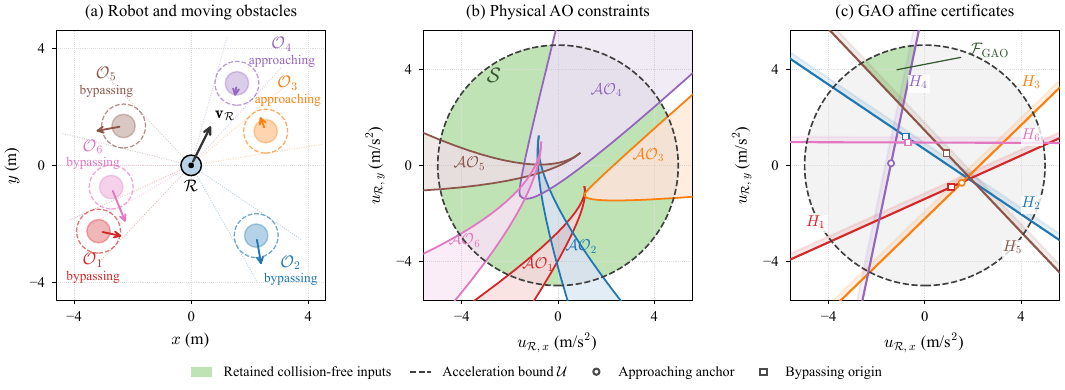}\label{fig:ao_gao_world}}\hfill
\subfloat[Physical AO constraints]{%
\includegraphics[width=.32\textwidth,trim=172.588bp 15bp 176.088bp 11bp,clip]{figs/method/ao_gao_comparison.pdf}\label{fig:ao_gao_physical}}\hfill
\subfloat[GAO affine certificates]{%
\includegraphics[width=.32\textwidth,trim=348.676bp 15bp 0bp 11bp,clip]{figs/method/ao_gao_comparison.pdf}\label{fig:ao_gao_affine}}
\par
\includegraphics[width=\textwidth,trim=0bp 2bp 0bp 173bp,clip]{figs/method/ao_gao_comparison.pdf}
\captionsetup{skip=2pt}
\caption{Construction of the GAO certified input set. (a) Instantaneous configuration of the robot and moving obstacles. Arrows indicate velocities, and dashed circles denote configuration-space obstacle disks. (b) Physical acceleration obstacles $\mathcal{AO}_i$ (colored) and the collision-free set $\mathcal S$ (green). (c) GAO constructs open half-planes $H_i$, whose intersection with the admissible input set $\mathcal U$ defines the certified set $\mathcal F_{\mathrm{GAO}}$ (green). Colored bands indicate the safe sides of the supporting lines. The lines themselves are excluded from $H_i$. Circular and square markers denote approaching anchors and bypassing origins, respectively, each translated by the corresponding obstacle acceleration. 
(b) and (c) share the same obstacle radii and accelerations, and the dashed circle denotes the actuator bound.}
\label{fig:ao_gao_comparison}
\end{figure*}

For each obstacle, define the collision-free half-plane $H_i=\{\mathbf u_{\mathcal R}:\bar{\mathbf n}_i^\top\mathbf u_{\mathcal R}>b_i\}$ from~\eqref{eq:unified_affine}. 
Let $\mathcal S=\mathcal U\setminus\bigcup_i\mathcal{AO}_i$ denote the actuator-feasible inputs whose future trajectories avoid contact with every obstacle for $t\ge0$. 
The certified feasible region satisfies
\begin{equation}\label{eq:gao_feasible_inclusion}
\mathcal F_{\mathrm{GAO}}=\mathcal U\cap\bigcap_{i=1}^{m}H_i
\ \subseteq\ \mathcal S.
\end{equation}
The inclusion follows pairwise from Theorem~\ref{thm:approaching_tangent_certificate}, Theorem~\ref{thm:byp_global_support}, and Remark~\ref{rem:degenerate}. With convex actuator limits, $\mathcal F_{\mathrm{GAO}}$ is convex, although $\mathcal S$ need not be. Fig.~\ref{fig:ao_gao_comparison} illustrates this geometric inclusion for six obstacles.

For QP implementation, we approximate the actuator disk $\|\mathbf u_{\mathcal R}\|\le u_{\max}$ by an inscribed polygon~\cite{alonsomora2018ccnr}:
\begin{equation}
\mathcal U_{\mathrm{qp}}=\{\mathbf u:A_u\mathbf u\le\mathbf c_u\}
\subseteq\mathcal U.
\end{equation}

\subsection{Nominal Tracking and Quadratic Optimization}

We compute the nominal proportional-derivative (PD) command
\begin{equation}
\mathbf u_{\mathrm{nom}}
=-k_p(\mathbf p_{\mathcal R}-\mathbf p_{\mathrm{target}})
-2\sqrt{k_p}\,\mathbf v_{\mathcal R},\qquad k_p>0,
\end{equation}
saturate it to the actuator limits, and obtain the control input by solving the following QP :
\begin{equation}\label{eq:qp_formulation}
\begin{aligned}
\min_{\mathbf u_{\mathcal R}}\quad &
\tfrac12\|\mathbf u_{\mathcal R}-\mathbf u_{\mathrm{nom}}\|^2\\
\text{s.t.}\quad &\bar{\mathbf n}_i^\top\mathbf u_{\mathcal R} > b_i,
\quad i=1,\ldots,m,\\
&A_u\mathbf u_{\mathcal R}\le\mathbf c_u.
\end{aligned}
\end{equation}
The QP in~\eqref{eq:qp_formulation} may be infeasible if the selected affine constraints are mutually incompatible or conflict with the actuator limits. 
To handle such infeasibility, we introduce the softened QP
\begin{equation}
\label{eq:qp_softened}
\begin{aligned}
\min_{\mathbf u_{\mathcal R},\,\boldsymbol\delta}\quad &
\tfrac12\|\mathbf u_{\mathcal R}-\mathbf u_{\mathrm{nom}}\|^2
+\sum_{i=1}^m\rho_i\delta_i^2\\
\text{s.t.}\quad &\bar{\mathbf n}_i^\top\mathbf u_{\mathcal R}+\delta_i > b_i,
\quad i=1,\ldots,m,\\
&A_u\mathbf u_{\mathcal R}\le\mathbf c_u,\qquad
\boldsymbol\delta\ge\mathbf0.
\end{aligned}
\end{equation}
where $\rho_i \gg0$.

\begin{remark}[Feasibility]
\label{rem:soft_safety}
When~\eqref{eq:qp_formulation} is infeasible, the softened QP~\eqref{eq:qp_softened} provides a feasible fallback, with violations of the affine constraints penalized in the objective.
\end{remark}

At each sampling instant, the controller updates the constraints, computes the control input, and holds it until the next update. Collision freedom follows if each applied input satisfies~\eqref{eq:unified_affine} for all obstacles and the assumptions in Section~\ref{subsec:collision_model} hold throughout each control interval.

\section{Simulation Experiments}
\label{sec:sim_exp}
\subsection{Navigation Evaluation Protocol}
We evaluate GAO in three sets of simulation experiments.
Section~\ref{sec:sim_1} assesses single-robot avoidance among dense, noncooperative dynamic obstacles.
Section~\ref{sec:sim_2} compares reciprocal avoidance in multi-agent configurations.
Section~\ref{sec:sim_large_scale} evaluates reciprocal navigation with up to 1,000 agents, both with and without static obstacles.
The evaluation measures safety, navigation time, path length, control effort, computation latency, and reciprocal coordination across agent counts.

In Sections~\ref{sec:sim_1} and~\ref{sec:sim_2}, we compare GAO with six baselines.
ORCA \cite{orca} is a first-order geometric method based on velocity obstacles.
MPC-CBF \cite{xax_mpc} represents formal safety filtering and is implemented as a nonlinear programming (NLP) problem over a receding horizon.
DWA \cite{dwa} is a sampling-based planner.
AVO \cite{avo} extends the VO paradigm to explicitly incorporate acceleration constraints.
AVOCADO \cite{avocado} uses nonlinear opinion dynamics to infer surrounding agents' cooperation levels within an adaptive VO approach.
SICNav \cite{sicnav} uses bilevel model predictive control to jointly optimize trajectories and predict crowd motions.

Within each comparative benchmark, all planners use the same workspace, robot radius, obstacle radius, target condition, and control update rate.
For planners that output velocity commands, changes in the executed velocity are constrained by the maximum acceleration limit.
To ensure a consistent and fair comparison, all optimization problems are formulated in CasADi \cite{casadi} and solved using Ipopt \cite{ipopt}, although some methods can be solved more efficiently using alternative solvers.
All evaluations are conducted on a desktop workstation equipped with an Intel Core i9-13900K CPU.
The implementation settings and timing scope of the large-scale GAO experiments are specified separately in Section~\ref{sec:sim_large_scale}.

% Keep the single-robot trajectory figure and its result table on one page.
\begin{table*}[!tp]
\begingroup
\makeatletter\def\@captype{figure}\makeatother
\normalsize
    \centering
    \subfloat[ORCA]{\includegraphics[width=0.135\linewidth]{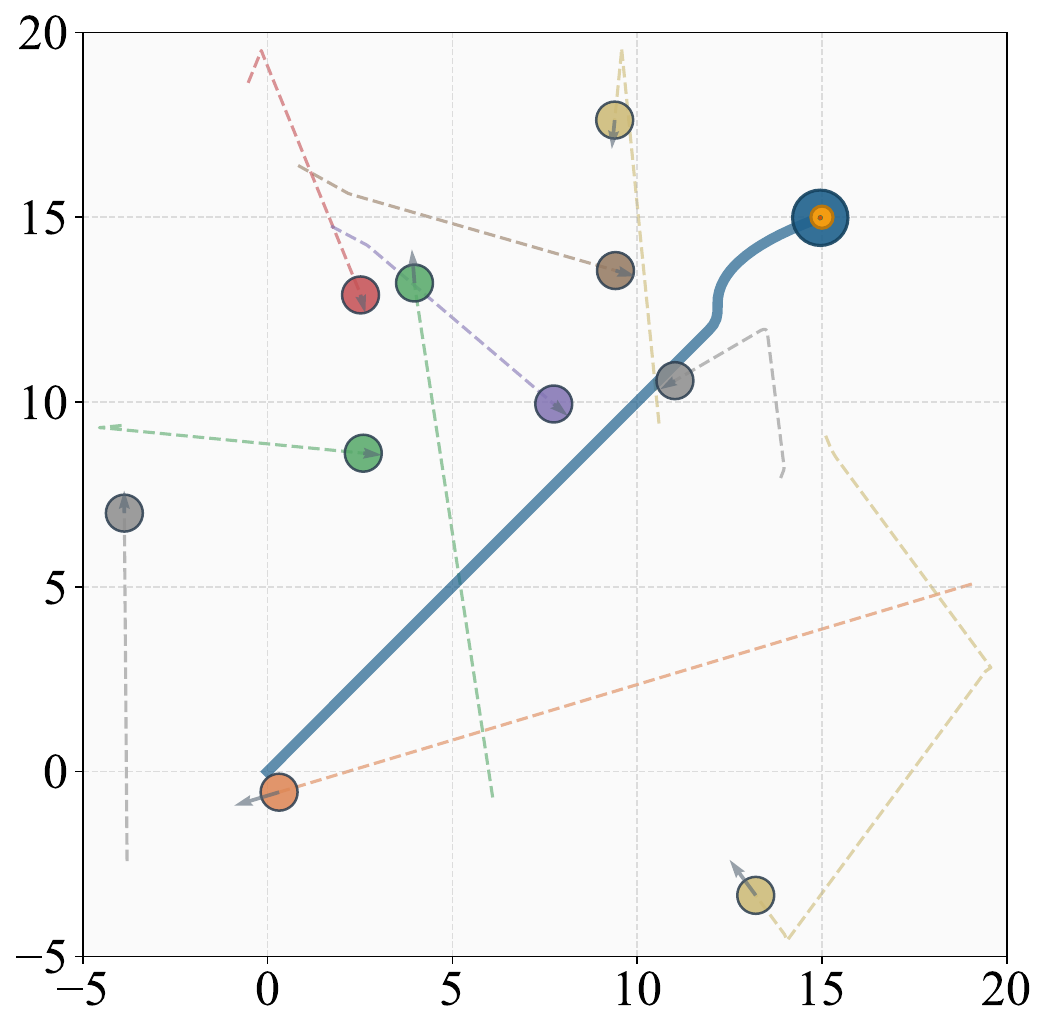}} \hfil
    \subfloat[DWA]{\includegraphics[width=0.135\linewidth]{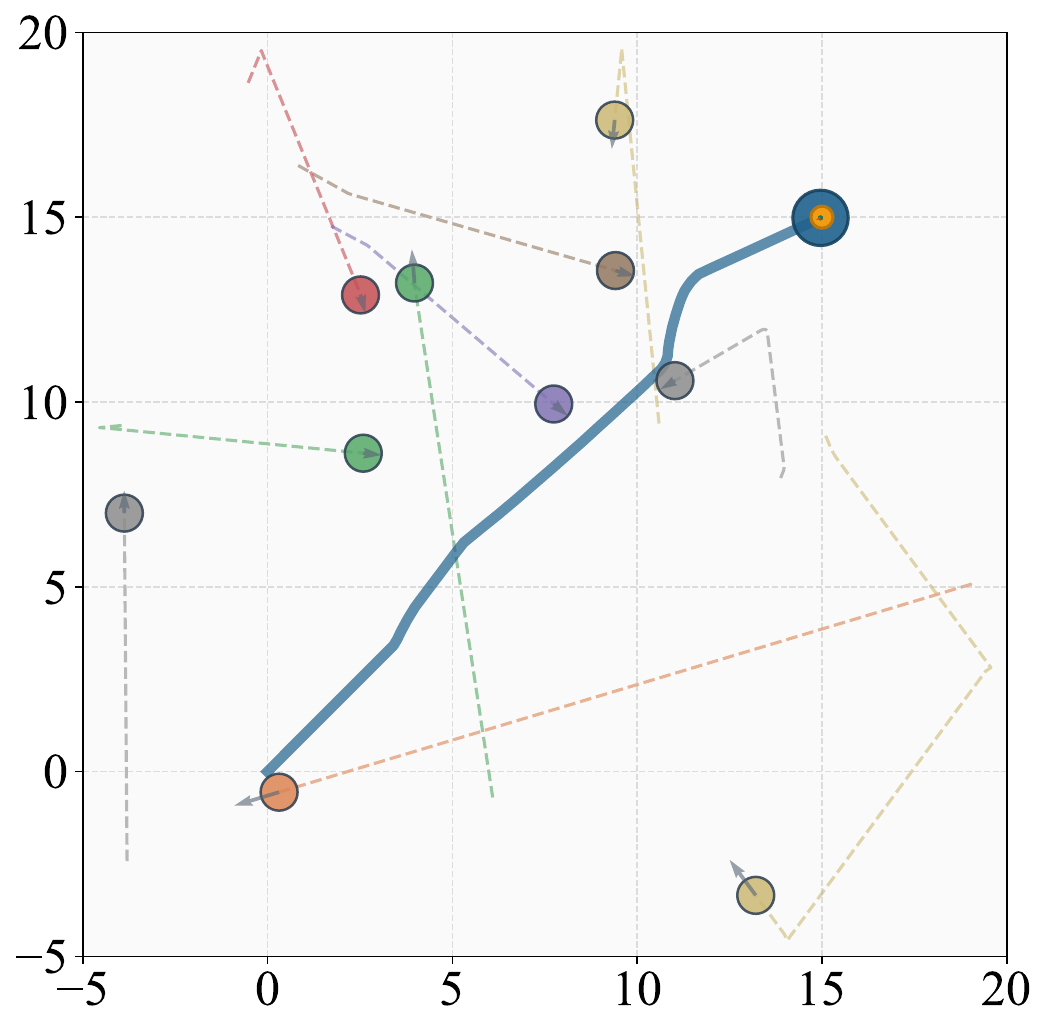}} \hfil
    \subfloat[MPC-CBF]{\includegraphics[width=0.135\linewidth]{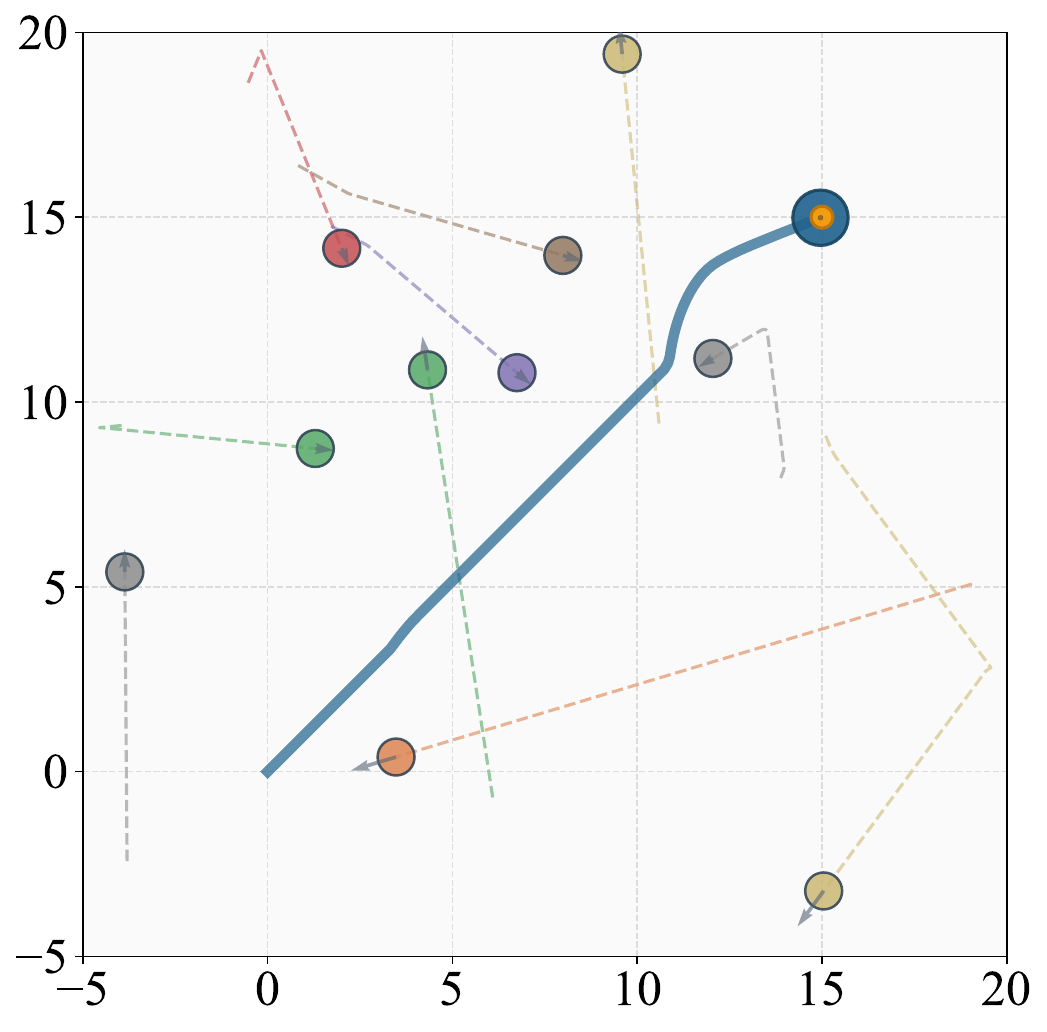}} \hfil
    \subfloat[SICNav]{\includegraphics[width=0.135\linewidth]{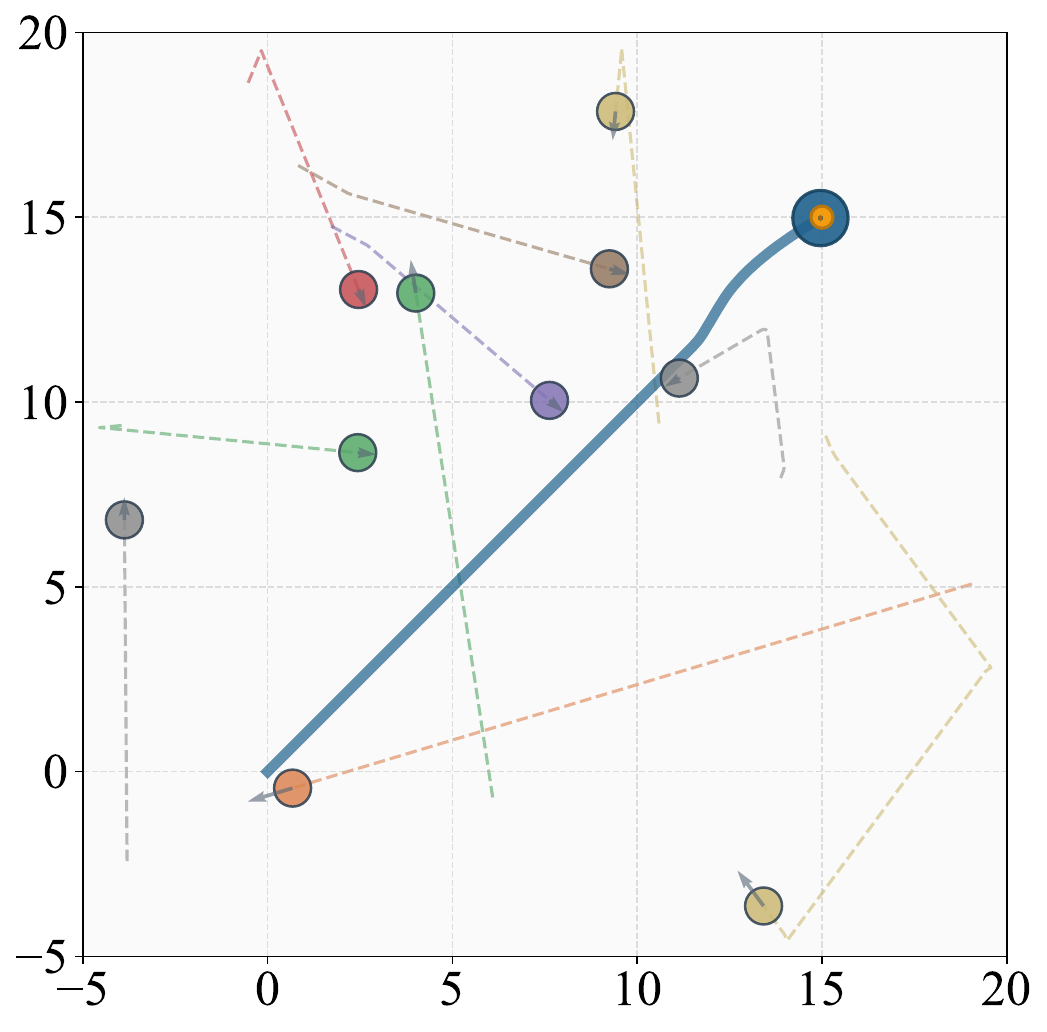}} \hfil
    \subfloat[AVO]{\includegraphics[width=0.135\linewidth]{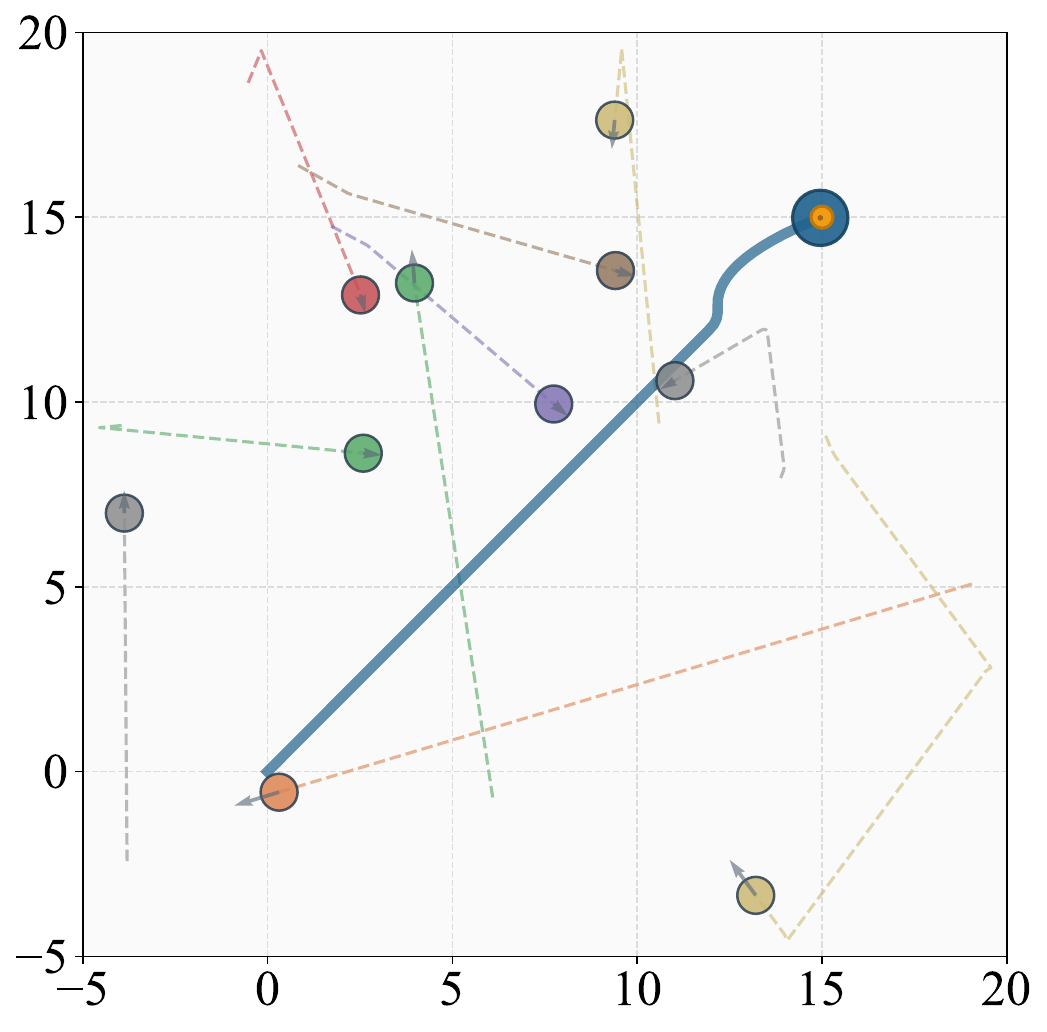}} \hfil
    \subfloat[AVOCADO]{\includegraphics[width=0.135\linewidth]{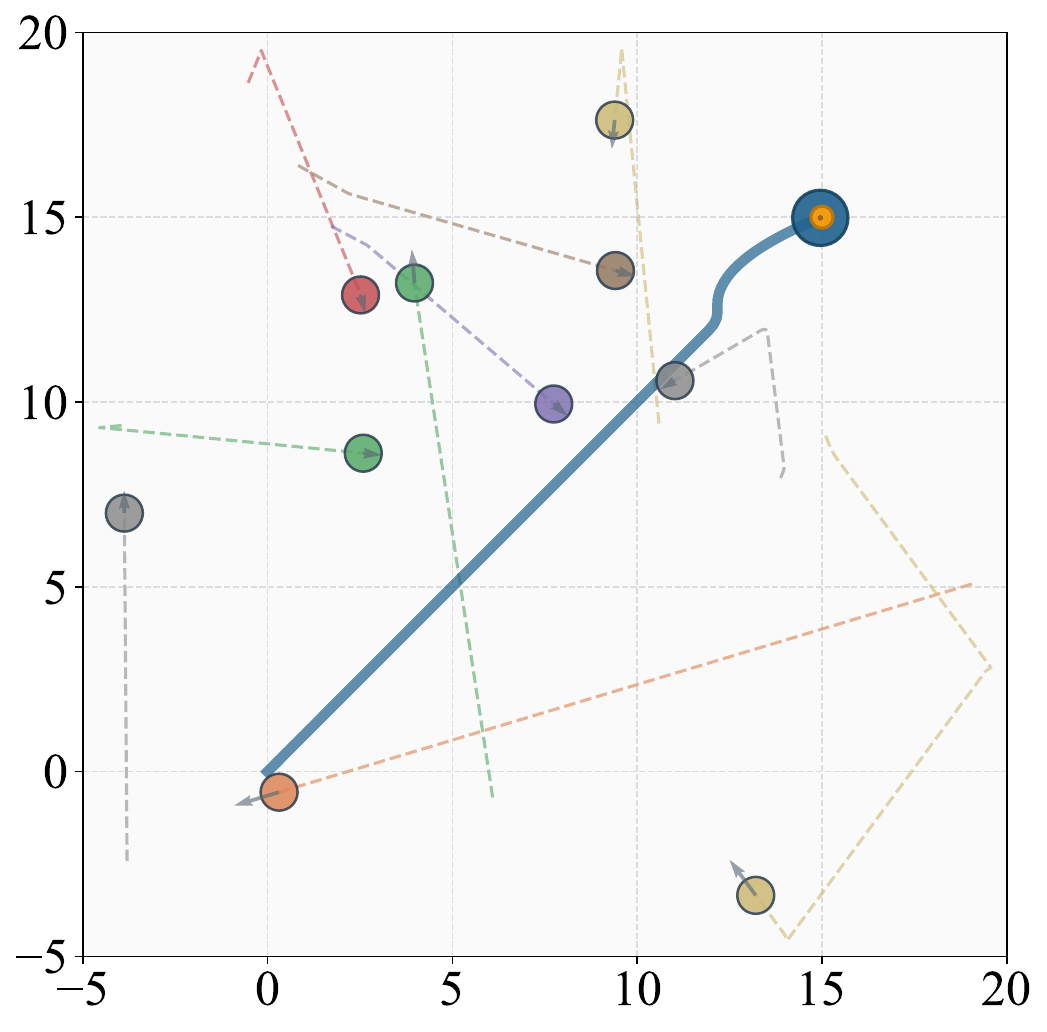}} \hfil
    \subfloat[GAO (Ours)]{\includegraphics[width=0.135\linewidth]{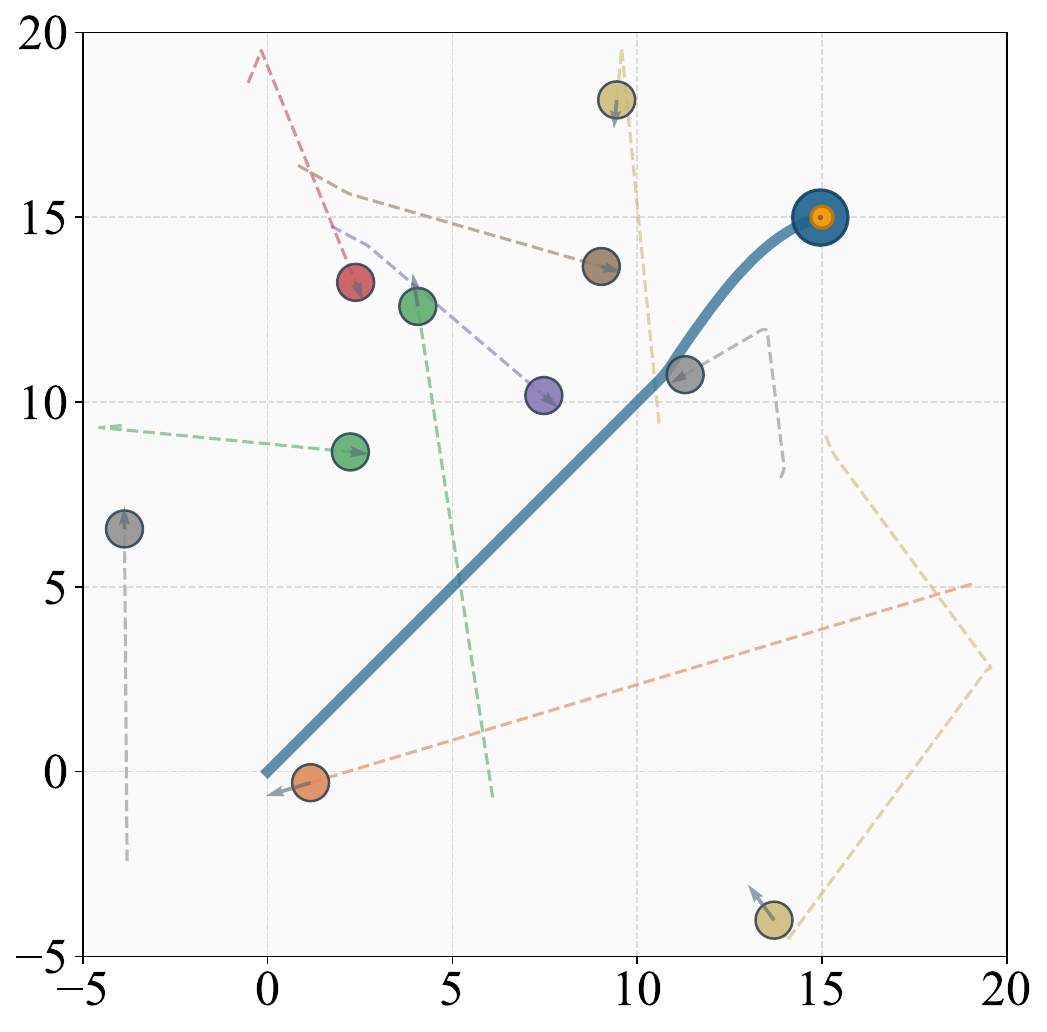}} \\
    
    \subfloat[ORCA]{\includegraphics[width=0.135\linewidth]{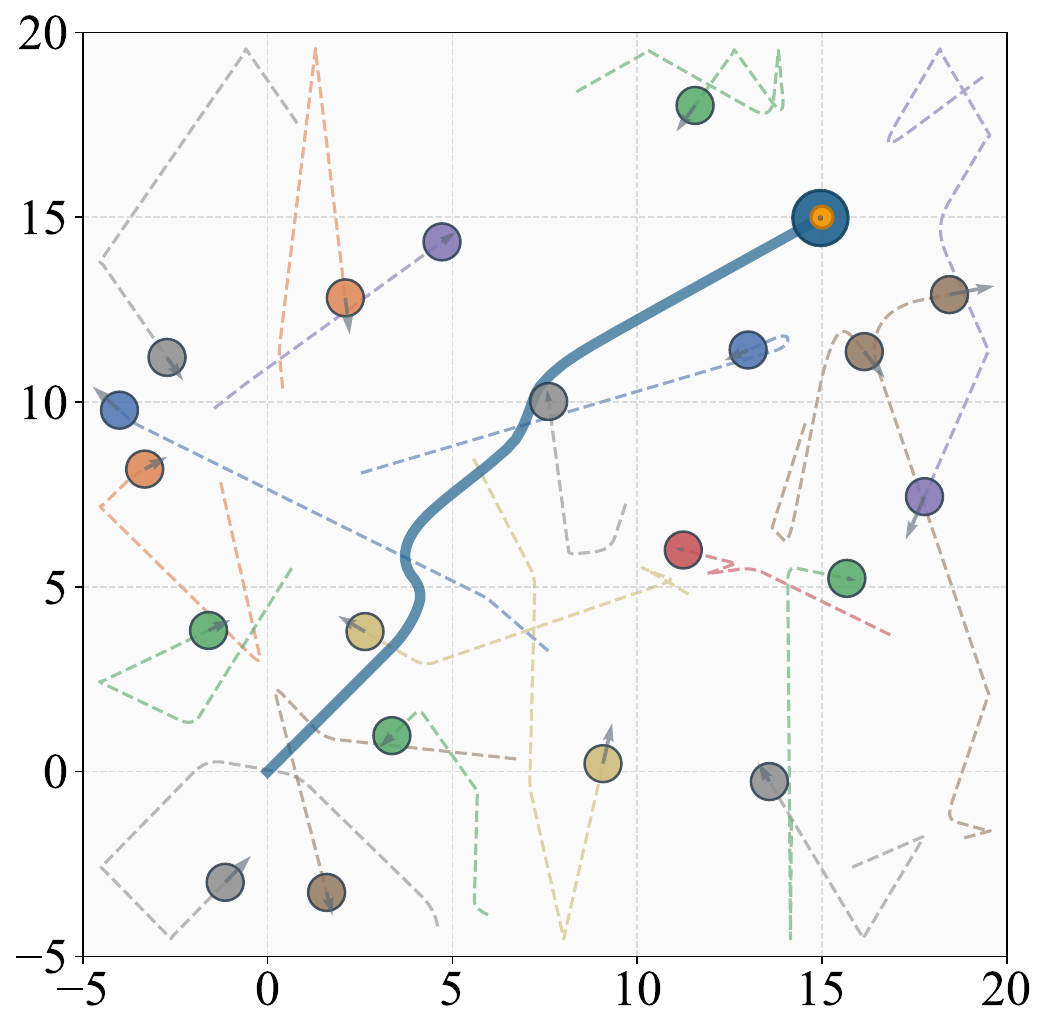}} \hfil
    \subfloat[DWA]{\includegraphics[width=0.135\linewidth]{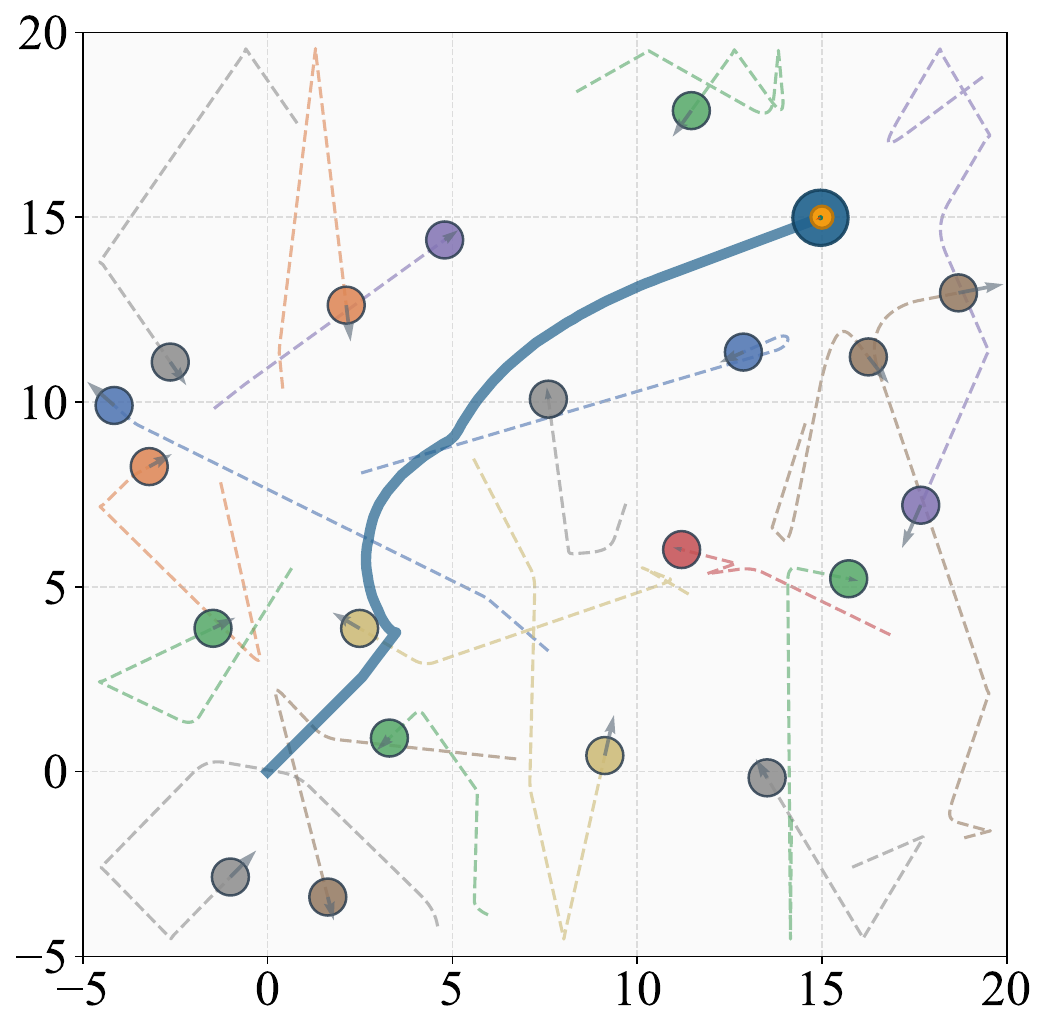}} \hfil
    \subfloat[MPC-CBF]{\includegraphics[width=0.135\linewidth]{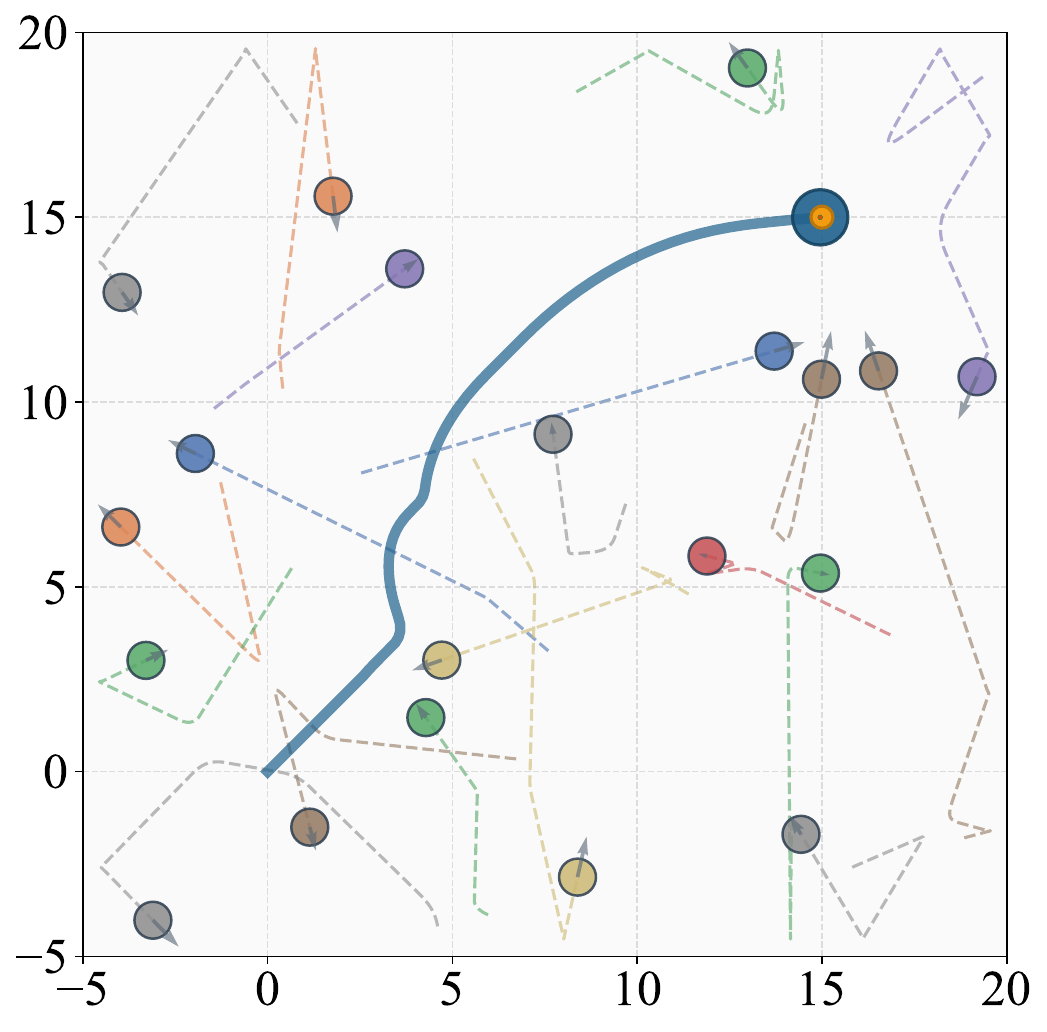}} \hfil
    \subfloat[SICNav]{\includegraphics[width=0.135\linewidth]{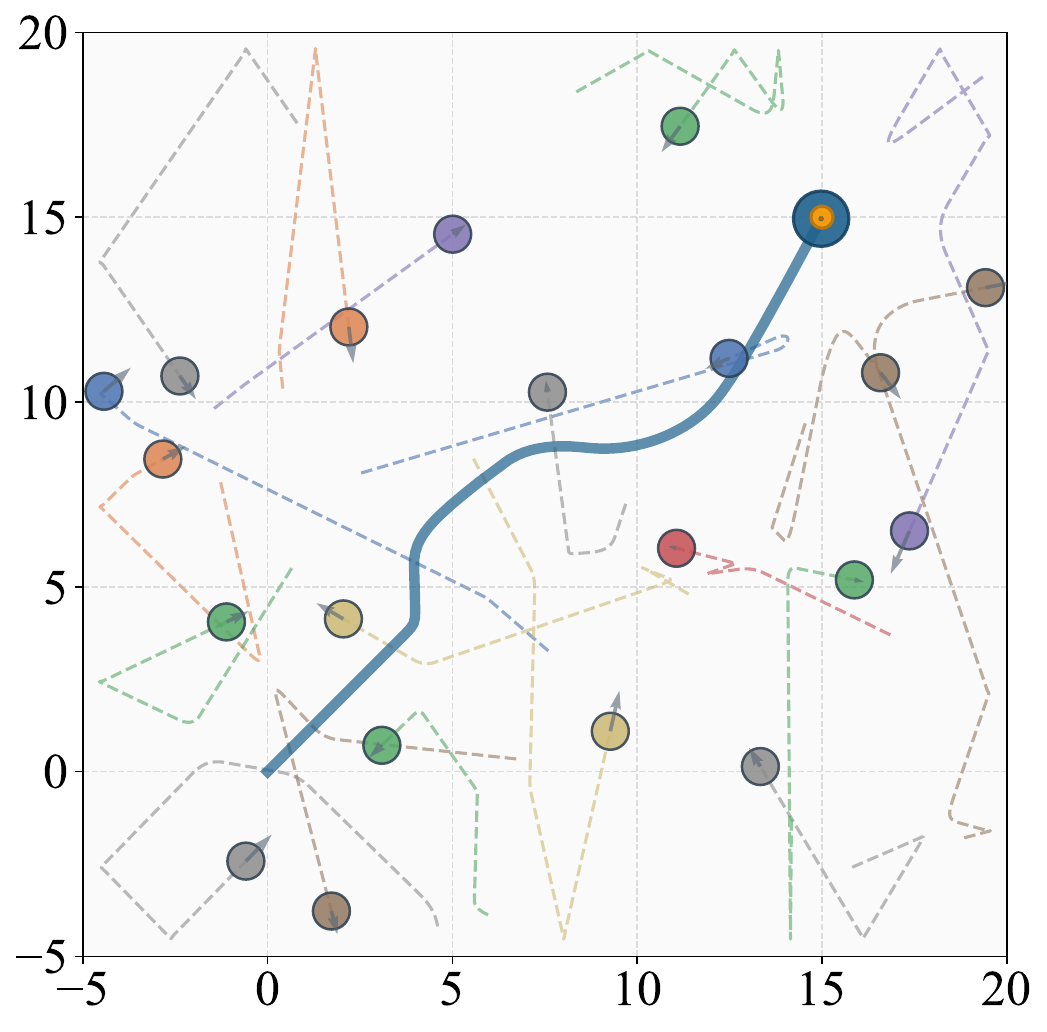}} \hfil
    \subfloat[AVO]{\includegraphics[width=0.135\linewidth]{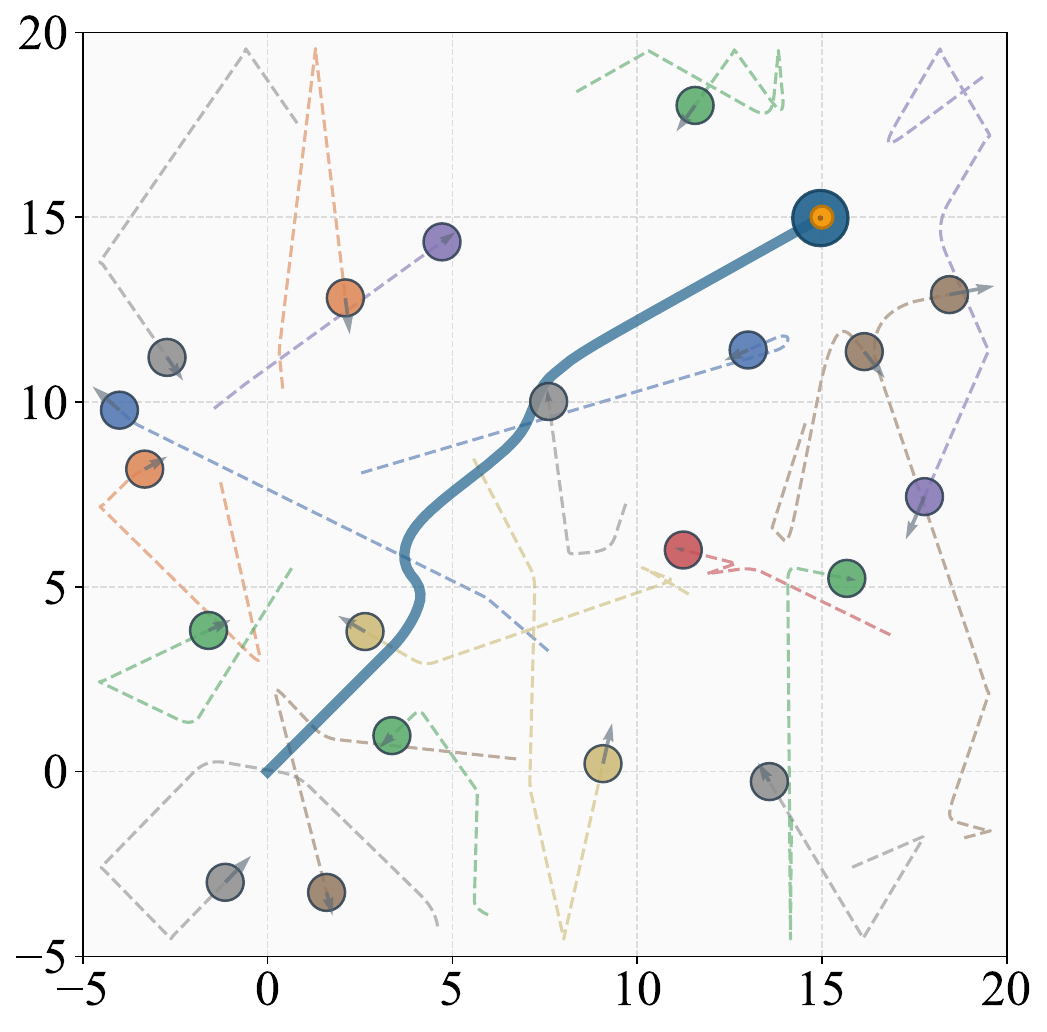}} \hfil
    \subfloat[AVOCADO]{\includegraphics[width=0.135\linewidth]{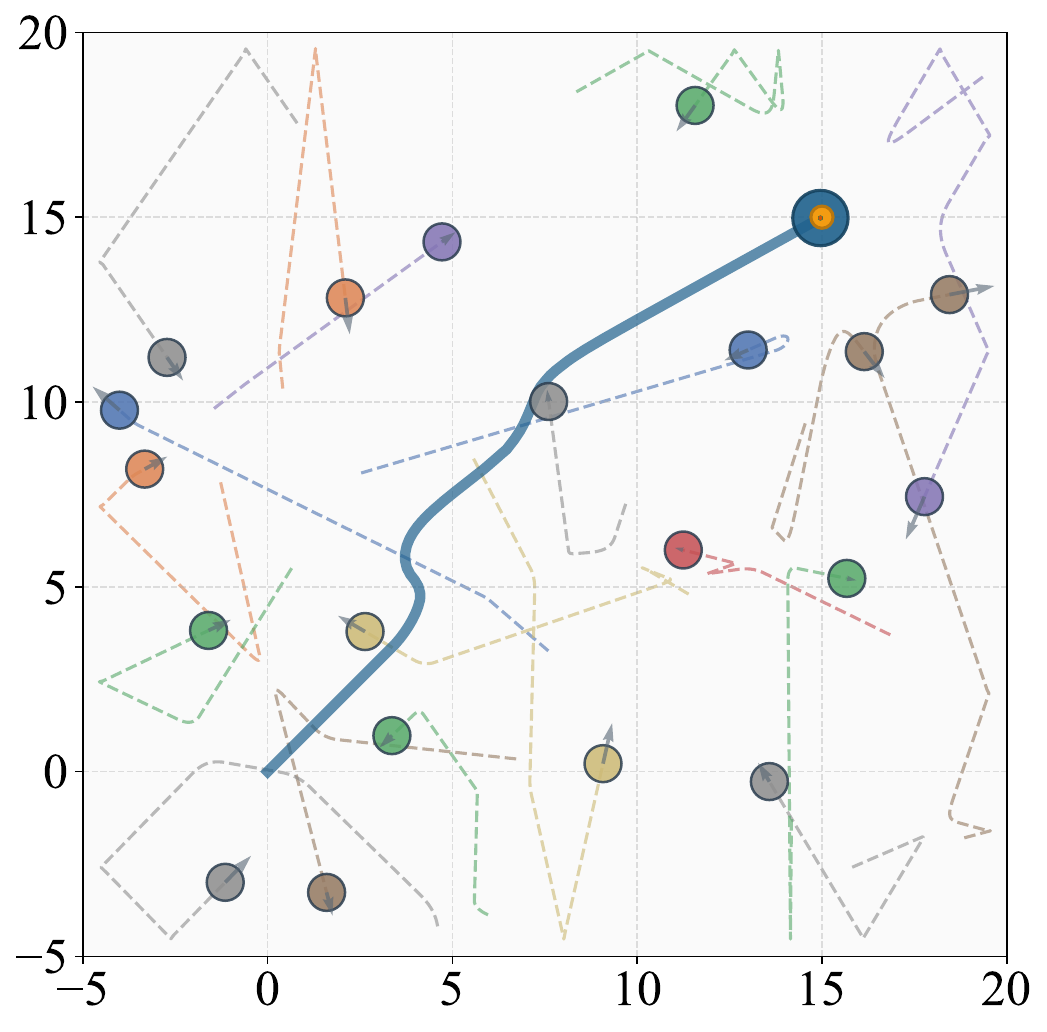}} \hfil
    \subfloat[GAO (Ours)]{\includegraphics[width=0.135\linewidth]{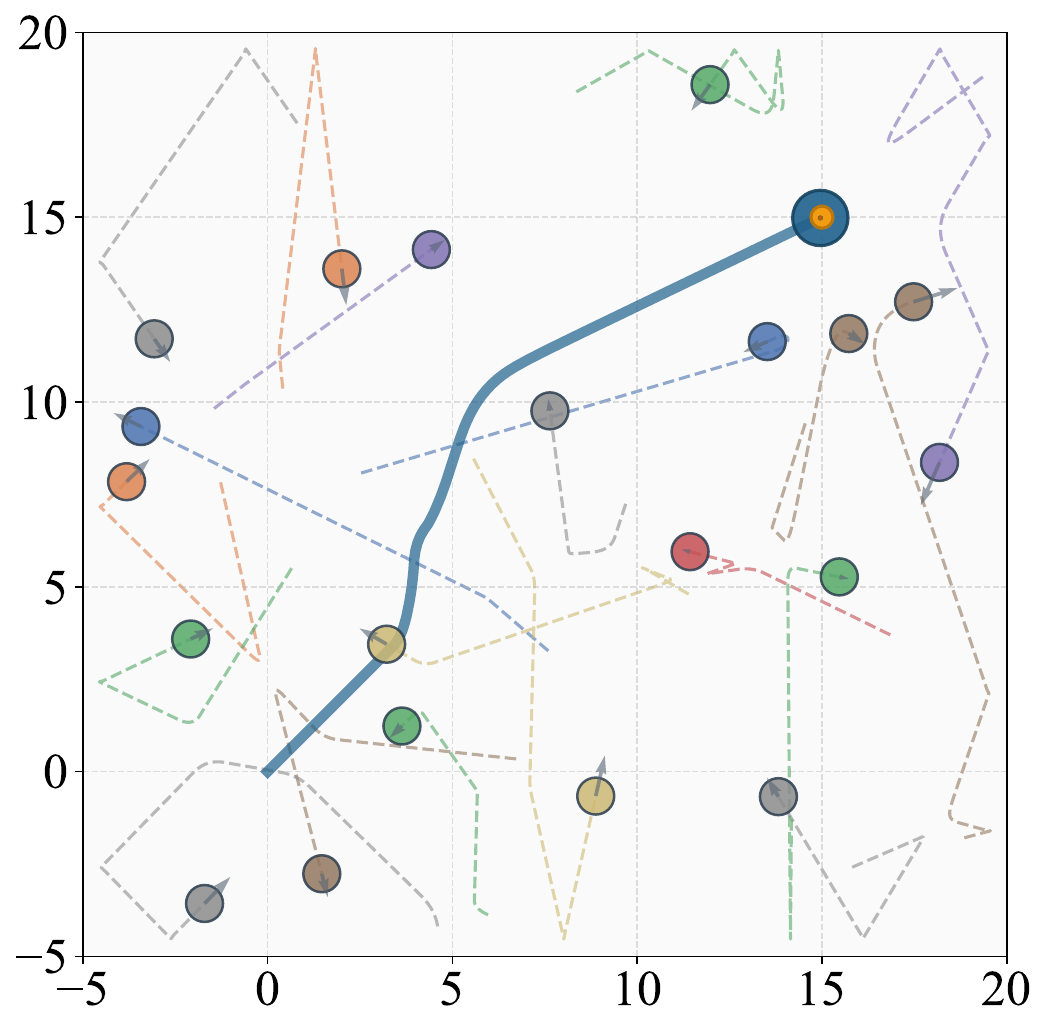}} \\

    \subfloat[ORCA]{\includegraphics[width=0.135\linewidth]{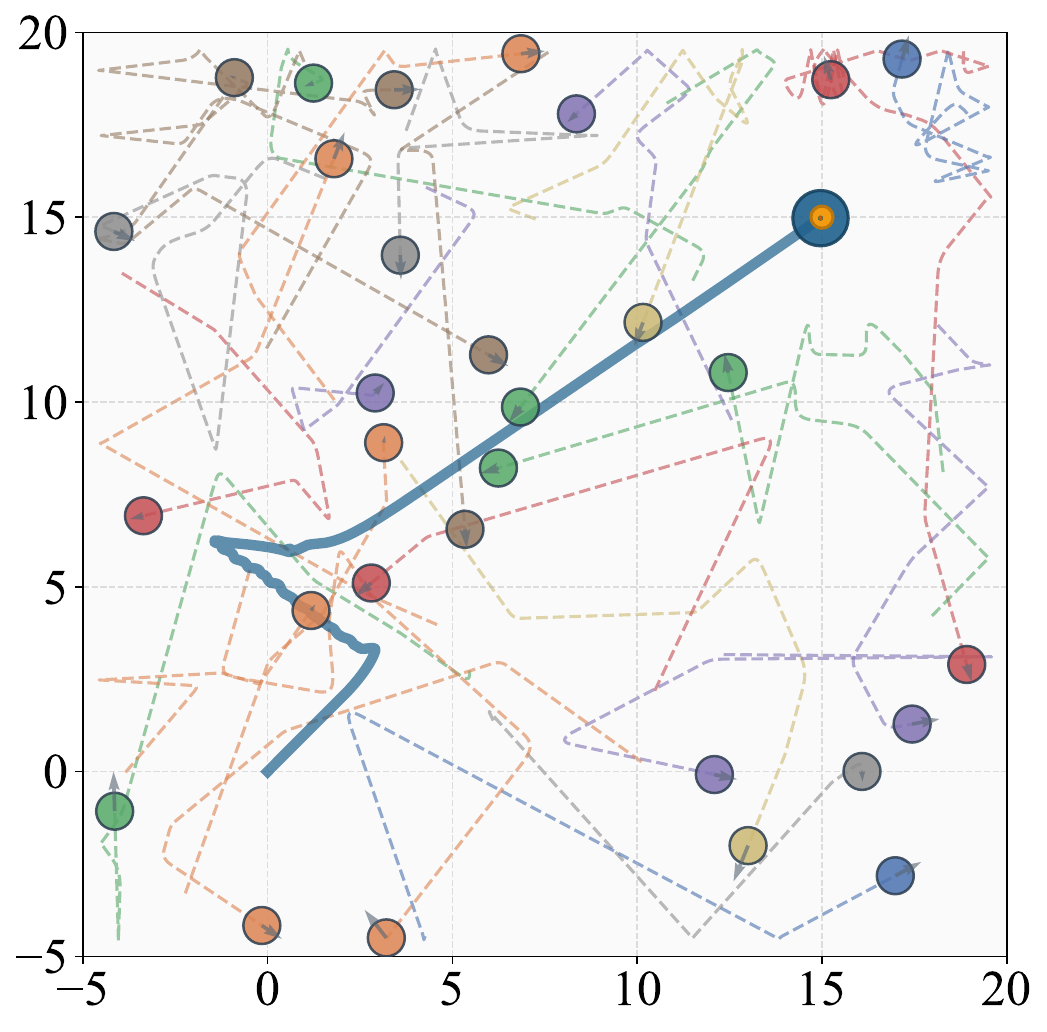}} \hfil
    \subfloat[DWA]{\includegraphics[width=0.135\linewidth]{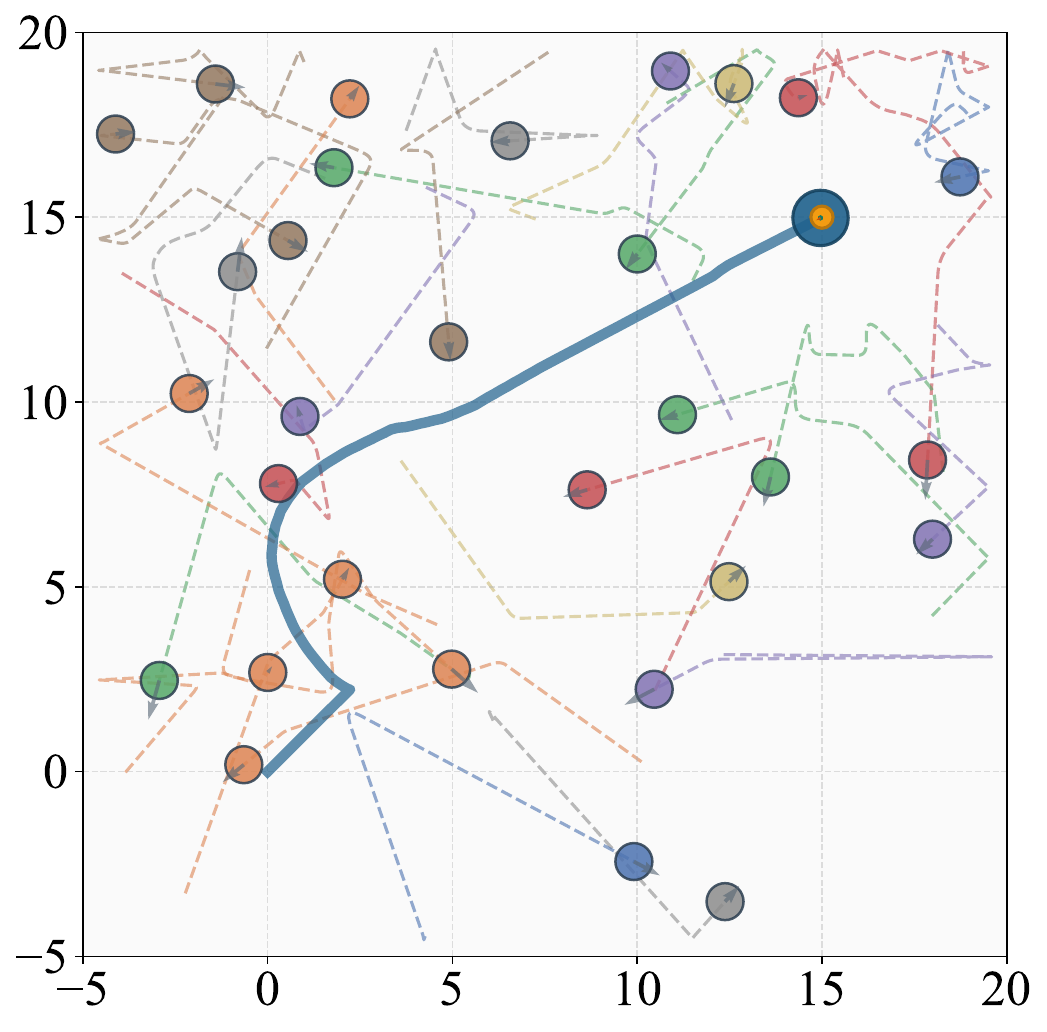}} \hfil
    \subfloat[MPC-CBF]{\includegraphics[width=0.135\linewidth]{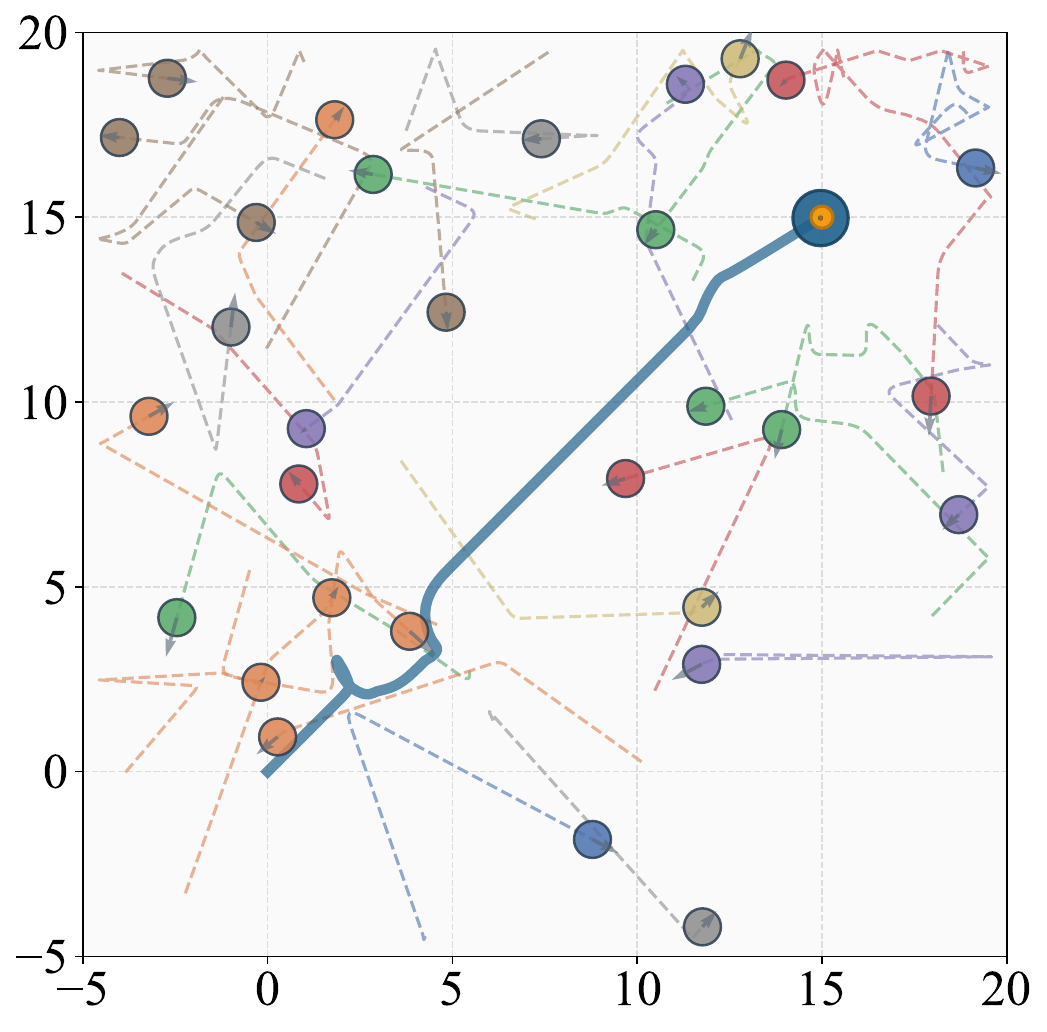}} \hfil
    \subfloat[SICNav]{\includegraphics[width=0.135\linewidth]{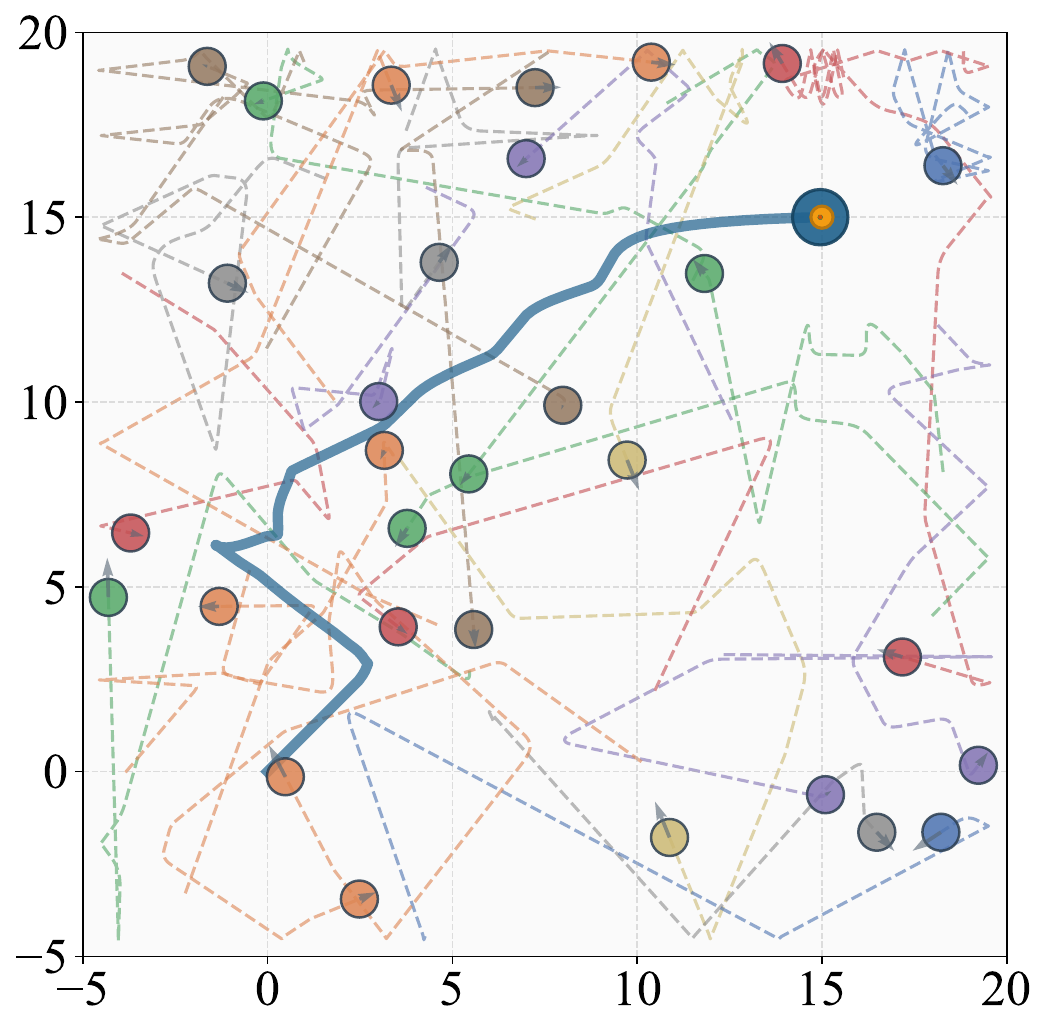}} \hfil
    \subfloat[AVO]{\includegraphics[width=0.135\linewidth]{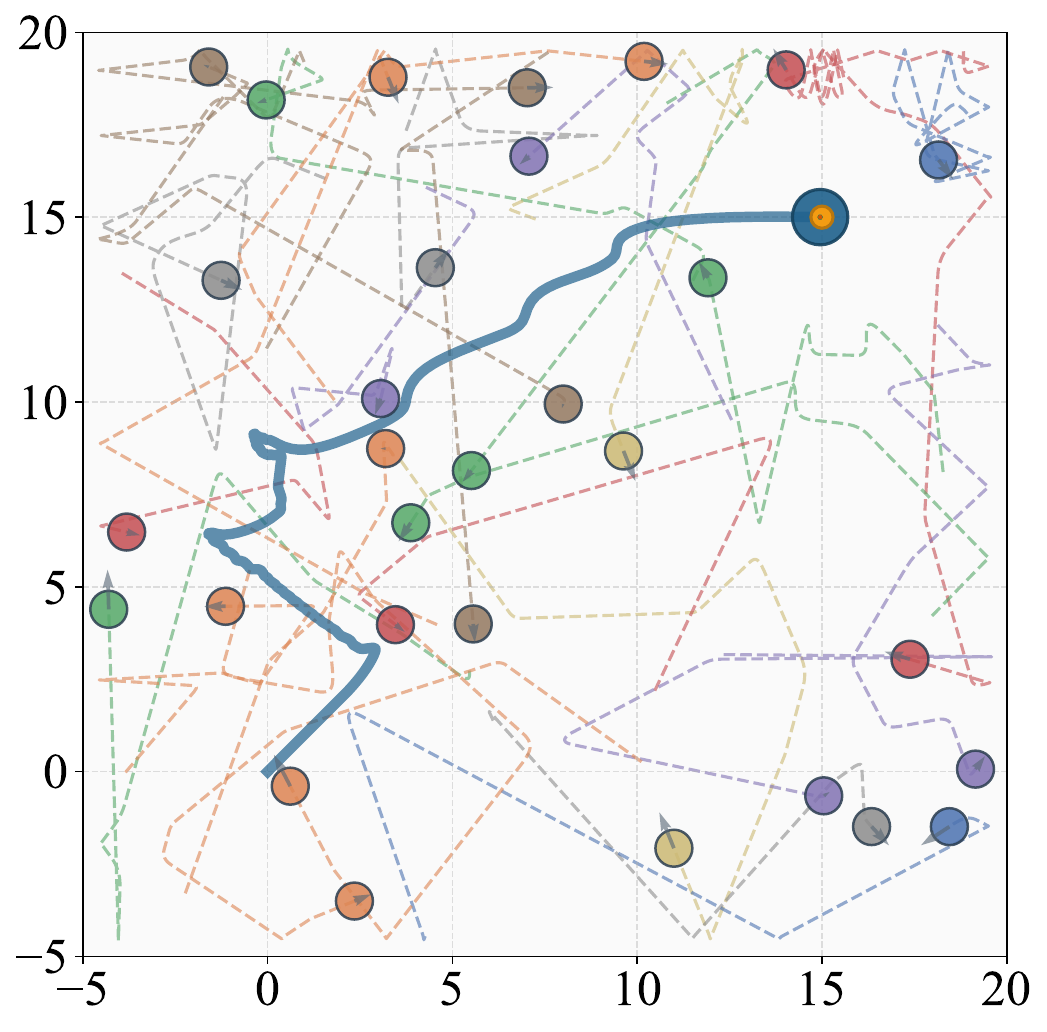}} \hfil
    \subfloat[AVOCADO]{\includegraphics[width=0.135\linewidth]{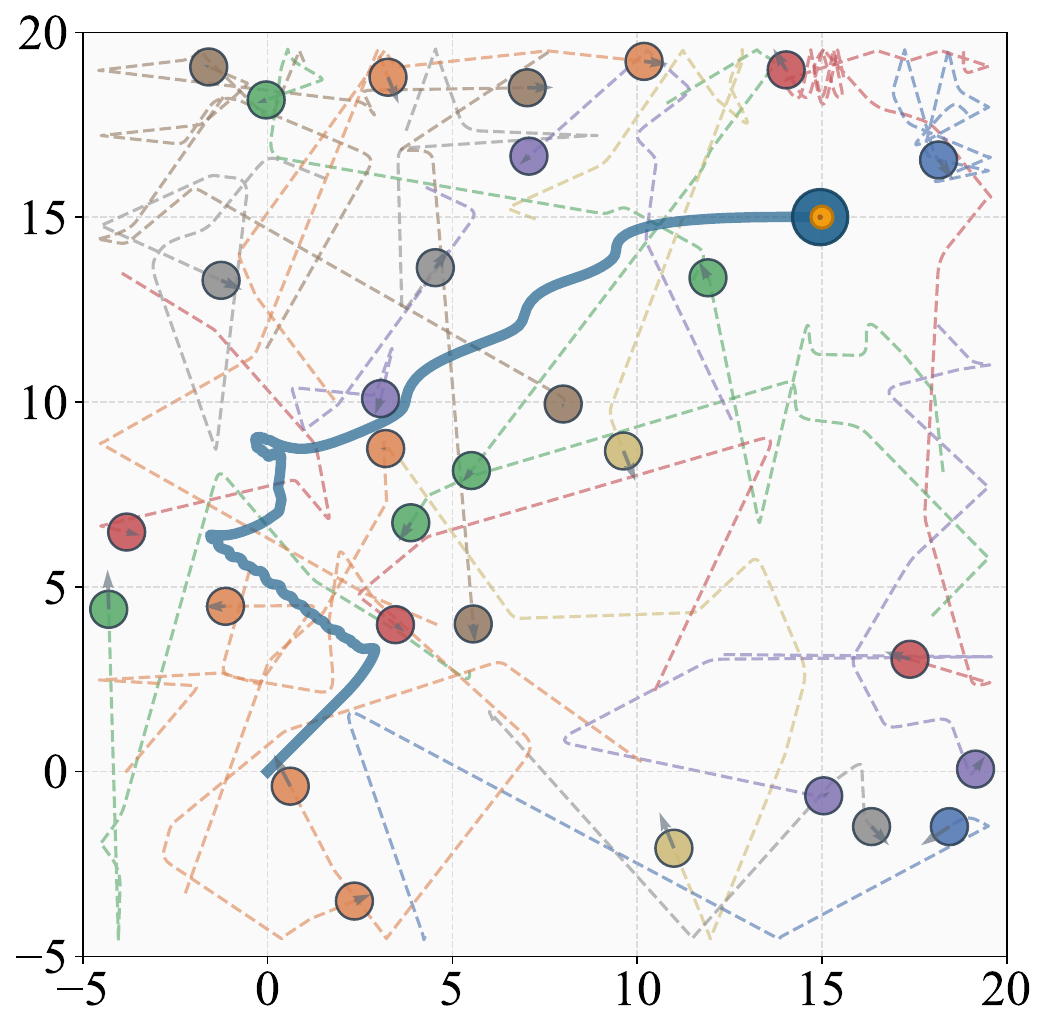}} \hfil
    \subfloat[GAO (Ours)]{\includegraphics[width=0.135\linewidth]{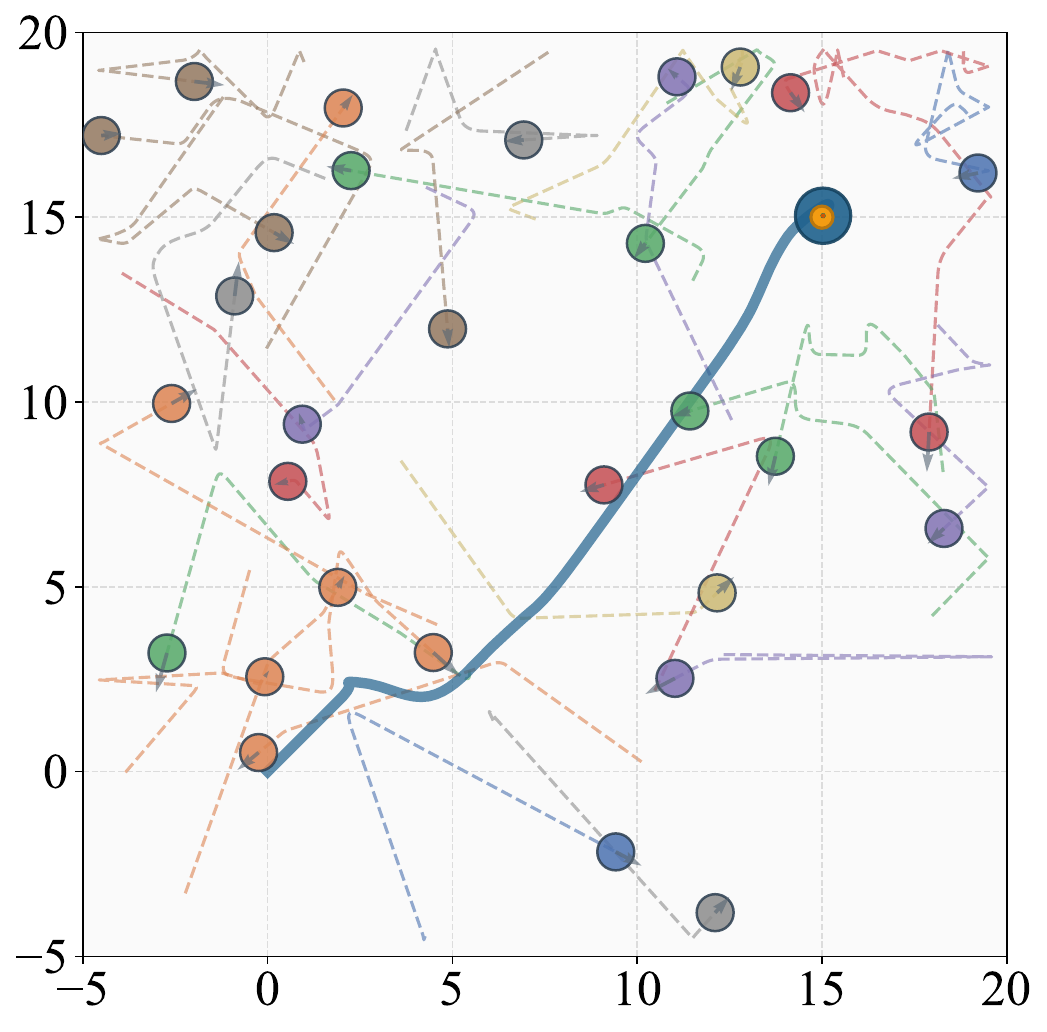}}
    
    \caption{Trajectory visualizations of the single-robot navigation experiments in noncooperative dynamic environments. The columns compare the evaluated planning algorithms, while the rows correspond to increasing spatial densities with $M=10$ (top), $M=20$ (middle), and $M=30$ (bottom) dynamic obstacles. Increasing obstacle density reduces free space and exposes differences in reactive behavior across planners.}
    \label{fig:sim_exp_1_comparison}
\par
\endgroup
\vspace{12pt}
\centering
\caption{Quantitative Comparison of Navigation Performance Across Different Scenarios. Values are reported as mean $\pm$ standard deviation, except for success rates.}
\label{tab:sim_1_results}
\resizebox{\textwidth}{!}{
\begin{tabular}{llccccc}
\toprule
\textbf{Scenario} & \textbf{Planner} & \textbf{Success Rate (\%) $\uparrow$} & \textbf{Navigation Time $t_f$ (s) $\downarrow$} & \textbf{Path Length $L$ (m) $\downarrow$} & \textbf{Control Effort $E_u$ (m$^2$/s$^3$) $\downarrow$} & \textbf{Computational Latency $\tau$ (ms) $\downarrow$} \\ 
\midrule
\multirow{7}{*}{\textbf{M=10}}
& \textbf{GAO (Ours)} & 99.0 & 15.04 $\pm$ 0.74 & 21.62 $\pm$ 0.94 & \textbf{14.83 $\pm$ 6.66} & 3.87 $\pm$ 0.98 \\
& ORCA & 97.0 & 15.27 $\pm$ 0.35 & 21.52 $\pm$ 0.51 & 19.85 $\pm$ 11.93 & \textbf{2.42 $\pm$ 0.60} \\
& MPC-CBF & \textbf{100.0} & \textbf{13.14 $\pm$ 1.02} & 22.07 $\pm$ 1.05 & 28.76 $\pm$ 14.87 & 15.99 $\pm$ 5.79 \\
& DWA & 96.0 & 15.78 $\pm$ 1.22 & 22.03 $\pm$ 1.31 & 33.73 $\pm$ 17.68 & 3.88 $\pm$ 2.23 \\
& AVO & 98.0 & 15.34 $\pm$ 0.50 & 21.59 $\pm$ 0.62 & 21.59 $\pm$ 13.84 & 4.14 $\pm$ 0.75 \\
& AVOCADO & 98.0 & 15.33 $\pm$ 0.49 & 21.58 $\pm$ 0.61 & 21.98 $\pm$ 14.21 & 4.14 $\pm$ 0.63 \\
& SICNav & 99.0 & 15.31 $\pm$ 0.61 & \textbf{21.43 $\pm$ 0.48} & 15.30 $\pm$ 8.15 & 27.20 $\pm$ 9.32 \\ 
\midrule
\multirow{7}{*}{\textbf{M=20}}
& \textbf{GAO (Ours)} & \textbf{98.0} & 15.23 $\pm$ 1.25 & 22.17 $\pm$ 1.68 & \textbf{19.26 $\pm$ 7.95} & 3.92 $\pm$ 0.96 \\
& ORCA & 86.0 & 15.63 $\pm$ 0.82 & 21.91 $\pm$ 0.79 & 30.47 $\pm$ 16.37 & \textbf{2.56 $\pm$ 0.68} \\
& MPC-CBF & 95.0 & \textbf{14.21 $\pm$ 1.89} & 23.10 $\pm$ 1.89 & 46.64 $\pm$ 16.57 & 19.58 $\pm$ 7.99 \\
& DWA & 89.0 & 17.23 $\pm$ 3.34 & 23.86 $\pm$ 4.02 & 57.47 $\pm$ 34.03 & 5.38 $\pm$ 3.72 \\
& AVO & \textbf{98.0} & 16.07 $\pm$ 2.00 & 22.30 $\pm$ 1.93 & 37.11 $\pm$ 23.62 & 4.28 $\pm$ 0.70 \\
& AVOCADO & 97.0 & 16.06 $\pm$ 2.04 & 22.28 $\pm$ 1.99 & 39.24 $\pm$ 30.03 & 4.36 $\pm$ 0.75 \\
& SICNav & 96.0 & 15.87 $\pm$ 1.65 & \textbf{21.87 $\pm$ 1.45} & 21.31 $\pm$ 12.43 & 33.89 $\pm$ 15.24 \\ 
\midrule
\multirow{7}{*}{\textbf{M=30}} 
& \textbf{GAO (Ours)} & \textbf{96.0} & 16.09 $\pm$ 2.45 & 23.18 $\pm$ 3.15 & \textbf{24.75 $\pm$ 11.99} & 4.21 $\pm$ 1.28 \\
& ORCA & 67.0 & 15.93 $\pm$ 1.26 & \textbf{22.10 $\pm$ 0.90} & 37.52 $\pm$ 20.67 & \textbf{3.40 $\pm$ 2.00} \\
& MPC-CBF & 85.0 & \textbf{15.65 $\pm$ 3.01} & 24.16 $\pm$ 2.73 & 65.57 $\pm$ 26.57 & 22.61 $\pm$ 9.30 \\
& DWA & 82.0 & 18.26 $\pm$ 2.94 & 24.49 $\pm$ 3.25 & 73.27 $\pm$ 33.24 & 7.12 $\pm$ 5.31 \\
& AVO & 85.0 & 17.14 $\pm$ 3.40 & 23.11 $\pm$ 2.72 & 53.07 $\pm$ 39.10 & 4.51 $\pm$ 0.85 \\
& AVOCADO & 89.0 & 17.27 $\pm$ 3.34 & 23.21 $\pm$ 2.59 & 60.29 $\pm$ 48.80 & 4.61 $\pm$ 0.85 \\
& SICNav & 92.0 & 16.66 $\pm$ 2.58 & 22.23 $\pm$ 1.55 & 26.13 $\pm$ 14.28 & 42.02 $\pm$ 21.19 \\ 
\bottomrule
\end{tabular}
}
\end{table*}

\subsection{Single Robot Navigation in Dynamic and Noncooperative Environments}
\label{sec:sim_1}
The first set of experiments evaluates the ego-robot's reactive safety and motion smoothness in a bounded planar workspace $\mathcal{W} \subset \mathbb{R}^2$ ($25\text{ m} \times 25\text{ m}$).
The environment is populated with $M \in \{10, 20, 30\}$ noncooperative dynamic obstacles.

\indent\textit{Scenario Configuration:}
The $M$ dynamic obstacles are simulated as a decentralized crowd governed by a softened Social Force Model (SFM). 
Assuming unit mass, the acceleration of the $i$-th obstacle is driven by mutual repulsion from its neighbors:
\begin{equation}
    \mathbf{a}_i(t) = \sum_{j \in \mathcal{N}_i} \frac{k}{\|\mathbf{p}_i - \mathbf{p}_j\|^2 + \epsilon_1} \cdot \frac{\mathbf{p}_i - \mathbf{p}_j}{\|\mathbf{p}_i - \mathbf{p}_j\| + \epsilon_2} ,
\end{equation}
where $k$ is the repulsion gain, $\epsilon_1, \epsilon_2 > 0$ are regularizing constants, and $\mathcal{N}_i = \{ j \neq i \mid \|\mathbf{p}_i - \mathbf{p}_j\| < R_i + R_j + d_{\text{margin}} \}$ is the local neighborhood. 
The ego-robot is excluded from the perceptual field of all $M$ obstacles. 
The obstacles therefore do not react to the ego-robot, which bears full responsibility for avoidance. 

The workspace boundary uses elastic reflection to maintain approximately constant obstacle density.
When obstacle $j$ reaches the boundary at time $t_r$, its velocity is updated as
\begin{equation}
    \mathbf{v}_j(t_r^+) = \mathbf{v}_j(t_r^-) - 2(\mathbf{v}_j(t_r^-) \cdot \mathbf{n})\mathbf{n} ,
\end{equation}
where $\mathbf{n}$ is the inward-pointing unit normal of the boundary.

\indent\textit{Evaluation Metrics:}
The seven methods are evaluated on the same set of 100 randomized trials for each density level $M$, using identical random seeds across methods.
This results in 2,100 runs in total, with performance assessed using five primary metrics.
Success Rate (\%) is the percentage of collision-free trials reaching the goal.
Navigation Time ($t_f$) measures task completion time, while Path Length ($L$) measures spatial efficiency and is defined as $L = \int_0^{t_f} \|\mathbf{v}(t)\| dt$.
Control Effort ($E_u$) is the integrated squared acceleration, computed as $E_u = \int_0^{t_f} \|\mathbf{u}(t)\|^2 dt$.
Navigation time, path length, and control effort are summarized over collision-free goal-reaching trials.
Computation Latency ($\tau$) measures the time required by the planner to compute a control command. We report the mean and standard deviation of the recorded execution times from collision-free goal-reaching trials.

\indent\textit{Results and Discussion:}
Fig. \ref{fig:sim_exp_1_comparison} compares the ego-robot trajectories across obstacle densities ($M \in \{10, 20, 30\}$), and Table \ref{tab:sim_1_results} summarizes the corresponding performance metrics.
Higher obstacle densities reduce free space and increase avoidance difficulty.

In the tested scenarios, GAO maintains high success rates as obstacle density increases, reaching 96\% at $M=30$, and has the lowest control effort across all three densities.
MPC-CBF achieves the highest success rate at $M=10$.
The contrast between low- and high-density cases suggests that GAO's advantage is more apparent when frequent avoidance interactions make task completion difficult.
These results indicate favorable performance under dense noncooperative interactions but do not establish unconditional safety for the practical controller.

Spatial and temporal efficiency show a different ranking.
At $M=30$, ORCA and SICNav achieve shorter paths than GAO, while MPC-CBF has the shortest mean navigation time.
However, the differences in path length and navigation time are modest.
GAO adjusts acceleration to accommodate avoidance constraints while tracking the nominal command.
Its objective does not explicitly minimize path length or arrival time.
The results illustrate a trade-off among success, control effort, and navigation efficiency.

GAO's mean planning computation time increases with obstacle density, reaching $3.87$, $3.92$, and $4.21\,\mathrm{ms}$ at $M=10$, $20$, and $30$, respectively.
It is comparable to AVOCADO and lower than SICNav.

\subsection{Reciprocal Collision Avoidance in Multi-Agent Scenarios}
\label{sec:sim_2}
We evaluate reciprocal avoidance with all agents using identical planning logic. 
The setup tests symmetric conflict resolution and collective progress under kinodynamic constraints.

\indent\textit{Scenario Configuration:}
We use the circle-interchange benchmark with $N\in\{4,8,12\}$ agents.
Agents are placed uniformly on a circle of radius $10\,\mathrm{m}$, with the antipodal point assigned as each agent's goal.
The symmetric geometry brings all agents toward the center simultaneously, creating a congested bottleneck.

\indent\textit{Evaluation Metrics:}
We use two metrics to evaluate reciprocal coordination.
Collective safety is a binary criterion, with any physical contact between two agents counted as a failure.
Navigation time ($t_f$) measures the time required for all agents to reach their goals and indicates coordination efficiency as interaction complexity increases.

% In contrast to the randomized trials in Section \ref{sec:sim_1}, the reciprocal benchmarks use a deterministic configuration with fixed initial and goal positions. 
% A single trial is conducted for each algorithm at each agent count $N$.

\begin{figure*}[htbp]
    \centering
    \subfloat[ORCA]{\includegraphics[width=0.16\linewidth]{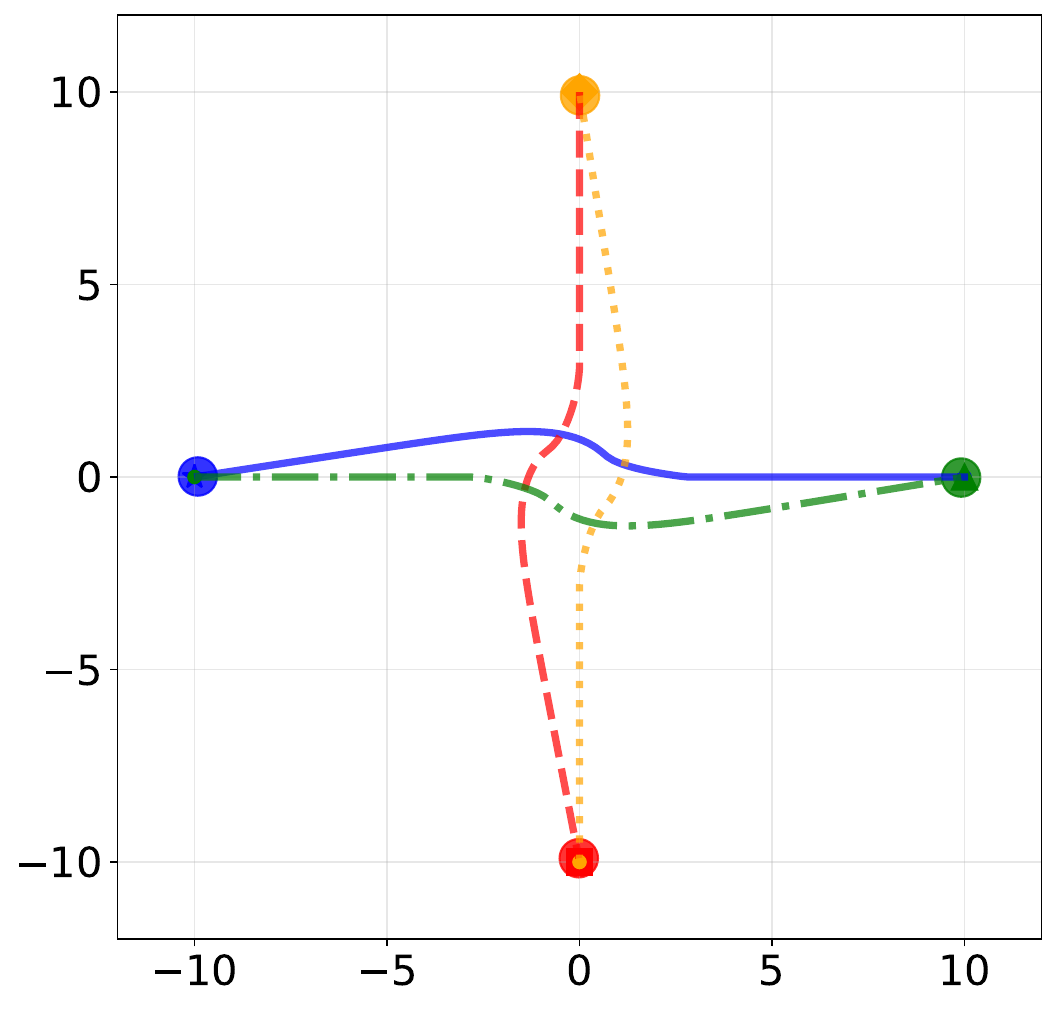}}\hfill
    \subfloat[DWA]{\includegraphics[width=0.16\linewidth]{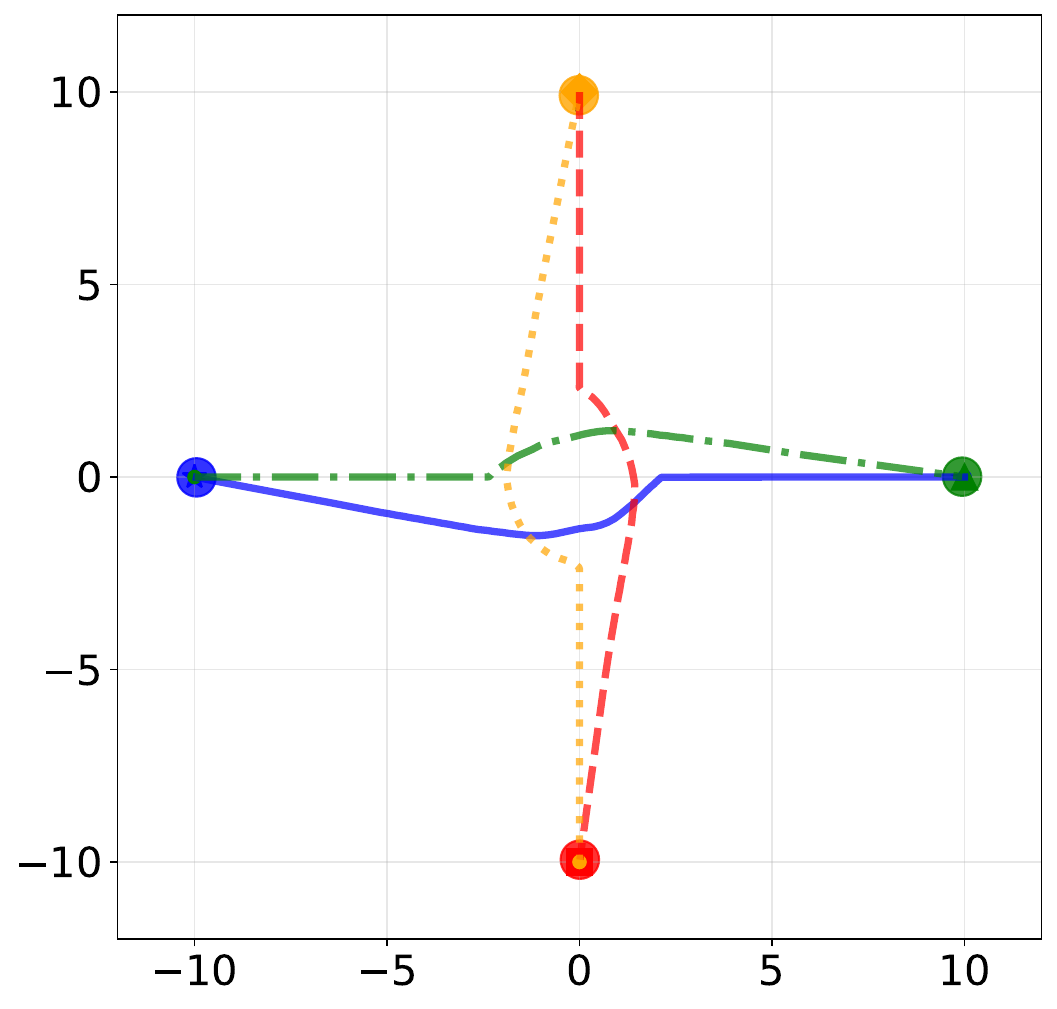}}\hfill
    \subfloat[MPC-CBF]{\includegraphics[width=0.16\linewidth]{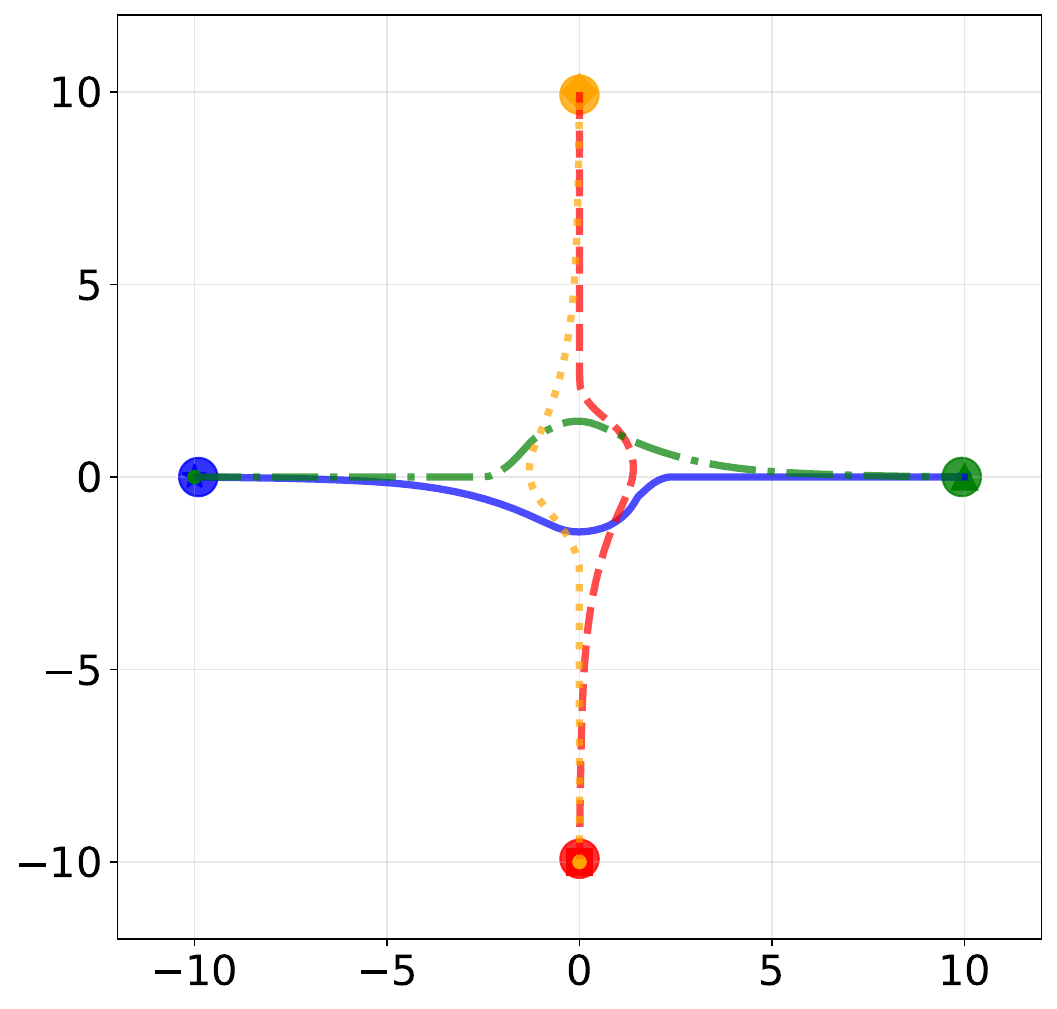}}\hfill
    \subfloat[AVO]{\includegraphics[width=0.16\linewidth]{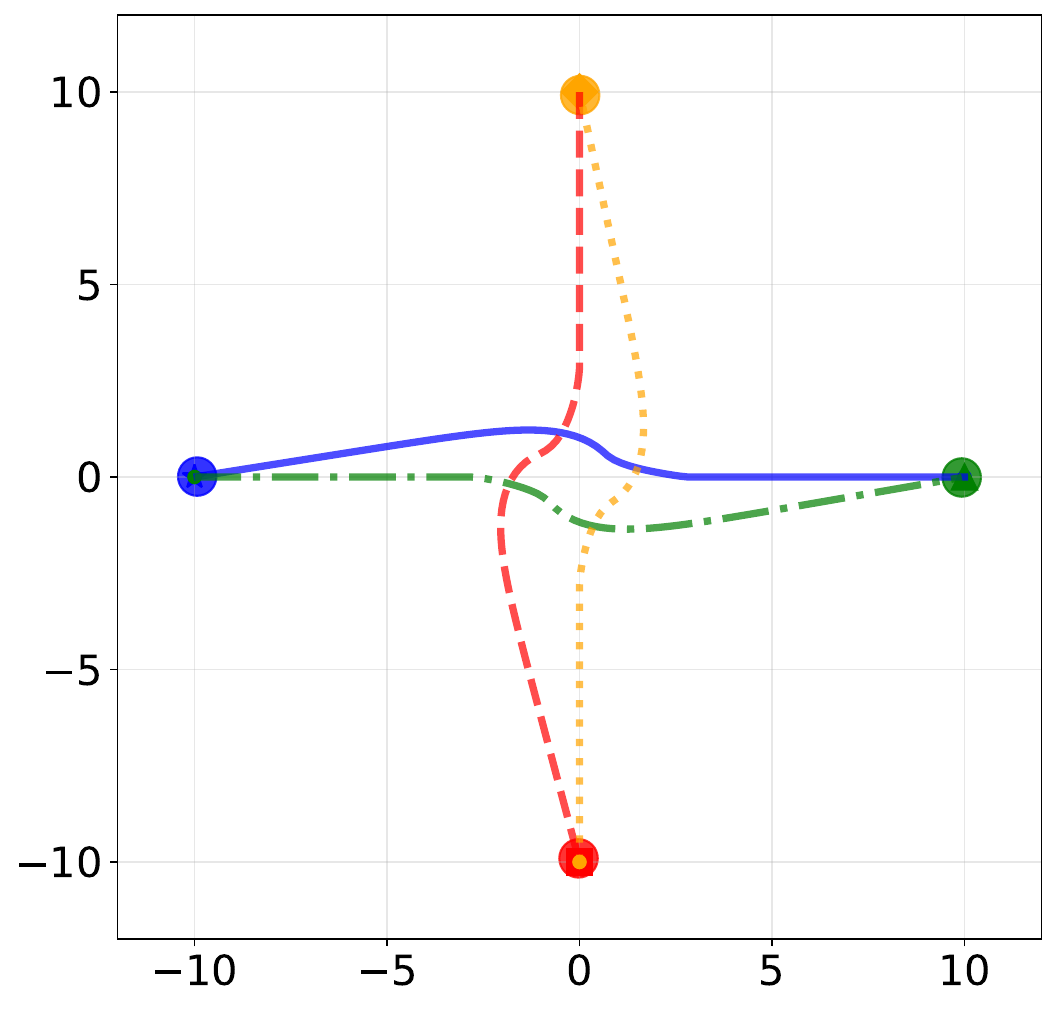}}\hfill
    \subfloat[AVOCADO]{\includegraphics[width=0.16\linewidth]{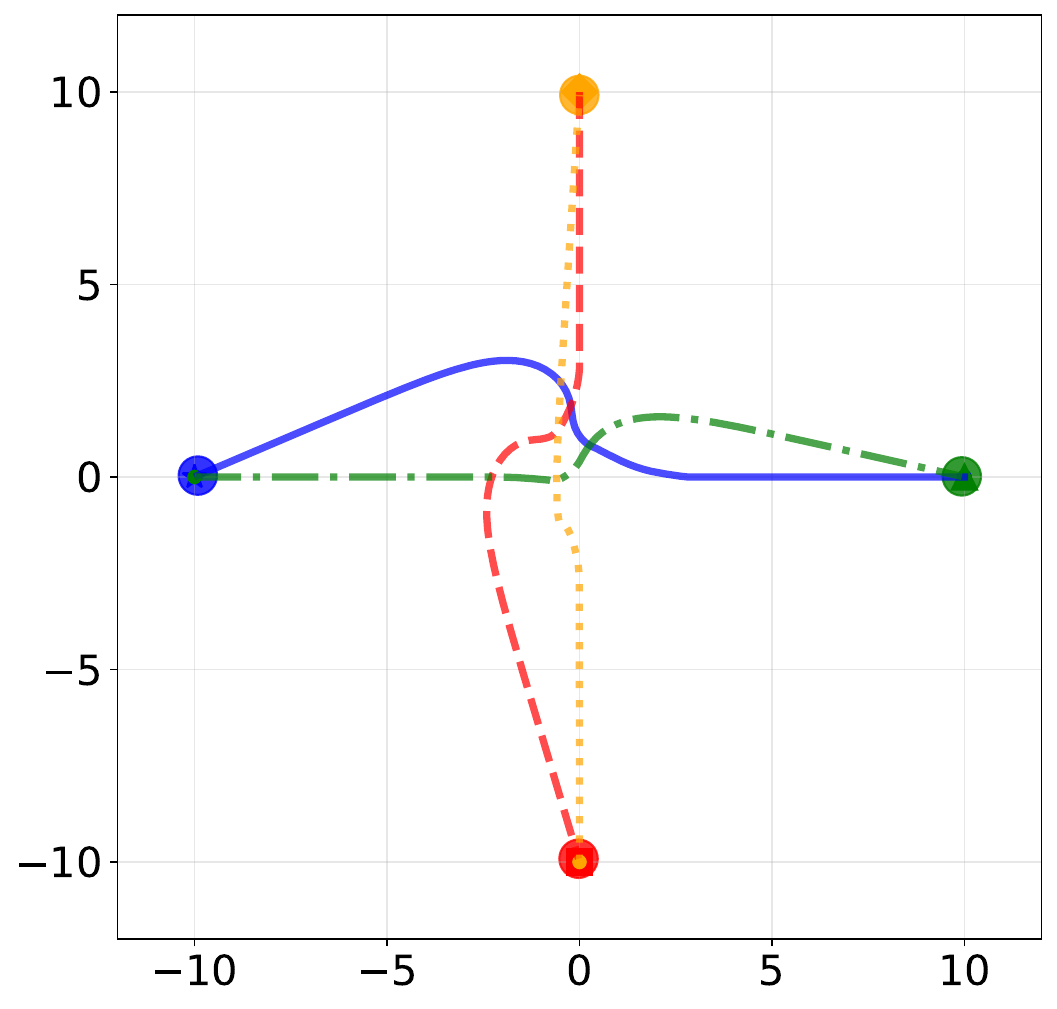}}\hfill
    \subfloat[GAO (Ours)]{\includegraphics[width=0.16\linewidth]{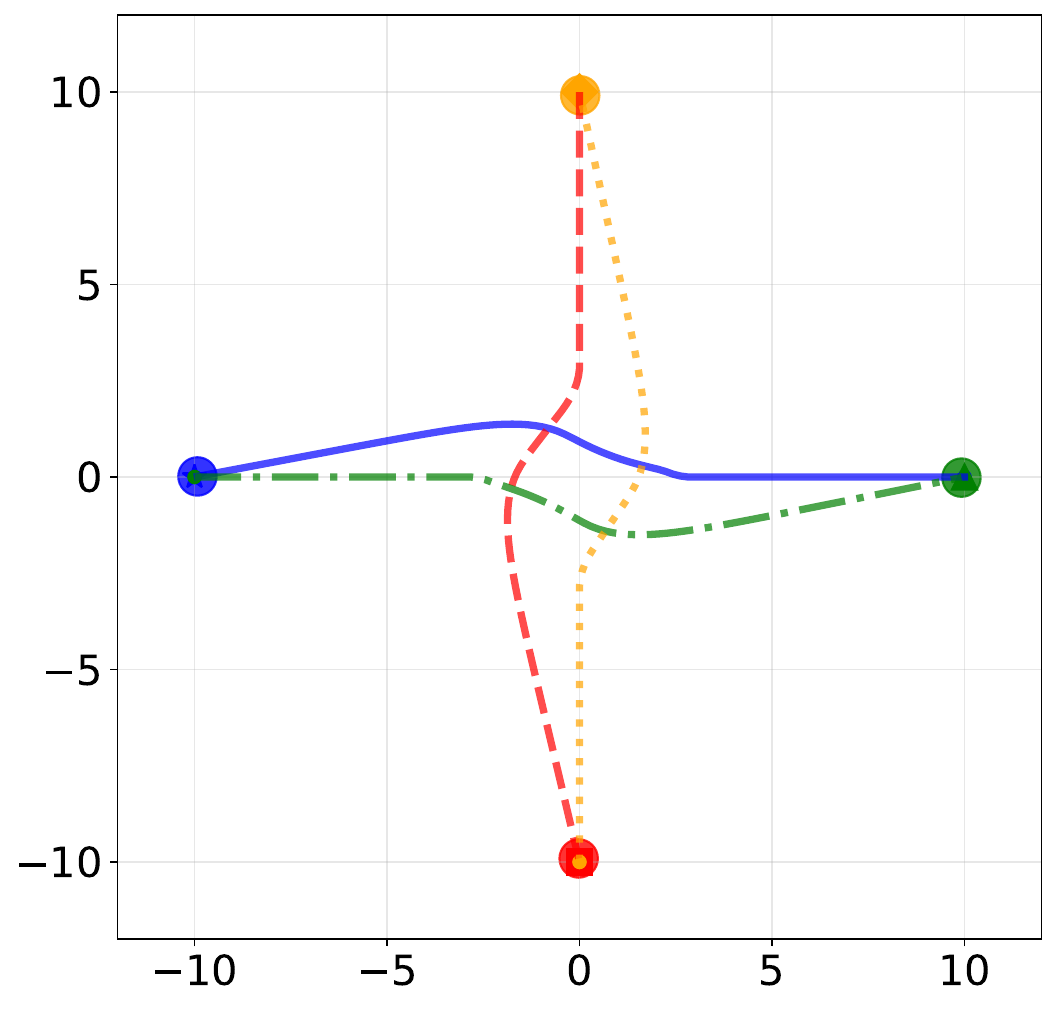}}
    \\
    \subfloat[ORCA]{\includegraphics[width=0.16\linewidth]{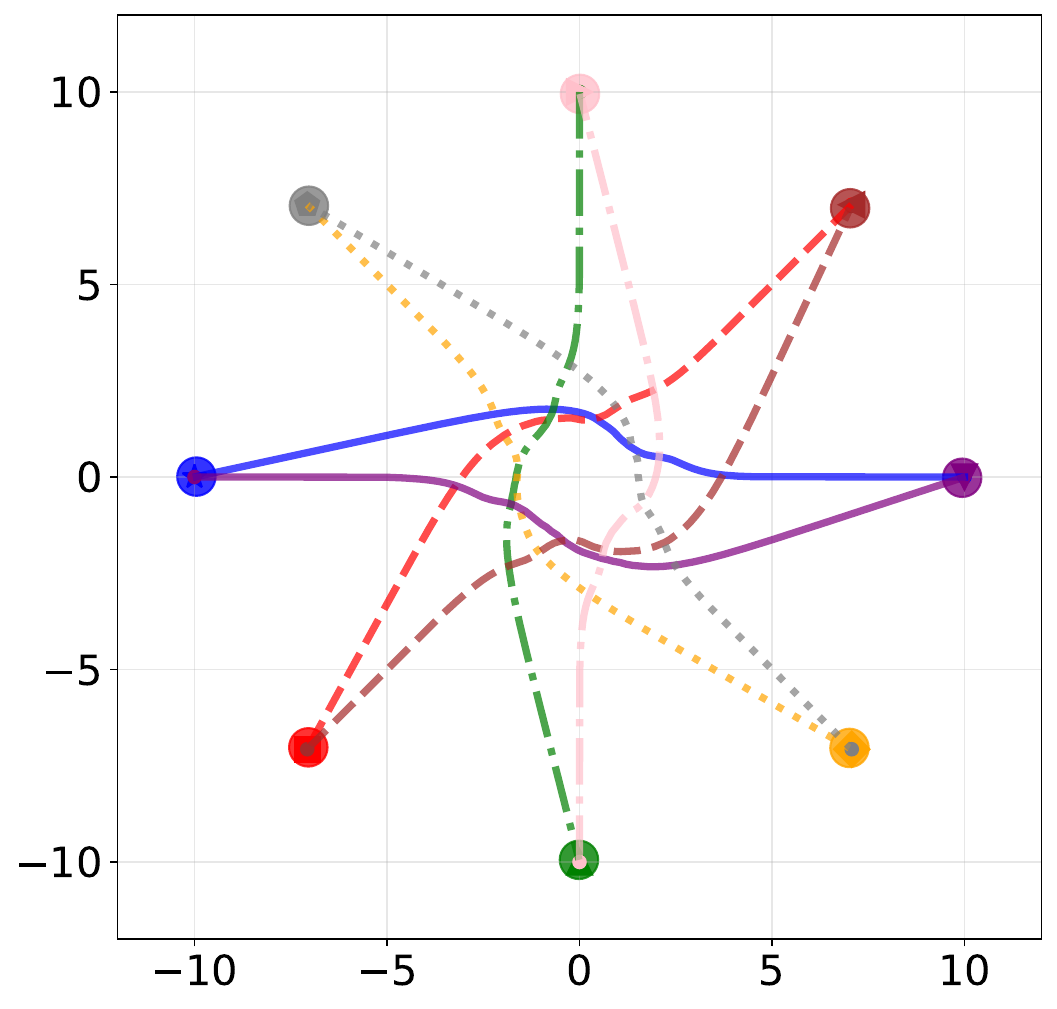}}\hfill
    \subfloat[DWA]{\includegraphics[width=0.16\linewidth]{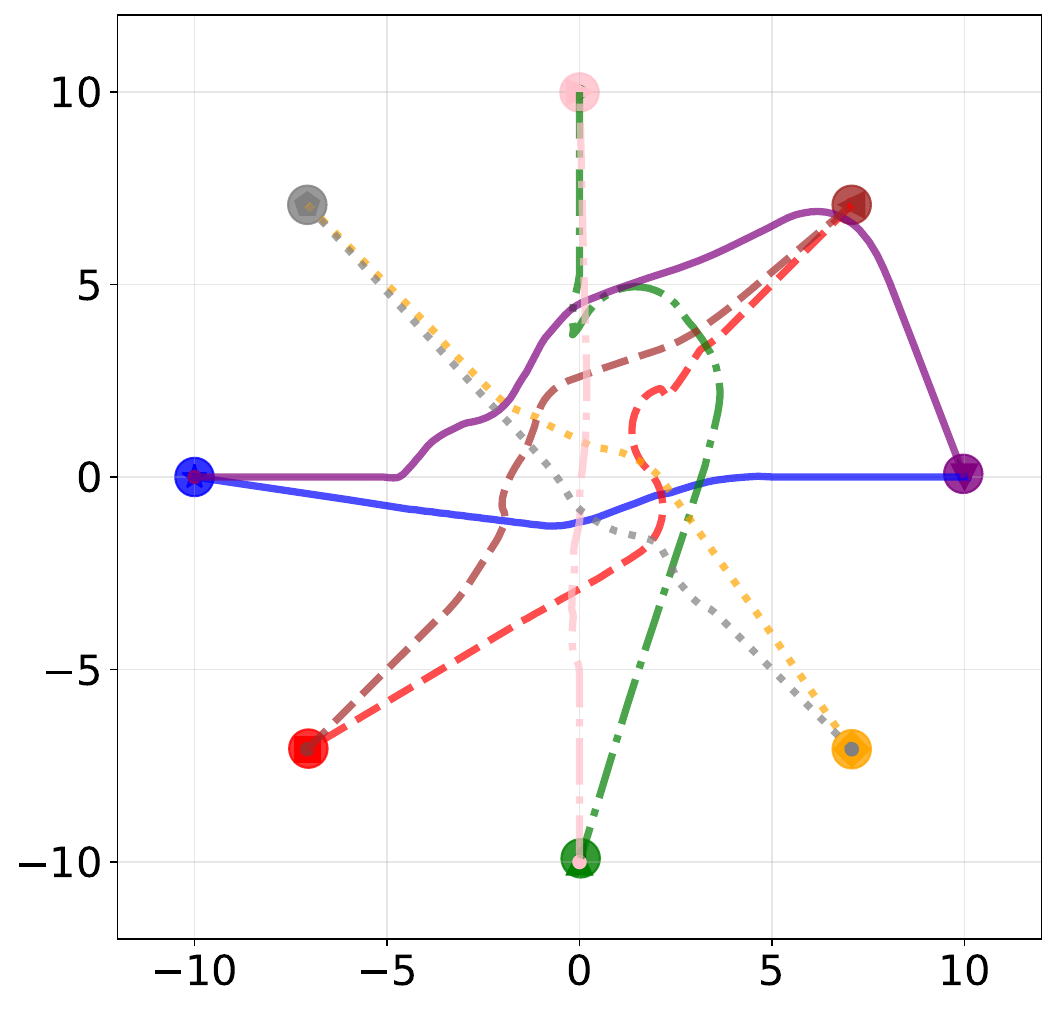}}\hfill
    \subfloat[MPC-CBF]{\includegraphics[width=0.16\linewidth]{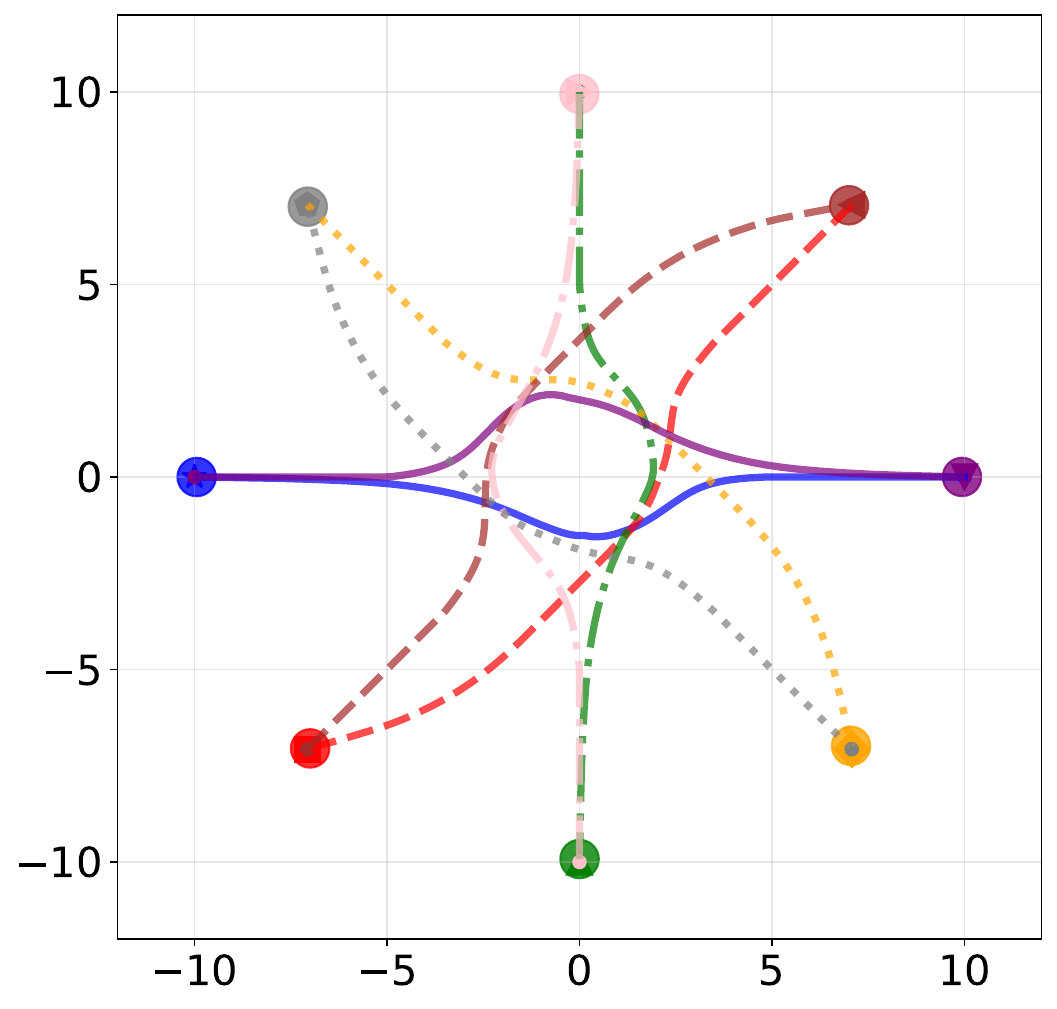}}\hfill
    \subfloat[AVO]{\includegraphics[width=0.16\linewidth]{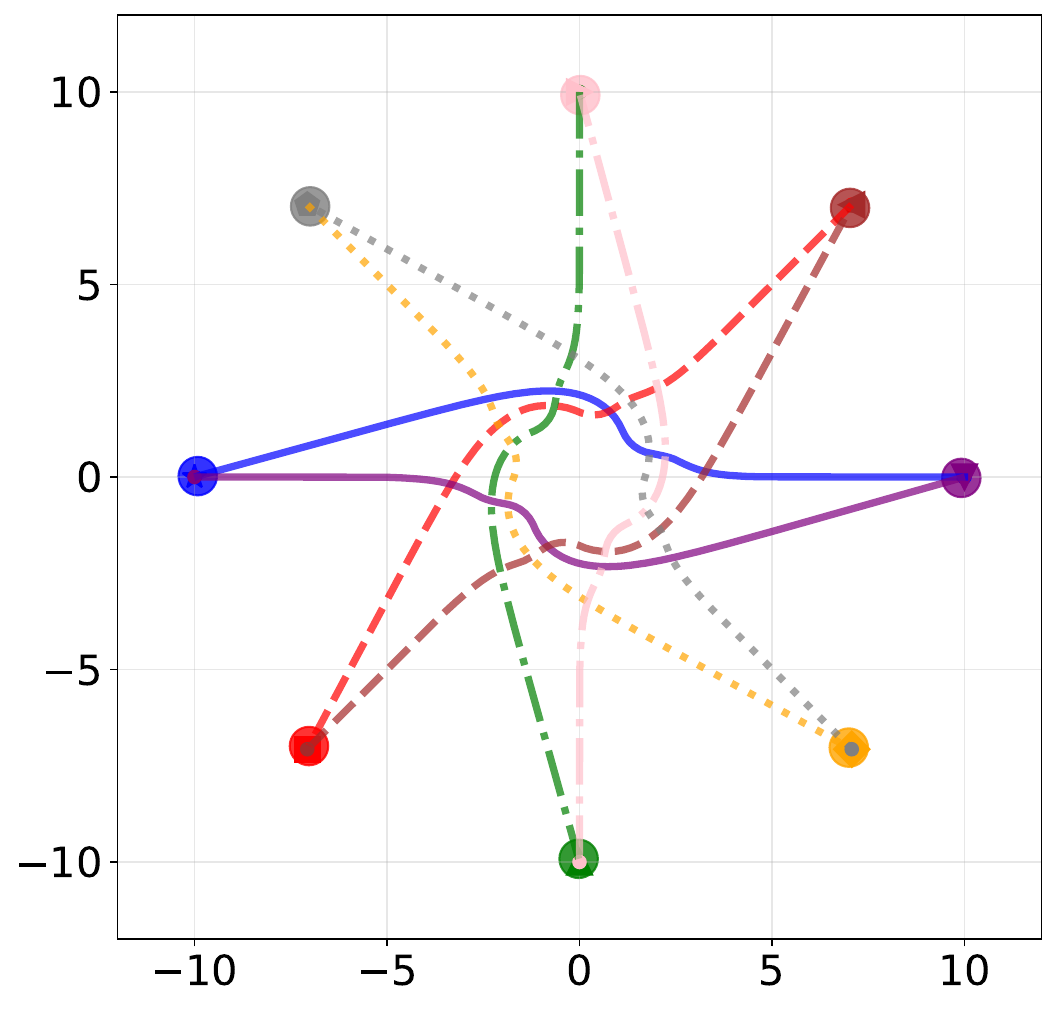}}\hfill
    \subfloat[AVOCADO]{\includegraphics[width=0.16\linewidth]{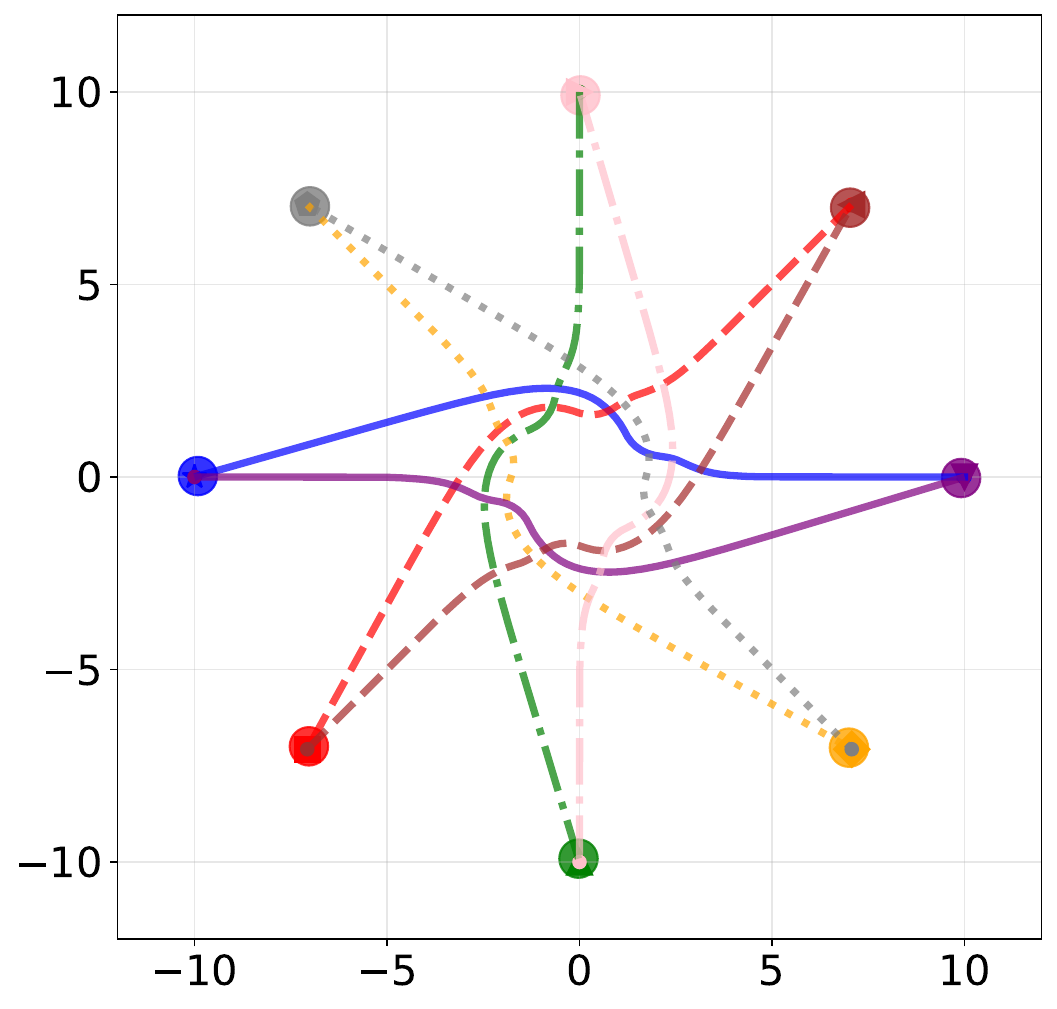}}\hfill
    \subfloat[GAO (Ours)]{\includegraphics[width=0.16\linewidth]{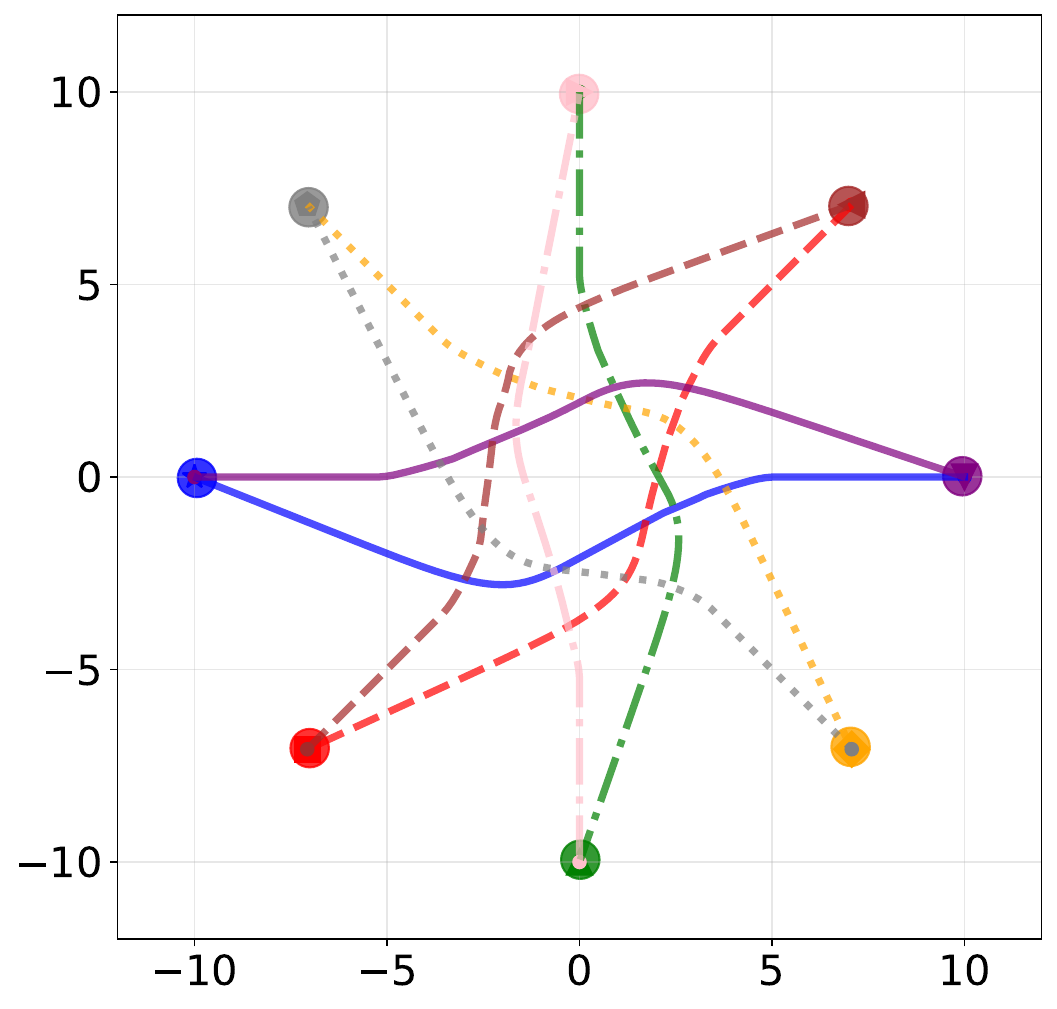}}
    \\
    \subfloat[ORCA]{\includegraphics[width=0.16\linewidth]{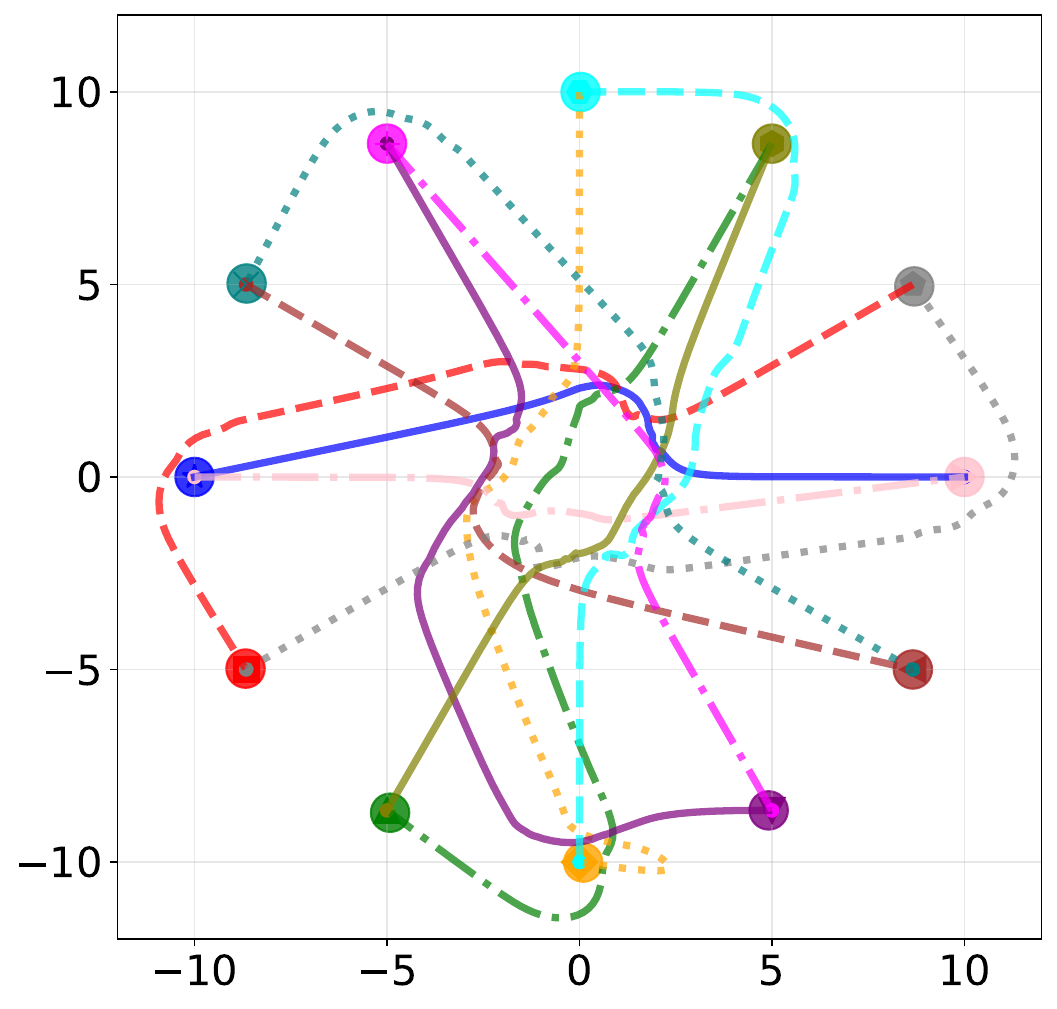}}\hfill
    \subfloat[DWA]{\includegraphics[width=0.16\linewidth]{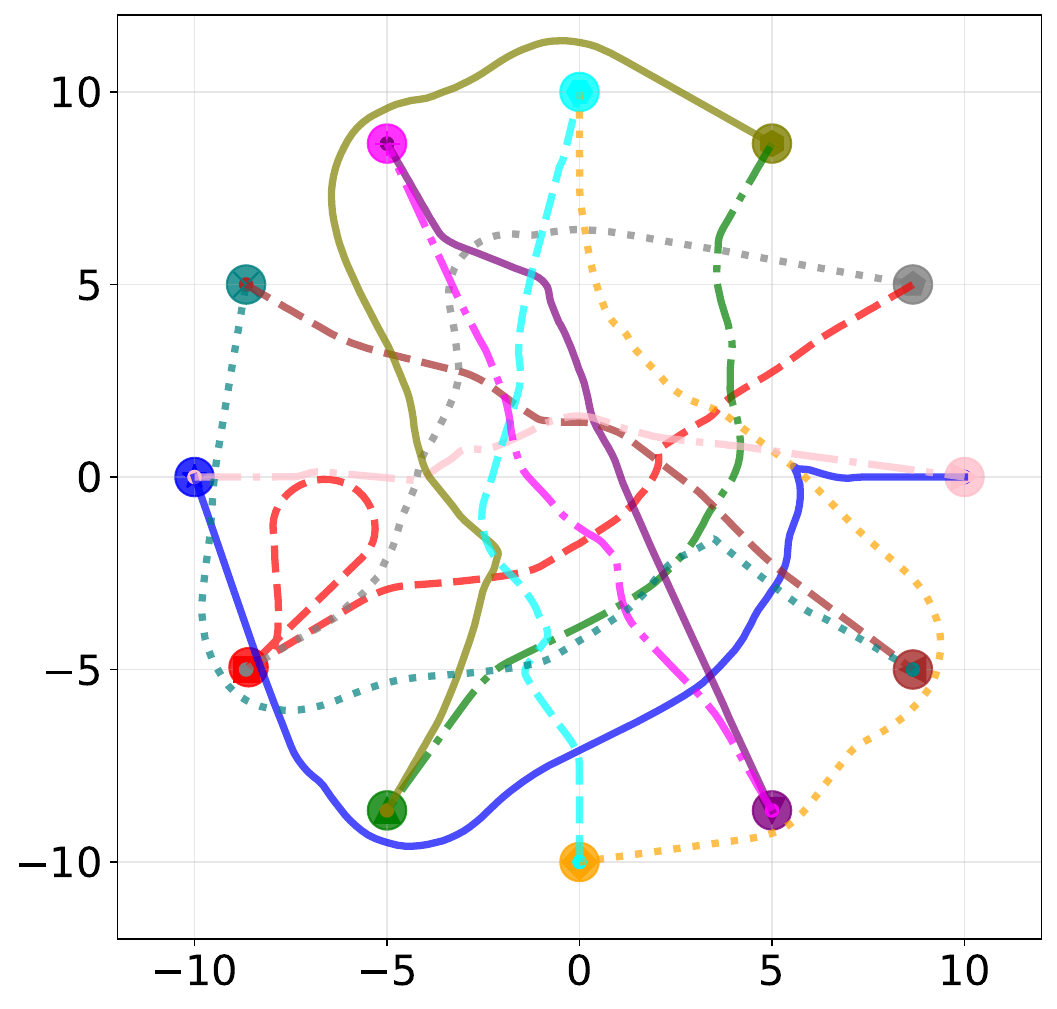}}\hfill
    \subfloat[MPC-CBF]{\includegraphics[width=0.16\linewidth]{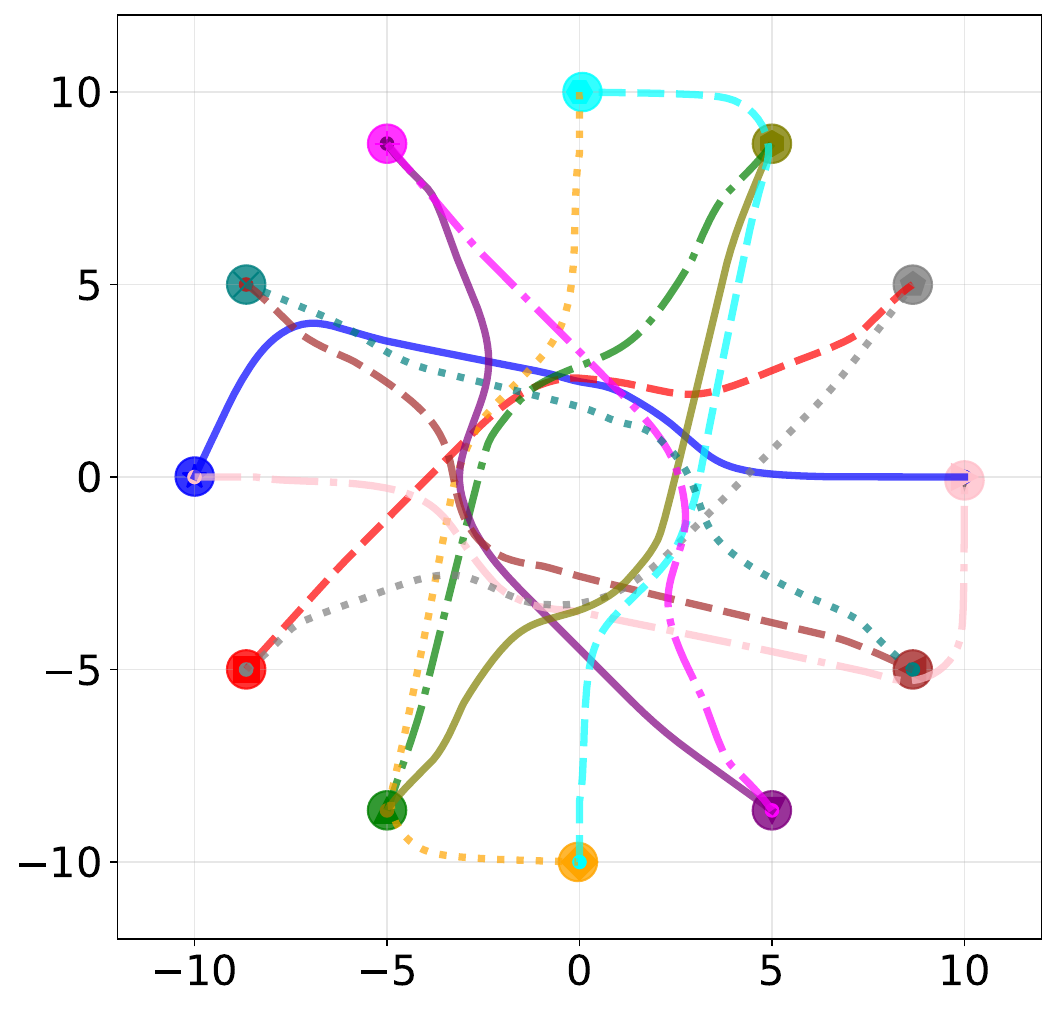}}\hfill
    \subfloat[AVO]{\includegraphics[width=0.16\linewidth]{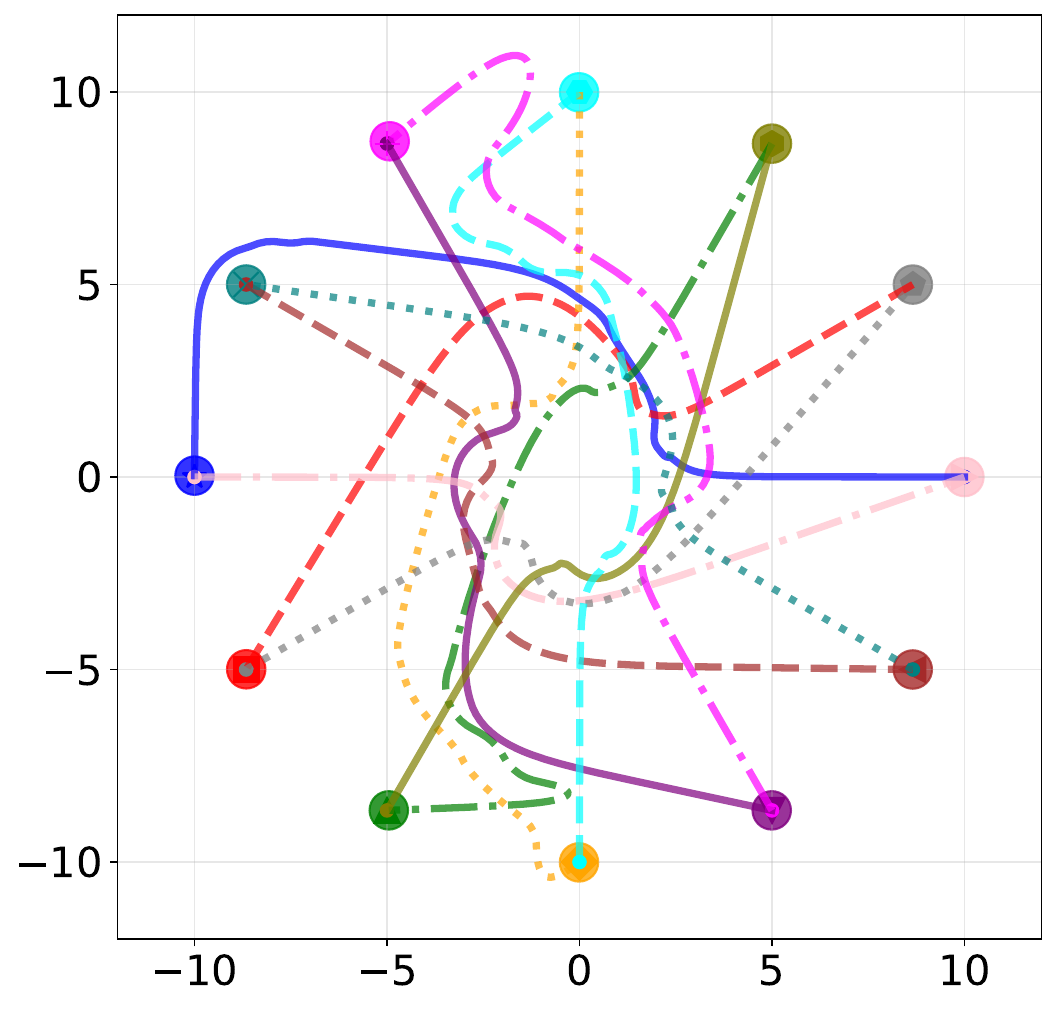}}\hfill
    \subfloat[AVOCADO]{\includegraphics[width=0.16\linewidth]{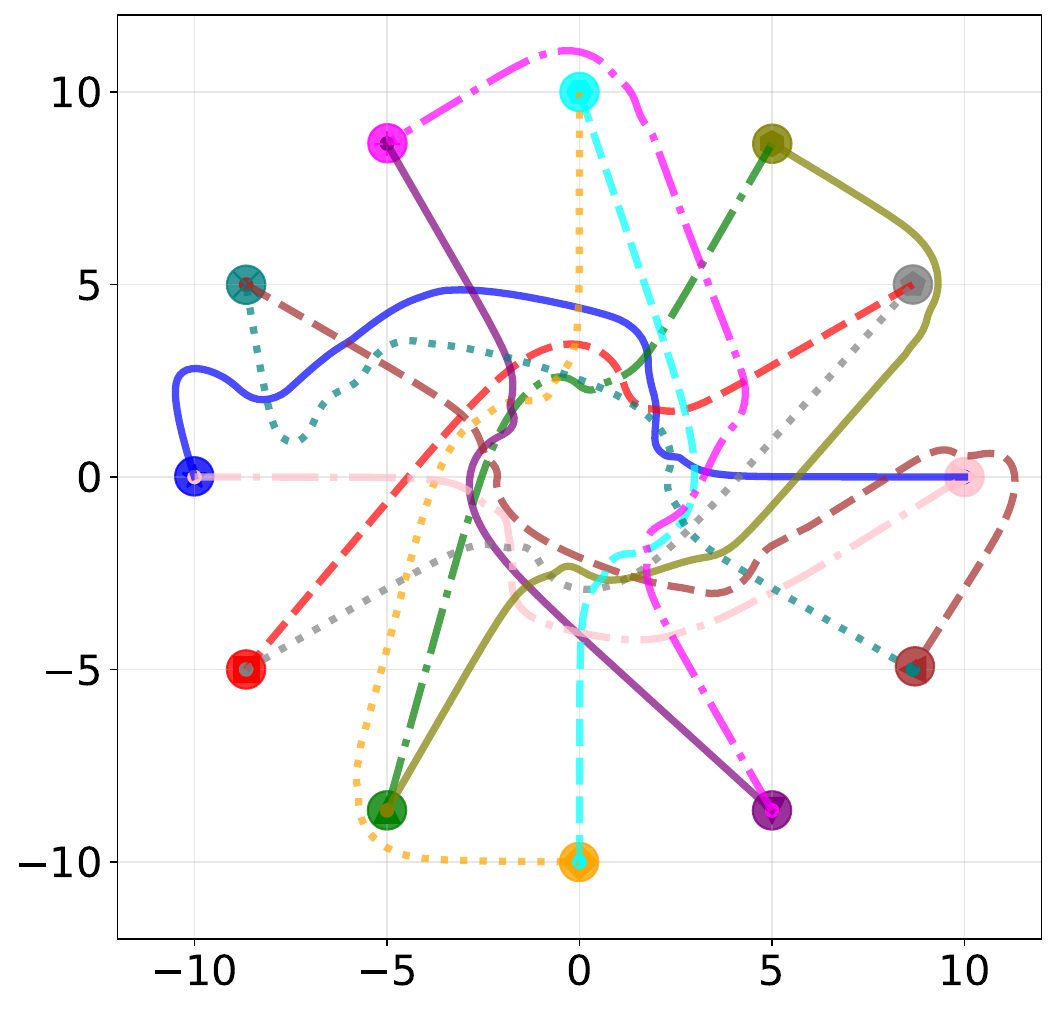}}\hfill
    \subfloat[GAO (Ours)]{\includegraphics[width=0.16\linewidth]{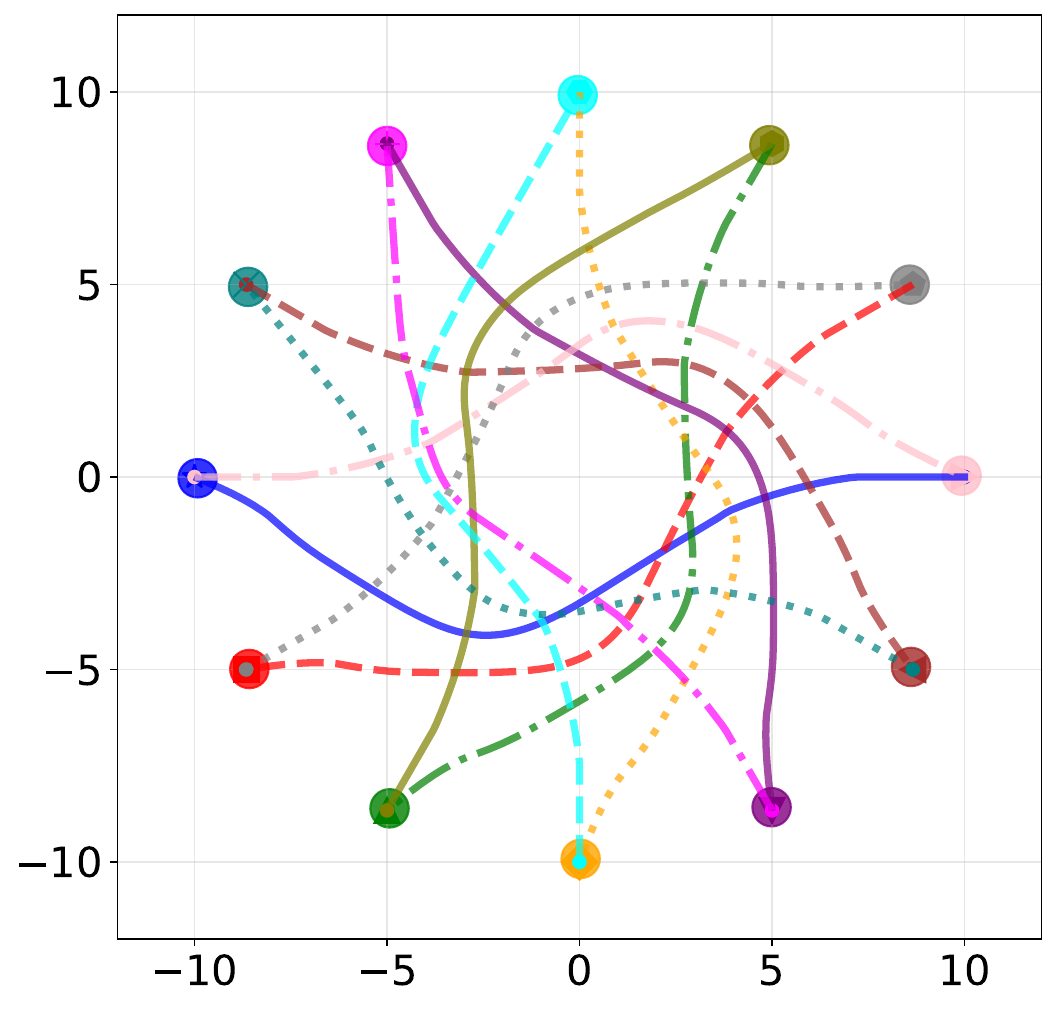}}
    
    \caption{Trajectory visualizations of the reciprocal circle interchange benchmark under varying agent densities. The columns compare six planning paradigms: ORCA, DWA, MPC-CBF, AVO, AVOCADO, and the proposed GAO framework. The rows correspond to increasing spatial congestion, with $N=4$ (top), $N=8$ (middle), and $N=12$ (bottom) agents. SICNav is unsuccessful in all three configurations and is omitted from the trajectory comparison. As the agent count increases, the trajectories show differences in how the planners pass through the central region. GAO completes the three illustrated interchange configurations.}
    \label{fig:sim_exp_2_comparison}
\end{figure*}

\indent\textit{Results and Discussion:}
Fig. \ref{fig:sim_exp_2_comparison} compares the circle-interchange trajectories for different agent counts ($N \in \{4, 8, 12\}$).
The trajectories show how the planners resolve head-on conflicts near the center, while Table \ref{tab:sim_2_results} reports the corresponding navigation time $t_f$.

\begin{table}[htbp]
    \centering
    \caption{Navigation Time $t_f$ (s) in the Circle Interchange Benchmark Across Varying Agent Densities. Failed denotes a collision.}
    \label{tab:sim_2_results}
    \begin{tabular}{lccc}
        \toprule
        \multirow{2}{*}{\textbf{Algorithm}} & \multicolumn{3}{c}{\textbf{Number of Agents ($N$)}} \\
        \cmidrule(lr){2-4}
        & \textbf{$N=4$} & \textbf{$N=8$} & \textbf{$N=12$} \\
        \midrule
        GAO (Ours) & \textbf{12.4} & \textbf{13.0} & \textbf{13.3} \\
        ORCA       & \textbf{12.4} & 14.5 & 22.1 \\
        DWA        & 15.5 & 17.8 & 23.5 \\
        MPC-CBF    & 13.5 & 14.1 & 16.6 \\
        AVO        & 12.6 & 13.9 & 20.6 \\
        AVOCADO    & 13.3 & 13.8 & 19.3 \\
        SICNav     & Failed & Failed & Failed \\
        \bottomrule
    \end{tabular}
\end{table}

GAO completes all three tested circle-interchange configurations without collision or system-wide stagnation.
The trajectories show agents passing through the central region toward their antipodal goals.
Its completion time increases only slightly across the three agent counts.
At $N=12$, it completes the task in 13.3 s, the fastest successful method.

SICNav is unsuccessful at all three agent counts under the evaluated implementation.
Its prediction model assumes that neighboring pedestrians react according to ORCA~\cite{sicnav}.
In our reciprocal tests, every agent executes SICNav, creating a mismatch between the assumed response policy and the controllers used by its neighbors.
This mismatch between prediction and execution may contribute to the observed failures.
The original study evaluates one SICNav robot among simulated and real pedestrians.
It does not report reciprocal navigation experiments in which every agent runs SICNav~\cite{sicnav}.

\subsection{Large-Scale Reciprocal Collision Avoidance}
\label{sec:sim_large_scale}
To enable efficient computation in large-scale multi-agent scenarios, we implement the solver in C++ and introduce parallelization to accelerate multi-agent computations.
We then extend the circle-interchange benchmark to $N\in\{64,200,1000\}$ agents to examine collective completion and computation efficiency at larger scales.
Agents start at rest with antipodal goals on circles of radii $64$, $200$, and $1000\,\mathrm{m}$, maintaining similar initial spacing.
Each scale is evaluated once without obstacles and once with $4$, $8$, or $20$ fixed circular obstacles, respectively.
Completion time $t_f$ is recorded when all agents settle at their goals.

\begin{table}[htbp]
    \centering
    \caption{Large-Scale Multi-Agent Navigation: Completion, Minimum Surface Clearance, and Computational Cost}
    \label{tab:sim_large_scale_results}
    \footnotesize
    \setlength{\tabcolsep}{2.5pt}
    \begin{tabular}{cccccccc}
        \toprule
        \multirow{2}{*}{\textbf{$N$}} & \multirow{2}{*}{\textbf{Obs.}} & \multirow{2}{*}{\textbf{Reached}} & \multirow{2}{*}{$t_f$ (s)} & \multirow{2}{*}{$d_{\min}$ (m)} & \multicolumn{3}{c}{\textbf{Team update (ms)}} \\
        \cmidrule(lr){6-8}
        & & & & & \textbf{Mean} & \textbf{P95} & \textbf{Active} \\
        \midrule
        64   & 0  & 64/64     & 130  & 0.19 & 0.41 & 1.12  & 0.71 \\
        64   & 4  & 64/64     & 142  & 0.13 & 0.38 & 0.92  & 0.64 \\
        200  & 0  & 200/200   & 290  & 0.11 & 1.30 & 3.15  & 1.55 \\
        200  & 8  & 200/200   & 307  & 0.15 & 1.40 & 3.15  & 1.94 \\
        1000 & 0  & 1000/1000 & 1221 & 0.10 & 3.44 & 10.70 & 4.03 \\
        1000 & 20 & 1000/1000 & 1248 & 0.05 & 3.99 & 13.02 & 4.68 \\
        \bottomrule
    \end{tabular}

    \par\smallskip
    \begin{minipage}{\columnwidth}
        \footnotesize
        Mean and P95 are computed over all control steps. Active denotes the mean computation time before any agent reaches its goal. Team updates are performed using 16 threads.
    \end{minipage}
\end{table}

\begin{figure}[htbp]
    \centering
    \captionsetup[subfloat]{farskip=0pt,nearskip=0pt,captionskip=2pt}
    \subfloat[64 agents, no obstacles\label{fig:sim_large_scale_64_noobs}]{%
        \includegraphics[width=0.48\columnwidth]{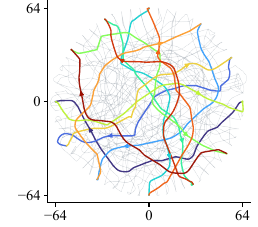}}\hfill
    \subfloat[64 agents, 4 obstacles\label{fig:sim_large_scale_64_obstacles}]{%
        \includegraphics[width=0.48\columnwidth]{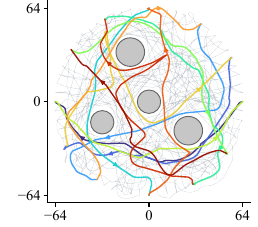}}
    \par\vspace{1pt}
    \subfloat[200 agents, no obstacles\label{fig:sim_large_scale_200_noobs}]{%
        \includegraphics[width=0.48\columnwidth]{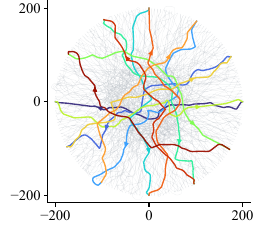}}\hfill
    \subfloat[200 agents, 8 obstacles\label{fig:sim_large_scale_200_obstacles}]{%
        \includegraphics[width=0.48\columnwidth]{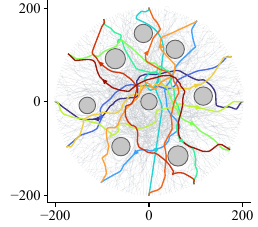}}
    \par\vspace{1pt}
    \subfloat[1,000 agents, no obstacles\label{fig:sim_large_scale_1000_noobs}]{%
        \includegraphics[width=0.48\columnwidth]{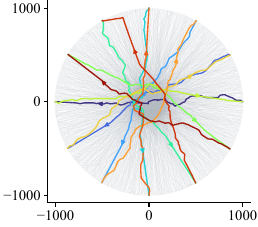}}\hfill
    \subfloat[1,000 agents, 20 obstacles\label{fig:sim_large_scale_1000_obstacles}]{%
        \includegraphics[width=0.48\columnwidth]{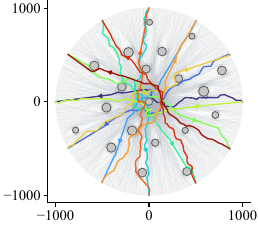}}
    \par\vspace{1pt}
    \hspace*{0.03\columnwidth}%
    \subfloat[Center detail\label{fig:sim_large_scale_1000_noobs_detail}]{%
        \includegraphics[width=0.43\columnwidth]{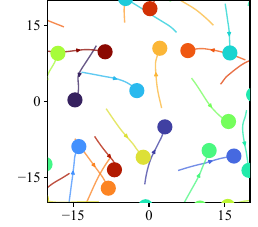}}\hfill
    \subfloat[Obstacle-edge detail\label{fig:sim_large_scale_1000_obstacles_detail}]{%
        \includegraphics[width=0.43\columnwidth]{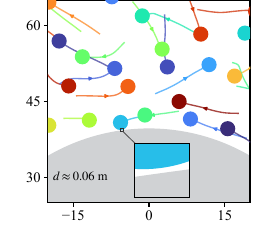}}\hspace*{0.03\columnwidth}

    \caption{Large-scale multi-agent circle-interchange scenarios. (a)–(f) show agent trajectories at different population scales, with 12 uniformly distributed agents highlighted in color. (g)–(h) show enlarged views of local interactions at $t=560\,\mathrm{s}$, including agent footprints and $6\,\mathrm{s}$ motion trails. Gray lines denote trajectories and gray disks represent static obstacles. All coordinates are in meters.}
    \label{fig:sim_large_scale}
\end{figure}

All agents reach their goals in the six deterministic cases (Table~\ref{tab:sim_large_scale_results}). Table~\ref{tab:sim_large_scale_results} also reports the average computation time before any agent reaches its goal, providing a reference for the computational cost when all agents are active. Mean team update time increases from below $1\,\mathrm{ms}$ for 64 agents to approximately $4\,\mathrm{ms}$ for 1,000 agents. The 95th-percentile time in the largest obstacle scene is approximately $13\,\mathrm{ms}$. Timing covers neighbor processing, constraint construction, optimization, and state updates. Obstacle scenes take longer to complete due to the additional constraints, while the minimum surface clearance $d_{\min}$ remains positive throughout every run.
Fig.~\ref{fig:sim_large_scale} illustrates navigation across different population sizes, showing how agents negotiate shared spaces and navigate around static obstacles, with enlarged views depicting local interactions in detail. 

\begin{figure}[t]
    \centering
    \captionsetup[subfloat]{farskip=0pt,nearskip=0pt,captionskip=2pt}
    \subfloat[Overview\label{fig:sim_isaac_overview}]{%
        \includegraphics[width=0.49\columnwidth]{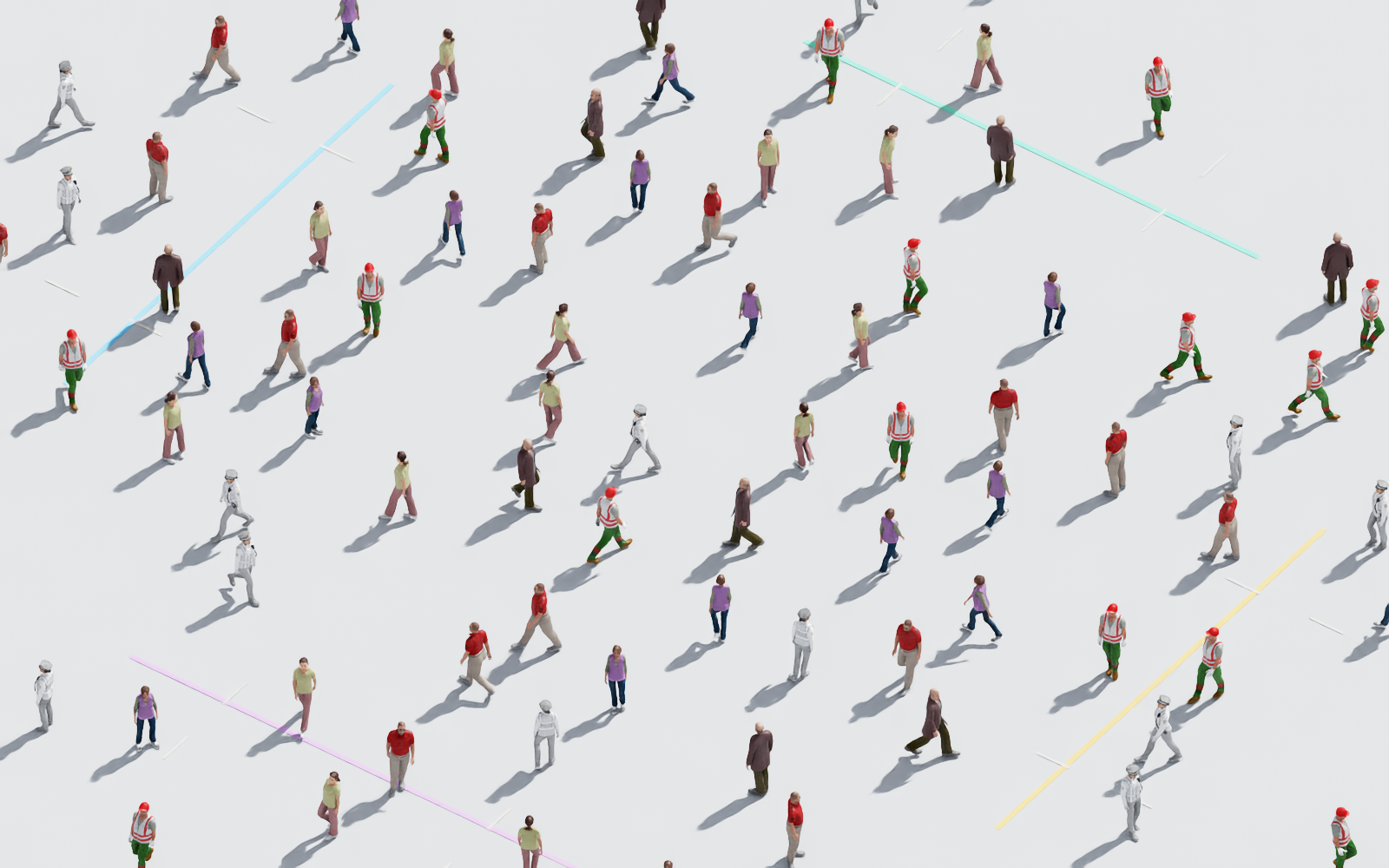}}\hfill
    \subfloat[Pedestrian interaction\label{fig:sim_isaac_interaction}]{%
        \includegraphics[width=0.49\columnwidth]{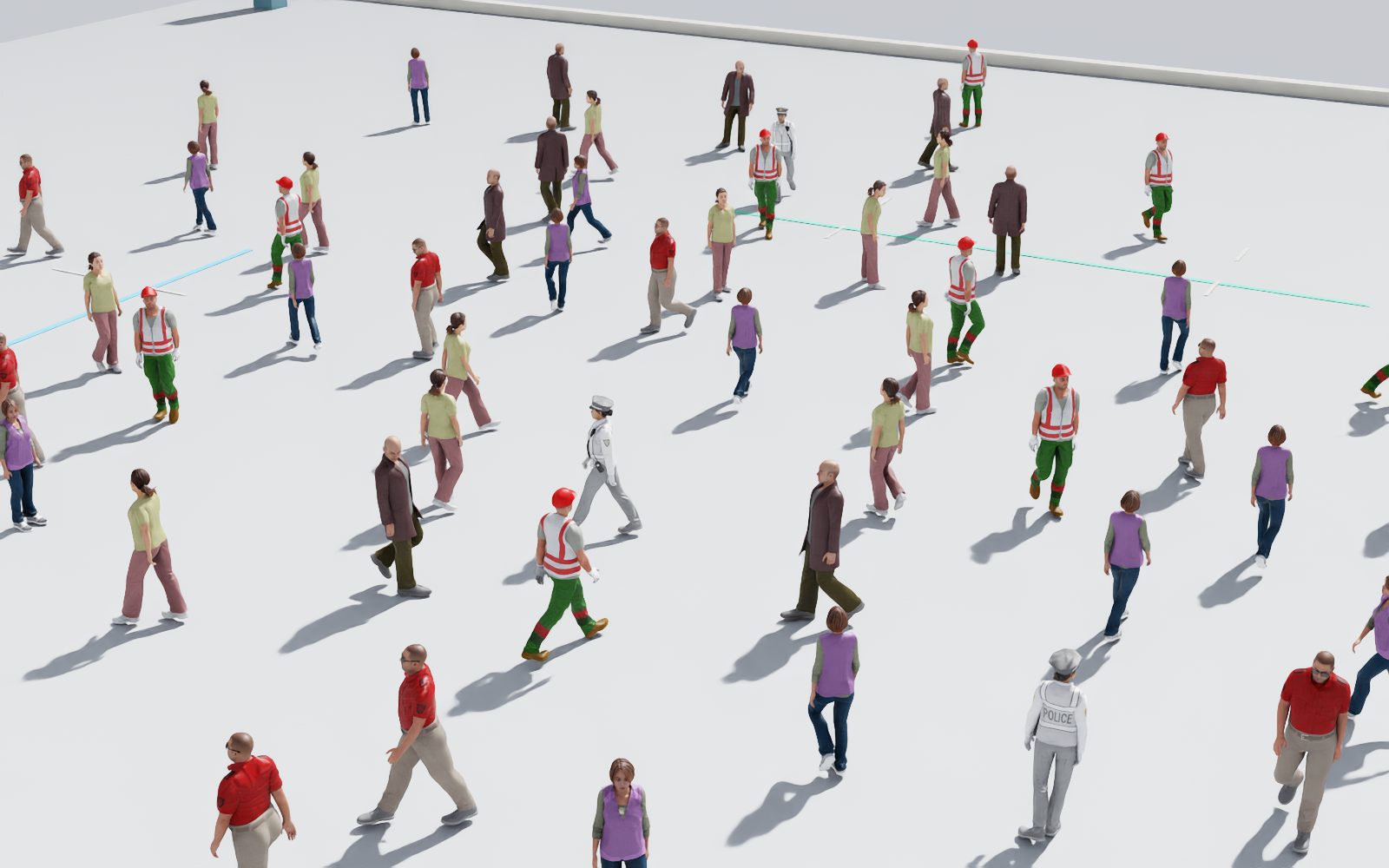}}
    \caption{GAO-driven pedestrian simulation in Isaac Sim. Two camera views of the crossing area at $t=30\,\mathrm{s}$ in the 100-pedestrian scenario, rendered from the same recorded character poses with only the viewpoint changed.}
    \label{fig:sim_isaac_pedestrians}
\end{figure}

We further integrate GAO into Isaac Sim in a 100-pedestrian scene with bidirectional and crossing flows (Fig.~\ref{fig:sim_isaac_pedestrians}).
GAO generates commands through acceleration-space optimization with affine safety constraints.
The resulting commands are converted into desired walking speeds and local navigation targets, which drive Isaac Sim's native locomotion controller to generate pedestrian gait and body motion.
Pedestrian flow continues during the 60-s recording, and mesh checks at $24\,\mathrm{Hz}$ detect no body contact.

\subsection{Overall Summary and Multi-Dimensional Evaluation}
\label{sec:overall_summary}
\begin{figure}[!t]
  \centering
  \includegraphics[width=\linewidth]{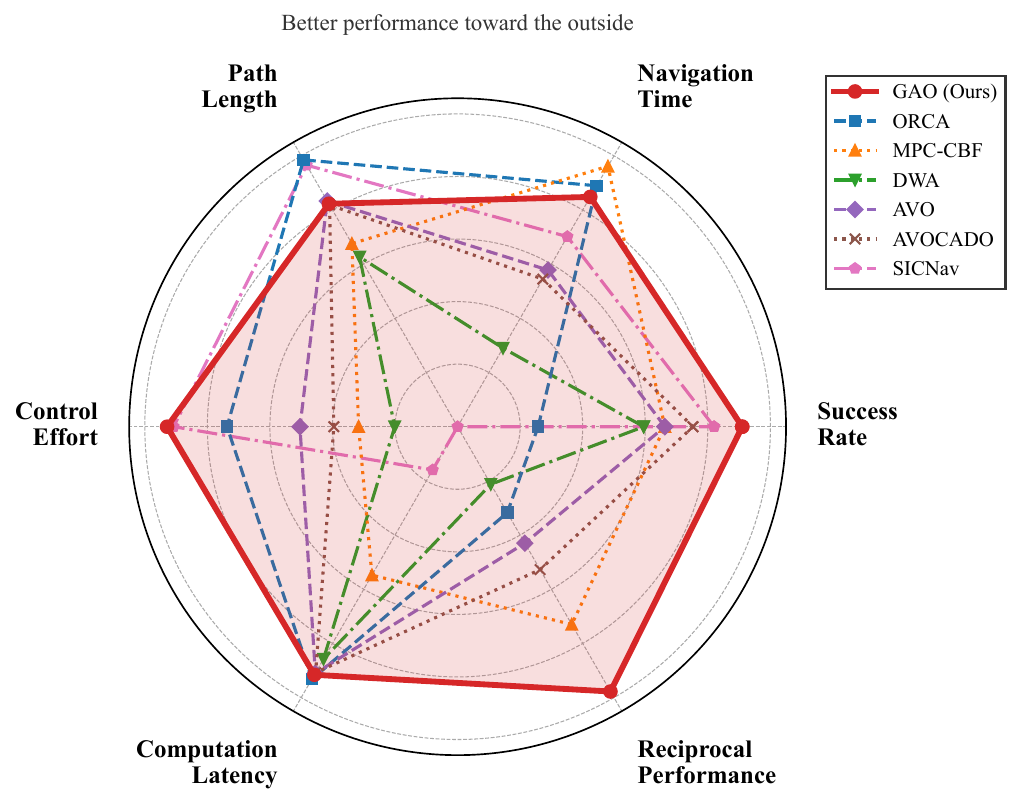}
  \caption{Performance comparison of GAO and six baselines. Radial values represent normalized performance, with better performance toward the outside on every axis. The first five axes use the $M=30$ metrics in Table~\ref{tab:sim_1_results}. Higher success rates and lower navigation time, path length, control effort, and computation latency map outward. Reciprocal performance uses the $N=12$ completion times in Table~\ref{tab:sim_2_results}, with shorter successful completion times mapped outward and failed runs scored as zero. All values are linearly normalized to $[0,1]$.}
  \label{fig:radar_chart}
\end{figure}

Fig. \ref{fig:radar_chart} summarizes the noncooperative navigation experiments (Section \ref{sec:sim_1}) and reciprocal scenarios (Section \ref{sec:sim_2}) in a six-dimensional radar chart.
This visualization shows normalized performance scores derived from success rate, navigation time, path length, control effort, and computation latency at $M=30$.
The sixth axis, reciprocal performance, is based on the $N=12$ completion time, with failed runs assigned a score of zero.
It is restricted to the comparative benchmarks in Sections~\ref{sec:sim_1} and~\ref{sec:sim_2}. 
The large-scale GAO results in Section~\ref{sec:sim_large_scale} are reported separately in Table~\ref{tab:sim_large_scale_results}.
The additional runs extend the observed collective completion to 1,000 agents in both obstacle-free and obstacle-populated scenes.

The comparison in Fig.~\ref{fig:radar_chart} highlights GAO's practical advantage in combining reliable task completion with economical control and continued progress through interactions.
For a deployed robot, reaching the goal without collision takes priority over arriving slightly earlier or taking a shorter route.
GAO achieves this balance with only modest differences in travel efficiency and a small absolute gap to the lowest mean computation latency.
The reciprocal results further show that agents resolve local conflicts while keeping the whole group moving toward its goals.

% Keep the figure-heavy closing pages compact without stretching vertical gaps.
\raggedbottom
\setlength{\textfloatsep}{10pt}
\setlength{\floatsep}{8pt}
\setlength{\intextsep}{8pt}
\section{Physical Experiments}
\label{sec:phy_exp}
We deploy GAO on physical robot platforms and test safety around noncooperative pedestrians, reciprocal avoidance among multiple robots, and behavior under high-speed operating conditions.

\subsection{Experimental Platform and System Architecture}
All experiments are conducted in a $10\text{ m} \times 10\text{ m}$ indoor motion capture area. 
The motion-capture system provides 6-DoF pose measurements at 100 Hz. 
The motion capture data are transmitted via the Virtual Reality Peripheral Network (VRPN) protocol to a central computation server.

The control system comprises state estimation, GAO planning, and command transmission.
An adaptive Kalman filter processes motion capture data to estimate the positions, velocities, and accelerations of all agents.
The GAO planner builds acceleration-space constraints from these estimates and solves the QP.
A ROS 2 control module then sends the computed commands to each robot over a 5 GHz WiFi network.
The unified control loop operates at 30 Hz in the reported experiments.

The GAO planner generates Cartesian acceleration commands.
For differential-drive platforms, these commands are converted to linear and angular commands through a headway-point mapping \cite{headway}.

\begin{figure}[htbp]
    \centering
    \begin{minipage}[c]{0.58\linewidth}
        \centering
        \subfloat[Custom mobile manipulator \label{fig:platform_1}]{%
            \includegraphics[width=\linewidth, keepaspectratio]{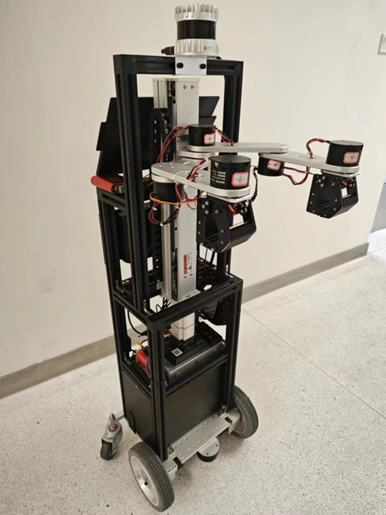}%
        }
    \end{minipage}%
    \hfill
    \begin{minipage}[c]{0.4\linewidth}
        \centering
        \subfloat[TurtleBot3 Burger \label{fig:platform_2}]{%
            \includegraphics[width=0.5\linewidth, keepaspectratio]{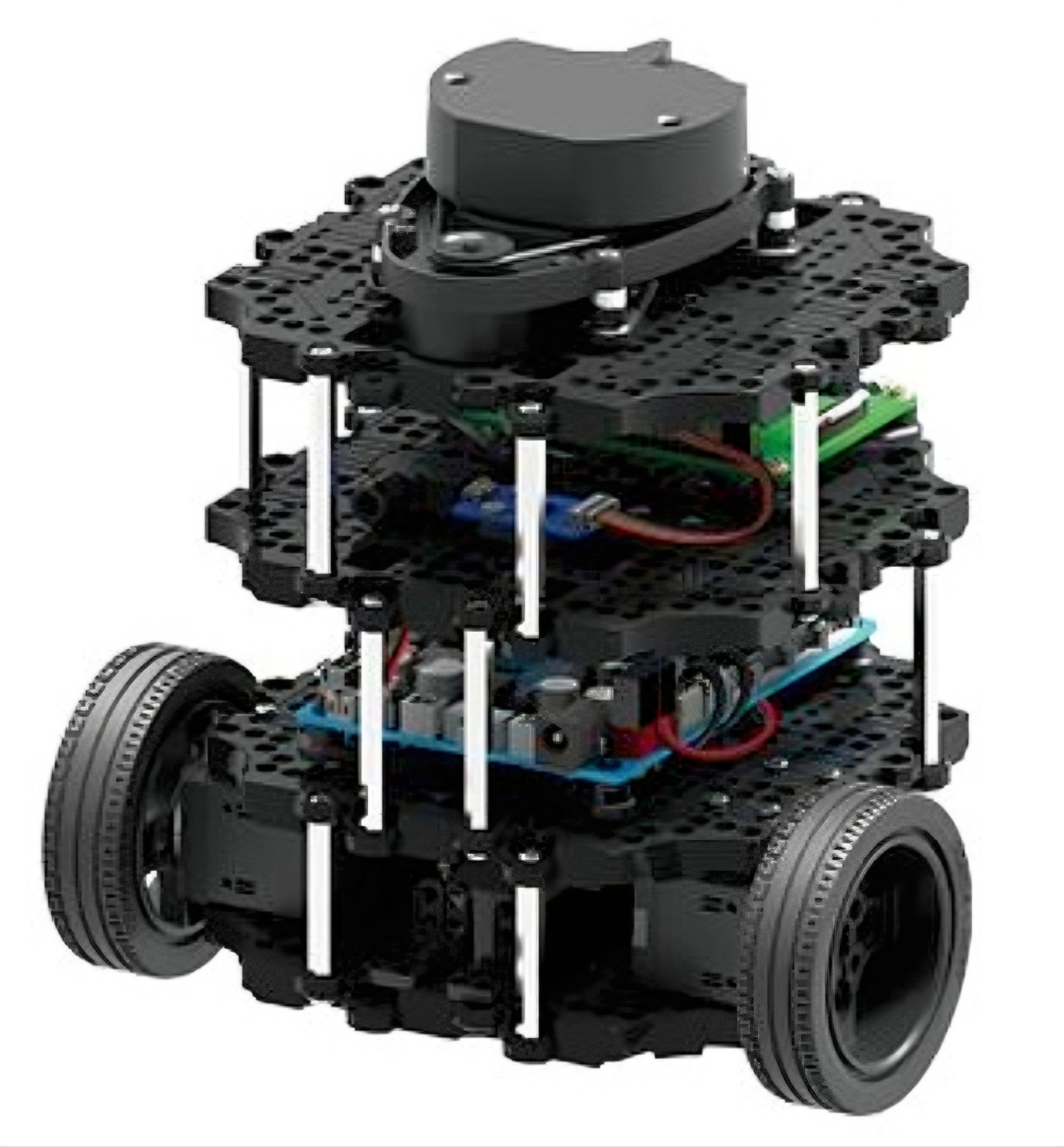}%
        } \\
        \vspace{0.2cm}
        \subfloat[AgileX Scout Mini \label{fig:platform_3}]{%
            \includegraphics[width=\linewidth, keepaspectratio]{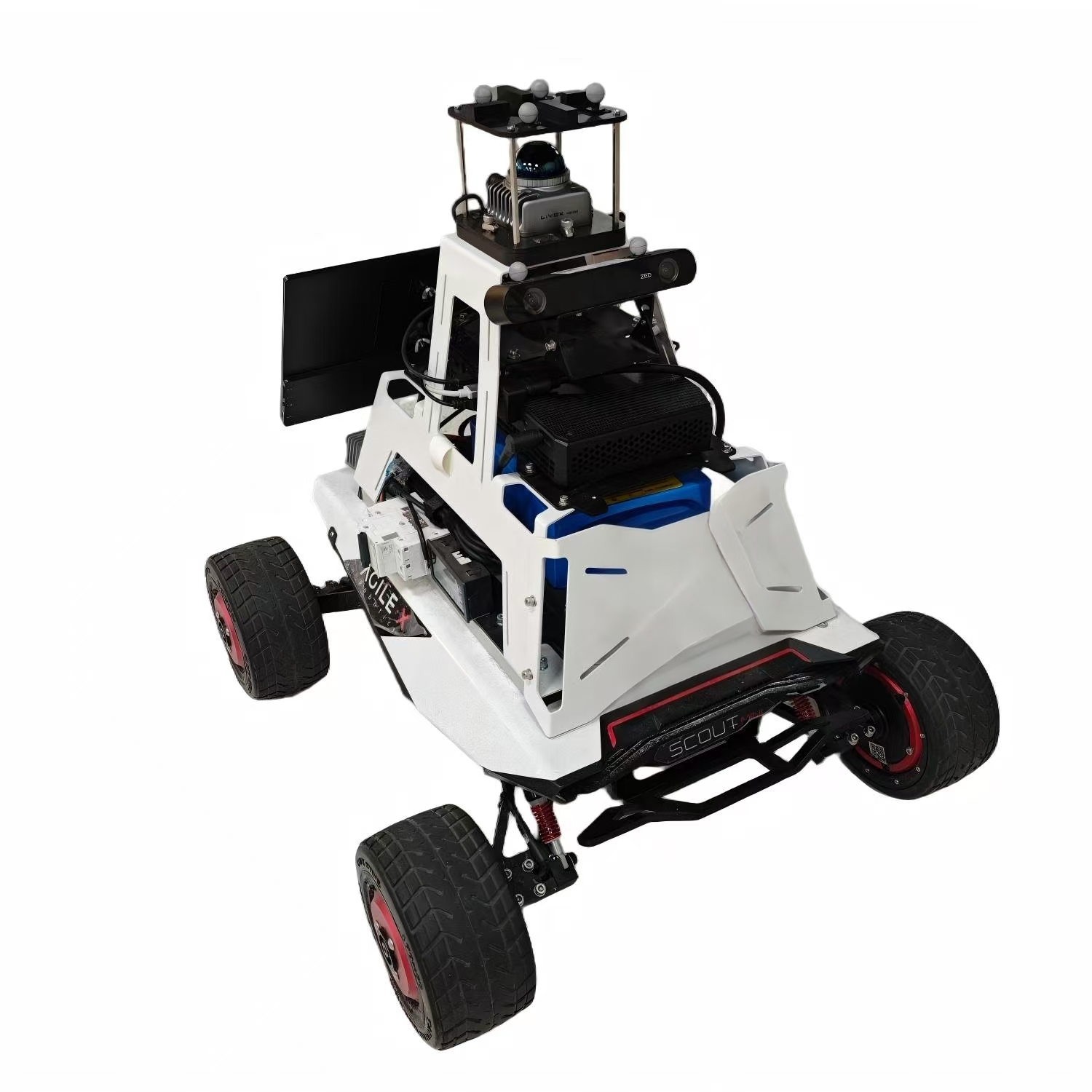}%
        }
    \end{minipage}
    \caption{The three physical mobile platforms used to evaluate the proposed GAO framework across distinct operational scenarios. (a) A custom-built differential-drive platform ($v_{\text{cruise}} = 0.5\text{ m/s}$, $a_{\max} = 1.5\text{ m/s}^2$) deployed for dynamic navigation. (b) The TurtleBot3 Burger used for the reciprocal multi-agent interchange benchmark. (c) The AgileX Scout Mini ($v_{\text{cruise}} = 2.0\text{ m/s}$, $a_{\max} = 4.0\text{ m/s}^2$) deployed for high-speed obstacle avoidance.}
    \label{fig:phy_platforms}
\end{figure}

We evaluate cross-platform applicability using three types of mobile robot:
a custom-developed differential-drive mobile robot (Fig. \ref{fig:platform_1}), 6 TurtleBot3 Burgers (Fig. \ref{fig:platform_2}), and an AgileX Scout Mini platform (Fig. \ref{fig:platform_3}).

\subsection{Navigation in Dynamic Environments}
\label{sec:phy_unstructured}
This experiment tests GAO among six noncooperative pedestrians with varying headings and accelerations.

\begin{figure}[htpb]
  \centering
  \includegraphics[width=0.97\linewidth]{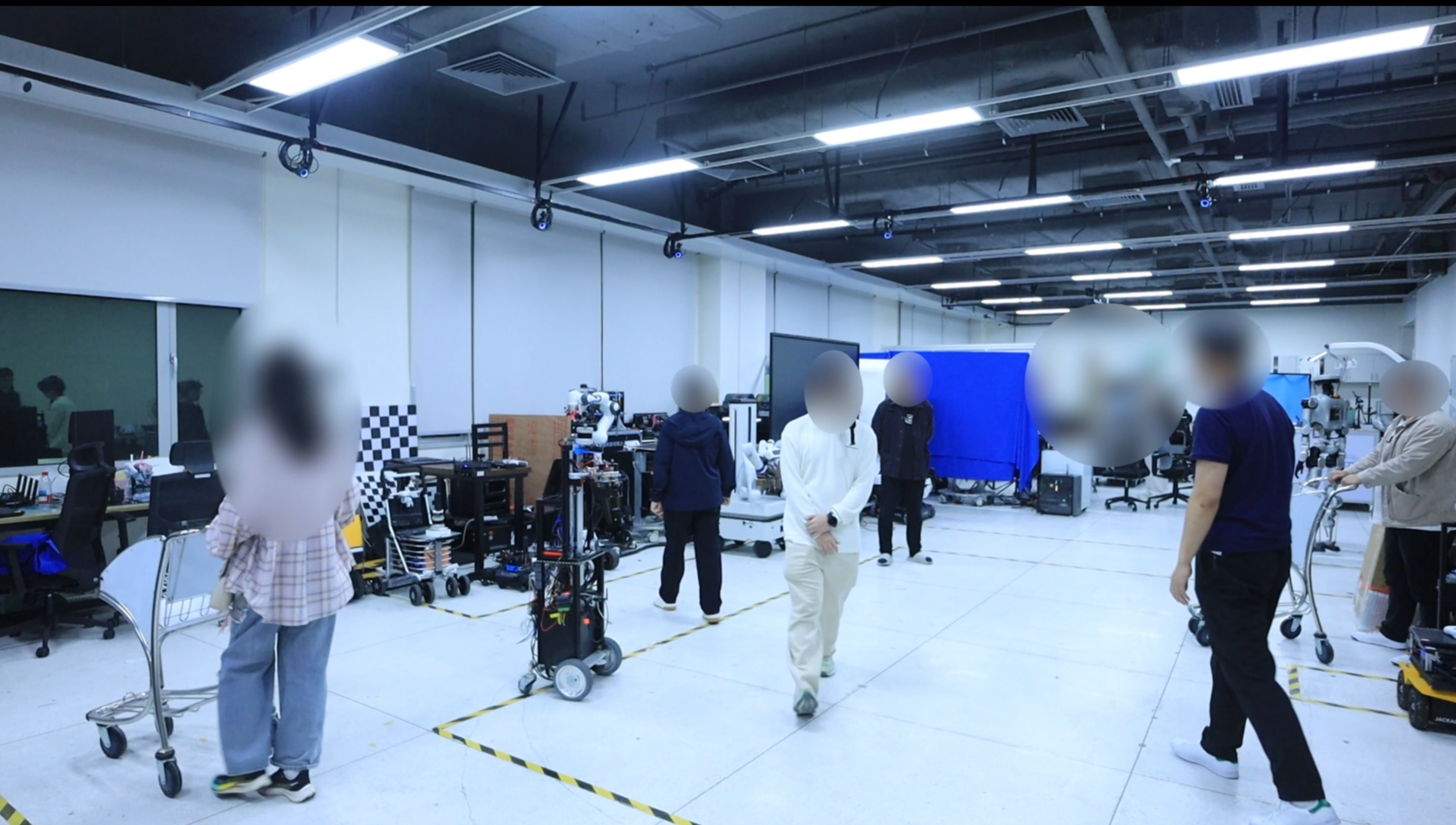}
  \caption{Experimental setup for the dynamic navigation scenario.}
  \label{fig:phy_scenario}
\end{figure}

\indent\textit{Experimental Setup:}
The robot $\mathcal{R}$ performs point-to-point navigation in the scenario shown in Fig. \ref{fig:phy_scenario}.
Its nominal global path extends from the initial position $\mathbf{p}_{\text{start}} = [-2.4, -4.2]^\top$ to the target location $\mathbf{p}_{\text{target}} = [1.0, 1.0]^\top$ (coordinates in meters).
The pedestrians move independently through the workspace, creating successive encounters that require deviations from the nominal path.

\begin{figure}[!t]
  \centering
  \captionsetup[subfloat]{farskip=0pt,nearskip=0pt,captionskip=2pt}
  \subfloat[Trajectories and local interactions\label{fig:exp_1_result_traj}]{%
    \includegraphics[width=\linewidth,trim=0bp 0bp 0bp 17bp,clip]{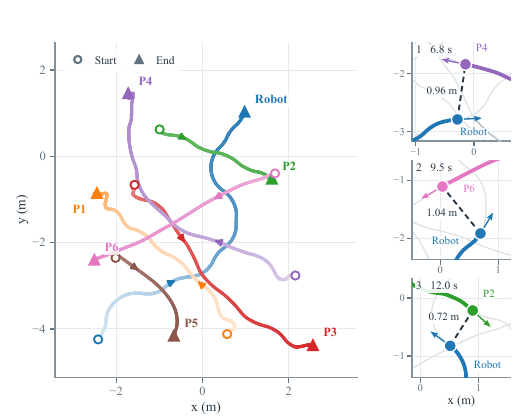}%
  }\par\vspace{2pt}
  \subfloat[Speed and acceleration\label{fig:exp_1_result_va}]{%
    \includegraphics[width=\linewidth,trim=0bp 0bp 0bp 5bp,clip]{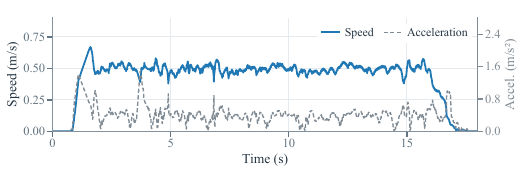}%
  }\par\vspace{2pt}
  \subfloat[Robot--pedestrian distance (log scale)\label{fig:exp_1_result_dis}]{%
    \includegraphics[width=\linewidth,trim=0bp 0bp 0bp 2bp,clip]{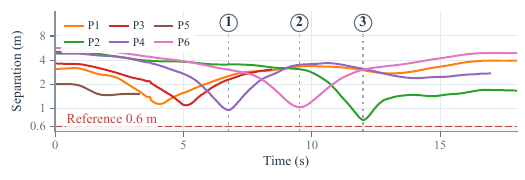}%
  }
  \caption{Physical navigation among six noncooperative pedestrians. 
  (a) Full trajectories and three local encounters. 
  Trajectory colors darken over time. 
  Circles mark the starting positions, while triangles indicate the endpoints.
  Arrows indicate motion direction. 
  Events 1--3 are the three smallest pedestrian-specific minimum distances. 
  (b) Robot speed (blue solid, left axis) and acceleration (gray dashed, right axis). 
  (c) Robot--pedestrian center distances on a logarithmic vertical axis. 
  The numbered markers correspond to (a), and the dashed red line indicates the $0.6\text{ m}$ reference distance. }
  \label{fig:exp_1_results}
\end{figure}

\indent\textit{Results and Discussion:}
The robot reaches its goal without collision, freezing, or stop-and-go motion (Fig. \ref{fig:exp_1_result_traj}).
It maintains a speed near $0.5\text{ m/s}$ during avoidance and gradually decelerates as it approaches the goal.

The highlighted encounters with P4, P6, and P2 occur at $t = 6.8$, $9.5$, and $12.0\text{ s}$, respectively (Fig. \ref{fig:exp_1_result_traj}).
The estimated acceleration varies during obstacle avoidance (Fig. \ref{fig:exp_1_result_va}).
The closest encounter occurs with P2 at $t = 12.0\text{ s}$, with a minimum center-to-center distance of $0.72\text{ m}$, above the $0.6\text{ m}$ contact threshold (Fig. \ref{fig:exp_1_result_dis}).
The curved trajectory and sustained speed indicate that these adjustments preserve progress while accommodating independently moving pedestrians.

\subsection{Reciprocal Multi-Agent Interchange}
\label{sec:phy_reciprocal}
We adapt the simulated circle-interchange benchmark to physical robots.

\begin{figure}[htpb]
  \centering
  \includegraphics[width=\linewidth]{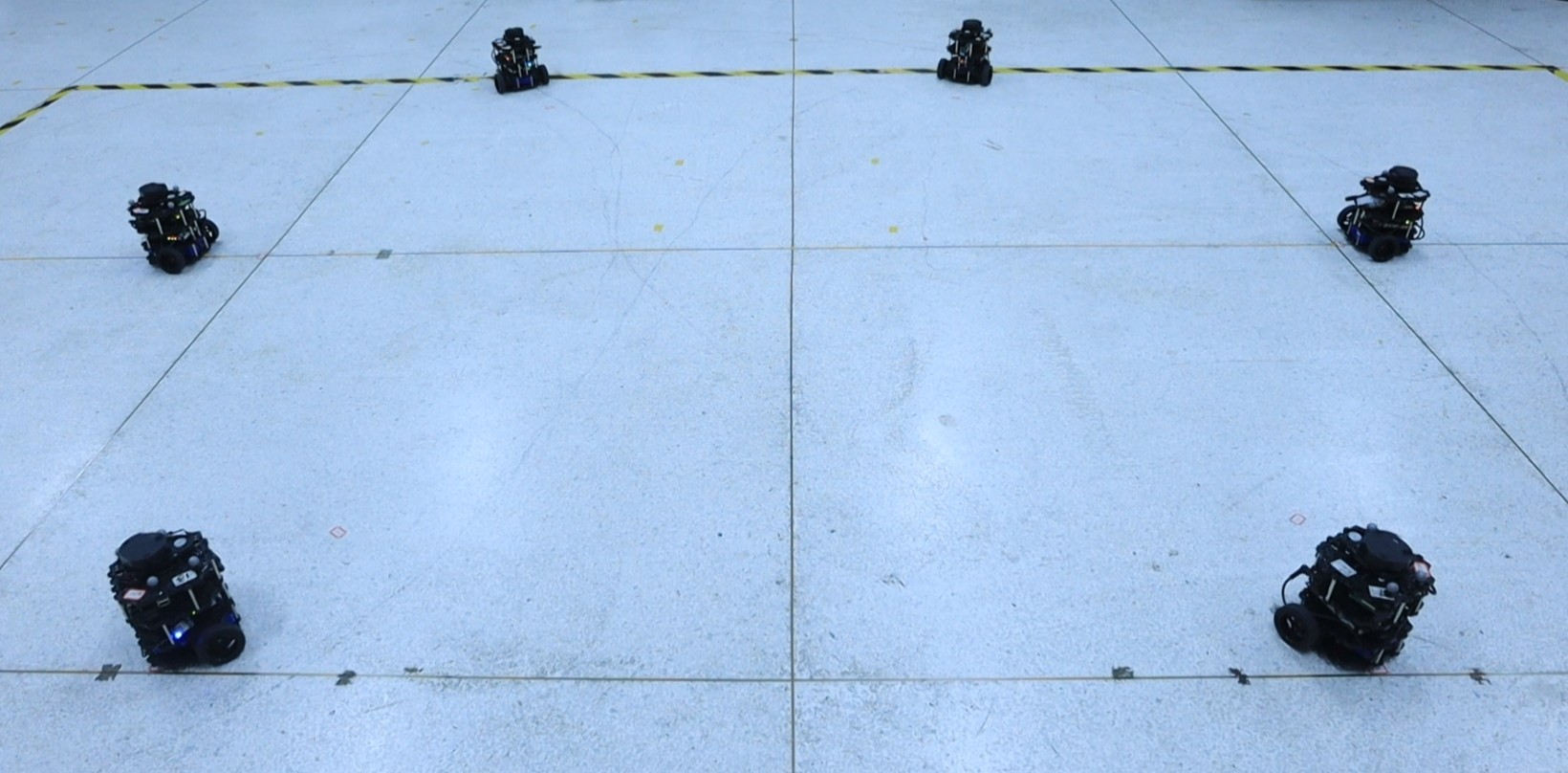}
  \caption{Experimental setup for the reciprocal multi-agent interchange.}
  \label{fig:phy_exp2_setup}
\end{figure}

\indent\textit{Experimental Setup:}
Six TurtleBot3 Burger robots are uniformly positioned on a circle of radius $1.5\text{ m}$ (Fig. \ref{fig:phy_exp2_setup}).
Each robot is assigned a goal position diametrically opposite its starting point. 
The symmetric geometry brings the robots toward the center simultaneously.
The robots initiate motion simultaneously, with GAO commands computed independently for each robot using state estimates provided by the adaptive Kalman filter.
Without communication with each other, the robots must coordinate avoidance and maintain progress toward their goals.

\begin{figure}[!h]
  \centering
  \captionsetup[subfloat]{farskip=0pt,nearskip=0pt,captionskip=2pt}
  \subfloat[Trajectories and local interactions\label{fig:phy_exp2_traj}]{%
    \includegraphics[width=\linewidth,trim=0bp 0bp 0bp 18bp,clip]{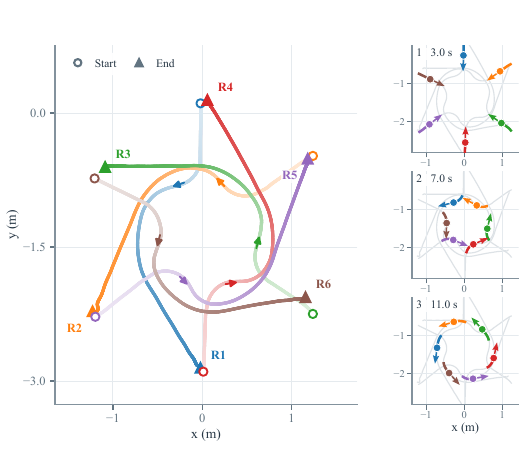}%
  }\par\vspace{2pt}
  \subfloat[Speed and acceleration\label{fig:phy_exp2_va}]{%
    \includegraphics[width=\linewidth,trim=0bp 0bp 0bp 11bp,clip]{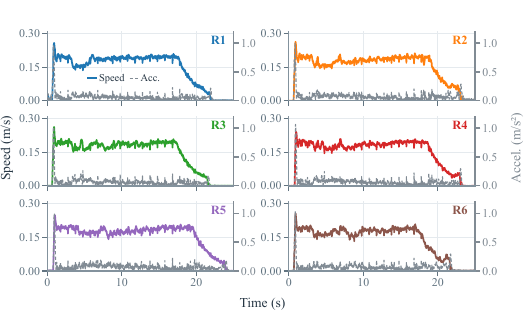}%
  }\par\vspace{2pt}
  \subfloat[Pairwise distance (log scale)\label{fig:phy_exp2_dis}]{%
    \includegraphics[width=\linewidth,trim=0bp 0bp 0bp 4bp,clip]{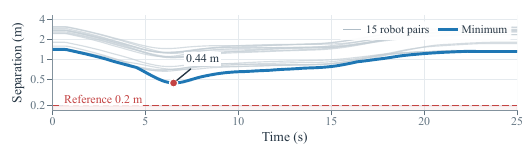}%
  }
  \caption{Six-robot reciprocal navigation. 
  (a) Full trajectories and three interaction states around the most compact group configuration. 
  Colors darken with time. 
  Circles and triangles mark the start and end positions. 
  Arrows indicate motion direction.
  (b) Individual speed (colored solid lines, left axes) and acceleration (gray dashed lines, right axes), with identical scales across robots.
  (c) All 15 pairwise center distances (gray) and their minimum (blue), with a logarithmic vertical axis and a $0.2\text{ m}$ reference (red dashed). }
  \label{fig:phy_exp2_results}
\end{figure}

\indent\textit{Results and Discussion:}
All six robots reach their antipodal goals in approximately $23.0\text{ s}$ without collision or freezing (Fig. \ref{fig:phy_exp2_traj}).
The local views at $t = 3.0$, $7.0$, and $11.0\text{ s}$ show convergence toward the center, mutual turning, and subsequent separation.
The robots maintain forward motion as they converge toward the center and subsequently separate, with modest speed adjustments (Fig. \ref{fig:phy_exp2_va}).

The acceleration profiles vary briefly during startup and stopping with smaller fluctuations during the interchange (Fig. \ref{fig:phy_exp2_va}).
The minimum pairwise distance is $0.44\text{ m}$ at $t = 6.5\text{ s}$, above the $0.2\text{ m}$ contact threshold (Fig. \ref{fig:phy_exp2_dis}).
The robots' independent avoidance decisions result in compatible turning motions as they pass through the central region.
They complete the encounter at nearly constant cruising speeds and continue toward their respective goals.

\subsection{High-Speed Obstacle Avoidance}
\label{sec:phy_high_speed}
We also evaluate GAO in a high-speed obstacle-avoidance scenario.

\indent\textit{Experimental Setup:}
The AgileX platform operates with a cruising speed of  $v_{\text{cruise}} = 2.0\text{ m/s}$ and acceleration limit of $a_{\max} = 4.0\text{ m/s}^2$.
A fast-moving obstacle approaches the robot head-on.

\begin{figure}[!htbp]
  \centering
  \captionsetup[subfloat]{farskip=0pt,nearskip=0pt,captionskip=2pt}
  \newsavebox{\gaoHighSpeedTrajectoryBox}
  \sbox{\gaoHighSpeedTrajectoryBox}{%
    \begin{minipage}[t]{0.43\linewidth}
      \centering\vspace{0pt}
      \subfloat[Trajectories\label{fig:exp_3_2_traj}]{%
        \includegraphics[width=\linewidth]{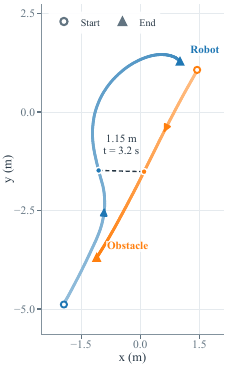}%
      }
    \end{minipage}%
  }
  \usebox{\gaoHighSpeedTrajectoryBox}\hfill
  \begin{minipage}[t][\dimexpr\ht\gaoHighSpeedTrajectoryBox+\dp\gaoHighSpeedTrajectoryBox\relax][s]{0.55\linewidth}
    \centering\vspace{0pt}\offinterlineskip
    \subfloat[Speed\label{fig:exp_3_2_speed}]{%
      \includegraphics[width=\linewidth]{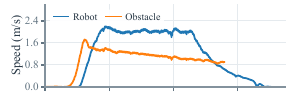}%
    }\par\vfill
    \subfloat[Acceleration\label{fig:exp_3_2_acc}]{%
      \includegraphics[width=\linewidth]{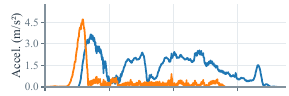}%
    }\par\vfill
    \subfloat[Distance\label{fig:exp_3_2_dis}]{%
      \includegraphics[width=\linewidth]{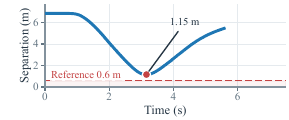}%
    }\par\vspace{0pt}
  \end{minipage}
  \caption{High-speed avoidance of a moving obstacle. (a) Robot (blue) and obstacle (orange) trajectories, with the closest encounter marked at $t = 3.2\text{ s}$. 
  Colors darken with time. 
  Circles and triangles indicate the start and end positions. 
  (b), (c) Speed and acceleration. 
  (d) Center distance, with a minimum of $1.15\text{ m}$. 
  The dashed red line marks the $0.6\text{ m}$ reference distance.}
  \label{fig:exp_3_2_result}
\end{figure}

\indent\textit{Results and Discussion:}
The obstacle reaches a speed of approximately $1.7\text{ m/s}$ and a peak acceleration of $4.7\text{ m/s}^2$.
GAO uses estimated obstacle dynamics in its local constraints to guide avoidance while maintaining the robot's speed near $2.0\text{ m/s}$ through the encounter (Fig. \ref{fig:exp_3_2_speed}).

The robot's estimated acceleration reaches a peak of $3.7\text{ m/s}^2$, remaining below the $4.0\text{ m/s}^2$ limit (Fig. \ref{fig:exp_3_2_acc}).
The minimum distance is $1.15\text{ m}$ at $t = 3.2\text{ s}$, above the $0.6\text{ m}$ reference distance (Fig. \ref{fig:exp_3_2_dis}).
The sustained speed and curved trajectory indicate that this encounter is resolved mainly through turning.
The nonzero acceleration during the encounter reflects changes in the robot's direction of motion rather than substantial speed variations.

\subsection{Experimental Summary}
The three deployments show that GAO can generate executable avoidance motions across different mobile platforms, interaction patterns, and operating speeds.
Their common feature is continued progress through path adjustments, despite differences in whether neighboring agents participate in the avoidance.
These results support practical deployment of the GAO controller with state estimation and platform-specific command conversion.

\section{Conclusion}
\label{sec:conclusion}
This paper presents Geometric Acceleration Obstacles (GAO), a framework that establishes the geometric structure of acceleration obstacles (AO) and derives affine safety certificates for real-time collision avoidance. By analyzing the clearance along the entire relative trajectory, we establish a nonpenetration criterion that distinguishes physically valid grazing candidates from hidden penetrating candidates. This result shows that satisfying contact and tangency conditions alone is insufficient to establish correspondence with the actual AO boundary.

Building on this physical characterization, we reveal the configuration-dependent geometry of AO and develop two complementary constructions of global support. For approaching configurations, the validated boundary branches exhibit strict local convexity, and their tangents provide globally supporting lines. For bypassing configurations, spatial obstacle tangents establish global support through the acceleration origin, without requiring convexity or smoothness of the collision-set boundary. Both constructions yield affine safety certificates with proven geometric tightness, leading to a unified constraint formulation in robot acceleration-input space.

By combining the resulting pairwise certificates with convex actuator constraints, we obtain a certified convex form of collision-free inputs and formulate a quadratic program for nominal command tracking. Simulation results demonstrate competitive navigation performance, reduced control effort, and millisecond-level computation. Large-scale circle-interchange experiments further demonstrate collision-free completion with up to 1,000 agents. Physical experiments validate the framework in pedestrian encounters, reciprocal multi-robot navigation, and high-speed obstacle avoidance.

The theoretical guarantees rely on initially separated disk-shaped agents. 
Future research will focus on three critical directions:
(1) extending the GAO framework to three-dimensional workspaces for aerial and underwater navigation;
(2) generalizing the exact second-order boundary mapping to accommodate arbitrary convex obstacle geometries and dynamic ellipsoids;
and (3) integrating probabilistic safety bounds or conformal prediction into the GAO formulation to handle perceptual uncertainties and sensor noise in unstructured real-world environments.
\appendices
\section{Supplementary Spatial-Support Properties}
\label{sec:support_tightness_details}
\subsection{Translational and Angular Tightness}
\label{subsec:byp_tightness}
\begin{figure}[!htbp]
\centering
\captionsetup[subfloat]{farskip=0pt}
% Crop embedded headings and top whitespace; LaTeX supplies the subfigure labels and titles.
\subfloat[Inward translation]{%
\includegraphics[width=.48\columnwidth,trim=0bp 21bp 126bp 30bp,clip]{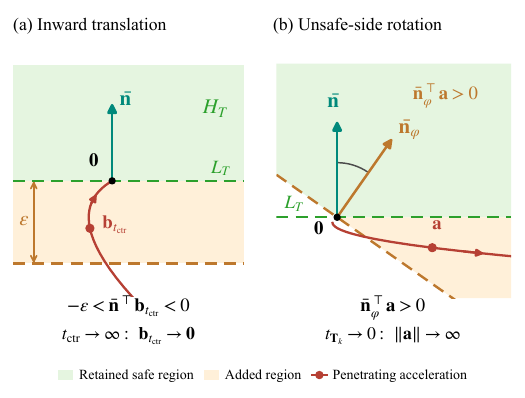}\label{fig:support_tightness_translation}}\hfill
\subfloat[Unsafe-side rotation]{%
\includegraphics[width=.48\columnwidth,trim=126bp 21bp 0bp 30bp,clip]{figs/method/support_tightness.pdf}\label{fig:support_tightness_rotation}}
\par
\includegraphics[width=\columnwidth,trim=0bp 3bp 0bp 171bp,clip]{figs/method/support_tightness.pdf}
\captionsetup{skip=2pt}
\caption{Tightness under translation and rotation. 
(a) Translating $L_T$ toward the collision set admits $\mathbf b_{t_{\mathrm{ctr}}}$ from the proof of Theorem~\ref{thm:byp_global_support}. 
(b) Rotating $\bar{\mathbf n}$ to $\bar{\mathbf n}_{\varphi}$ toward $\mathbf T_k-\mathbf p_0$ admits an acceleration that reaches $\mathbf T_k$ and then penetrates the disk. 
Green marks the retained safe region, and orange marks the newly admitted region containing the marked penetrating acceleration.}
\label{fig:support_tightness}
\end{figure}

Section~\ref{subsec:bypassing_geometry} establishes the support line $L_T$ and the certified half-plane $H_T$. We use the same selected tangent point $\mathbf T_k$, unit normal $\bar{\mathbf n}$, and $\mu=\bar{\mathbf n}^{\top}\mathbf v_0>0$ to examine translation and rotation of this line, as illustrated in Fig.~\ref{fig:support_tightness}.

We write $\bar{\mathbf n}_{\varphi}$ for the unit normal obtained by rotating $\bar{\mathbf n}$ through an angle $\varphi$.

\begin{theorem}[Geometric Tightness of Spatial Support]
\label{thm:support_tightness}
\label{prop:translation_tightness}
\label{prop:angular_tightness}
For the collision set and selected spatial tangent of Section~\ref{subsec:bypassing_geometry}, with $\mu>0$:
\begin{enumerate}
\item For every $\epsilon>0$, the translated half-plane $\{\mathbf a:\bar{\mathbf n}^{\top}\mathbf a>-\epsilon\}$ contains a penetrating acceleration.
\item In unbounded acceleration space, any rotated unit normal $\bar{\mathbf n}_{\varphi}$ satisfying $\bar{\mathbf n}_{\varphi}^{\top}(\mathbf T_k-\mathbf p_0)>0$ gives a half-plane $\{\mathbf a:\bar{\mathbf n}_{\varphi}^{\top}\mathbf a>0\}$ containing a penetrating acceleration.
\end{enumerate}
\end{theorem}

\begin{IEEEproof}
For translation, use the acceleration $\mathbf b_{t_{\mathrm{ctr}}}$ from~\eqref{eq:byp_origin_family}, which reaches the disk center and therefore strictly penetrates the disk.
By~\eqref{eq:byp_origin_projection}, its normal projection is negative and tends to zero as $t_{\mathrm{ctr}}$ increases.
Thus, for any $\epsilon>0$, there exists $t_{\mathrm{ctr}}>0$ such that
\begin{equation}
\label{eq:tightness_translation_relaxation}
-\epsilon<\bar{\mathbf n}^{\top}\mathbf b_{t_{\mathrm{ctr}}}<0.
\end{equation}
This penetrating acceleration lies in the translated half-plane, making the half-plane unsafe and thereby proving translational tightness.

For rotation, $\bar{\mathbf n}_{\varphi}^{\top}(\mathbf T_k-\mathbf p_0)>0$ selects the direction toward the spatial tangent point. Using the relative dynamics~\eqref{eq:relative_trajectory}, an acceleration that reaches $\mathbf T_k$ at time $t_{\mathbf T_k}>0$ is
\begin{equation}
\label{eq:tightness_contact_acceleration}
\mathbf a=\frac{2(\mathbf T_k-\mathbf p_0-t_{\mathbf T_k}\mathbf v_0)}{t_{\mathbf T_k}^2}.
\end{equation}
Its projection onto the rotated normal is
\begin{equation}
\label{eq:tightness_rotated_projection}
\bar{\mathbf n}_{\varphi}^{\top}\mathbf a
=\frac{2}{t_{\mathbf T_k}^2}
\left[\bar{\mathbf n}_{\varphi}^{\top}(\mathbf T_k-\mathbf p_0)
-t_{\mathbf T_k}\bar{\mathbf n}_{\varphi}^{\top}\mathbf v_0\right].
\end{equation}
The rotation condition gives
\begin{equation}
\begin{aligned}
&\lim_{t_{\mathbf T_k}\to0}
\left[\bar{\mathbf n}_{\varphi}^{\top}(\mathbf T_k-\mathbf p_0)
-t_{\mathbf T_k}\bar{\mathbf n}_{\varphi}^{\top}\mathbf v_0\right]\\
&\qquad=\bar{\mathbf n}_{\varphi}^{\top}(\mathbf T_k-\mathbf p_0)>0.
\end{aligned}
\notag
\end{equation}
Hence, for sufficiently small $t_{\mathbf T_k}>0$, the acceleration satisfies $\bar{\mathbf n}_{\varphi}^{\top}\mathbf a>0$ and lies in the rotated half-plane.
At contact, $\mathbf p(t_{\mathbf T_k};\mathbf a)=\mathbf T_k=R\bar{\mathbf n}$.
Using $\bar{\mathbf n}^{\top}(\mathbf T_k-\mathbf p_0)=0$ in~\eqref{eq:tightness_contact_acceleration}, the squared distance satisfies
\begin{equation}
\label{eq:tightness_penetration}
\begin{aligned}
\left.\frac{\mathrm d}{\mathrm dt}\|\mathbf p(t;\mathbf a)\|^2\right|_{t=t_{\mathbf T_k}}
&=2R\left(\mu+t_{\mathbf T_k}\bar{\mathbf n}^{\top}\mathbf a\right)\\
&=2R\left[\mu+\frac{2}{t_{\mathbf T_k}}\left(0-t_{\mathbf T_k}\mu\right)\right]\\
&=-2R\mu<0.
\end{aligned}
\end{equation}
Since $\|\mathbf p(t_{\mathbf T_k};\mathbf a)\|^2=R^2$, the negative derivative implies immediate penetration after contact.
Thus the half-plane is unsafe, establishing angular tightness.
\end{IEEEproof}

% A sufficiently small rotation in the opposite direction preserves safety but excludes some accelerations that promote continued passage.
% When $\mathbf v_0^\top(\mathbf T_k-\mathbf p_0)>0$, this restriction may favor braking.

\bibliographystyle{IEEEtran}
\bibliography{bib/bibliography}
\end{document}